\documentclass[10pt,twocolumn,letterpaper]{article}

\usepackage{iccv}
\usepackage{times}
\usepackage{epsfig}
\usepackage{amsmath}
\usepackage{amssymb}

\usepackage{graphicx}
\usepackage{comment}
\usepackage{amsmath,amssymb} %
\usepackage{color}
\usepackage[style=ieee]{biblatex}
\usepackage{multirow}
\usepackage[table, dvipsnames]{xcolor}
\usepackage{hhline,booktabs}
\usepackage[font=small,labelfont=small]{caption}
\usepackage{subcaption}
\usepackage{float}
\usepackage{makecell}
\usepackage{enumitem}
\usepackage[accsupp]{axessibility}  %

\usepackage[breaklinks=true,colorlinks,linkcolor=blue,citecolor=blue,bookmarks=false]{hyperref}

\newcommand{\Lagr}{\mathcal{L}}

\newcolumntype{L}[1]{>{\raggedright\let\newline\\\arraybackslash\hspace{0pt}}m{#1}}

\usepackage[breaklinks=true,bookmarks=false]{hyperref}

\iccvfinalcopy %

\begin{document}

\title{Analyzing and Mitigating JPEG Compression Defects in Deep Learning}

\author{
Max Ehrlich$^1$ \qquad Larry Davis$^1$ \qquad Ser-Nam Lim$^2$ \qquad Abhinav Shrivastava$^1$ \\
$^1$University of Maryland \qquad $^2$Facebook AI \\
{\tt \small \{maxehr, lsd\}@umiacs.umd.edu} \qquad {\tt \small sernamlim@fb.com} \qquad {\tt \small abhinav@cs.umd.edu}
}

\maketitle
\ificcvfinal\thispagestyle{empty}\fi

\begin{abstract}

    With the proliferation of deep learning methods, many computer vision problems which were considered academic are now viable in the consumer setting. One drawback of consumer applications is lossy compression, which is necessary from an engineering standpoint to efficiently and cheaply store and transmit user images. Despite this, there has been little study of the effect of compression on deep neural networks and benchmark datasets are often losslessly compressed or compressed at high quality. Here we present a unified study of the effects of JPEG compression on a range of common tasks and datasets. We show that there is a significant penalty on common performance metrics for high compression. We test several methods for mitigating this penalty, including a novel method based on artifact correction which requires no labels to train.

\end{abstract}

\section{Introduction}

The JPEG compression algorithm \cite{wallace1992jpeg} has remained the most popular compression algorithm since the early 90s. Despite rapid advances in video compression and application of those technologies to create superior still image compression algorithms, JPEG is still ubiquitous. JPEG is considered a simple solution to storage and transmission of user data. It is well supported and compresses data quite well despite glaring quality loss at low bitrates. Meanwhile, computer vision, driven by the deep learning revolution of 2012 \cite{krizhevsky2012imagenet}, reaches new milestones regularly with respect to performance, and is continually adopted in the mainstream industry and consumer-facing applications.

Despite this continuing application of deep learning to consumer methods, and the immense popularity of JPEG compression in consumer application, the effect of JPEG compression on deep learning models has been poorly studied. Many consumer applications rely on pretrained models either in whole or in part since labeled datasets are computationally expensive to create. These pretrained models are often designed for academic scenarios and use standard datasets. While many of these datasets, like ImageNet \cite{imagenet_cvpr09}, COCO \cite{lin2014microsoft}, and others, are JPEG compressed, it is often at reasonable quality levels. This is, in general, not reflective of the quality levels seen in consumer applications where the compression level is not always under the control of the deployed system. In light of this, we believe that a comprehensive study which characterizes performance loss due to JPEG compression is long overdue.

\begin{figure}
    \centering
    \includegraphics[width=0.4\textwidth]{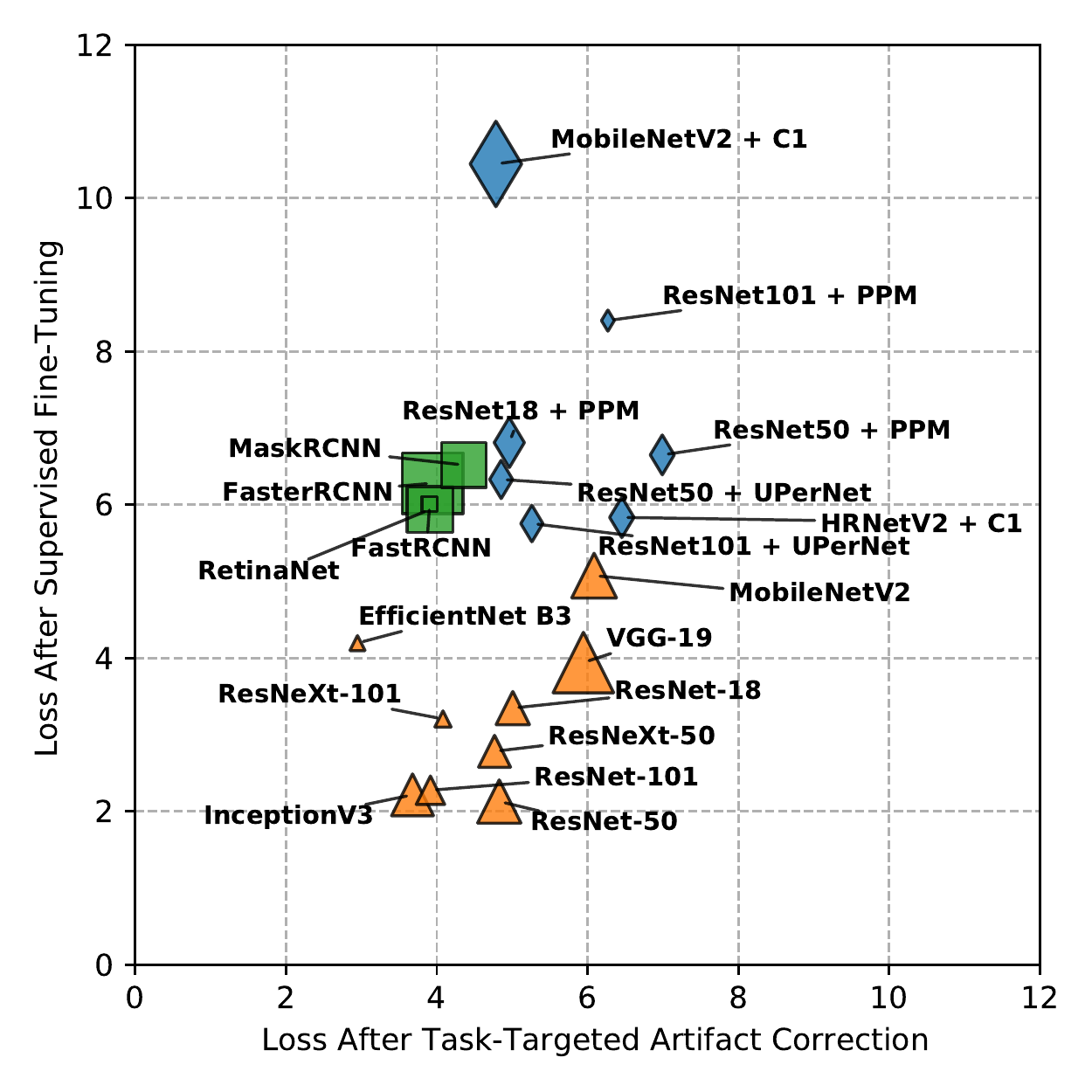}
    \vspace{-0.05in}
    \caption{Study results at a glance. Each point shows the performance loss, after applying mitigation, for a model evaluated on quality 10 JPEGs (lower is better), comparing Supervised Fine-Tuning to Task-Targeted Artifact Correction. {\color{orange} $\blacktriangle$ Orange triangles} are classification models evaluated with Top-1 accuracy on ImageNet,  {\color{OliveGreen} $\blacksquare$ Green squares} are detection models evaluated using mAP on MS-COCO, and {\color{blue}$\blacklozenge$ blue diamonds} are semantic segmentation models evaluated using mIoU on ADE20k. Point \textbf{size} shows the inference throughput (larger is better).}
    \vspace{-0.1in}
    \label{fig:glance}
\end{figure}

In this study, we quantify the effect of JPEG compression over standard performance metrics on a range of datasets and computer vision tasks. We do this by evaluating pretrained models in a systematic and consistent way. In addition to this, we also explore several strategies for mitigating any performance penalty, including a novel method which we call \textbf{Task-Targeted Artifact Correction}. This method uses the error of the downstream task logits on JPEG images  with respect to uncompressed versions of the same images, and as such is completely self-supervised: it requires no training labels making it ideal for consumer applications where labels are often expensive to obtain and unreliable. We show that for severe to moderate JPEG compression, there is a steep performance penalty and that this penalty can be mitigated effectively by using our proposed strategies. The study results are summarized in Figure~\ref{fig:glance}, which can be used to quickly choose a model and mitigation technique for a particular application.

Our study helps future scientists make informed decisions when faced with the following two important questions: (1) Is JPEG compression effecting the model? (2) If so, can anything be done about it? Our overall findings are as follows:

\begin{enumerate}[noitemsep,topsep=0pt]
    \item Heavy to moderate JPEG compression incurs a significant performance penalty on standard metrics, which we show on a set of standard computer vision tasks and datasets. We consider this study to be the primary contribution of our paper and have taken great care to ensure fair comparisons over a plethora of commonly used models. Furthermore, our benchmarking code is fully pluggable and will be released so that models can continue to be evaluated as they are developed. This study is \textbf{significantly} more comprehensive than any prior work studying any form of compression (see Section \ref{sec:pw:cd} for an overview of prior studies.)
    \item If the target application has enough labeled data, \textbf{and} preservation of the model result on uncompressed or losslessly compressed images is not important, then fine tuning the task model using JPEG as a data augmentation strategy effectively mitigates this performance loss.
    \item If the target application lacks labels, supervised training is impractical, or performance on uncompressed or losslessly compressed images must be preserved, then JPEG artifact correction can be used as a pre-processing step. We show that off-the-shelf artifact correction improves the performance of downstream tasks greatly, and our self-supervised technique makes a further improvement, approaching that of fine tuning the downstream task directly and surpassing it on some tasks all without requiring ground truth labels.
\end{enumerate}
\vspace{-0.4cm}

\section{Prior Work}

We begin by reviewing several recent works which attempt to characterize computer vision task performance under various image corruptions. We then review deep learning methods which make direct use of JPEG data and conclude with a review of artifact correction techniques which consider downstream tasks.
\vspace{-0.4cm}
\paragraph{Analysis of Compression Defects.}
\label{sec:pw:cd}
Recently, several works emerged that consider network performance in the presence of JPEG compression. In dos Santos \etal~\cite{dos2020good}, the authors revisit Gueguen \etal~\cite{gueguen2018faster} and compare the performance of the previous model under different compression settings. They find that the model still faces a performance penalty for compressed images even though it is trained on DCT coefficients. Mandelli~\etal~\cite{mandelli2020training} use two models from the EfficientNet~\cite{tan2019efficientnet} family and compare multimedia-forensics tasks (camera model identification and generated image detection) with computer vision tasks (ImageNet~\cite{imagenet_cvpr09} and LSUN~\cite{yu2015lsun} classification). The authors test on several JPEG quality factors as well as examine cropping defects (which misalign the JPEG grid). They find that while all models have a loss of performance on low-quality images, computer vision tasks tended to be more robust than multimedia-forensics tasks. Benz \etal~\cite{benz2020revisiting} show a method for improving batch normalization, which increases robustness to several image corruptions. While they do not treat compression specifically, the ImageNet-C~\cite{hendrycks2019benchmarking} dataset includes compressed images as a type of corruption. Hendrycs and Diettrich, in addition to introducing the ImageNet-C~\cite{hendrycks2019benchmarking} dataset, perform a limited benchmarking of several classification models over several JPEG compression settings, including other corruptions they introduce. An older work examining this problem by Zheng \etal~\cite{zheng2016improving} considers classification and ranking using a technique called stability training, which tries to match network output on an uncorrupted image with that of the corrupted image. They consider moderate JPEG compression.
\vspace{-0.4cm}
\paragraph{Deep Learning with JPEG Data.}

There have been several works in recent years which attempt to merge deep learning with low-level JPEG primitives. Ghosh and Chellappa~\cite{ghosh2016deep} include a DCT as part of their initial layer and show a performance improvement on classification. Gueguen \etal~\cite{gueguen2018faster} read JPEG DCT coefficients directly into their network and show that the DCT representation requires fewer parameters to learn comparable results to pixels, yielding a speed improvement. Ehrlich and Davis~\cite{ehrlich2019deep} formulate a fully JPEG domain residual network and again show a speed improvement. Lo and Hang~\cite{lo2019exploring} show a method for semantic segmentation on DCT coefficients and show both a performance and speed improvement over using pixels. Deguerre \etal~\cite{deguerre2019fast} show a method for object detection on DCT coefficients and again show performance and speed improvement over pixels. Ehrlich \etal~\cite{ehrlich2020quantization} formulate a JPEG artifact correction network on DCT coefficients and attain state-of-the-art results for color images. Choi and Han use a network to learn task-guided quantization matrices that maximize task performance after JPEG compression~\cite{choi2020task}.
\vspace{-0.4cm}
\paragraph{Artifact Correction for Task Improvement.}

Several recent works have studied using artifact correction in the presence of downstream tasks~\cite{galteri2017deep, galteri2019deep, katakol2020distributed}. Galteri \etal train a GAN~\cite{goodfellow2014generative} for JPEG artifact correction with the primary goal of making an image for human consumption. In~\cite{galteri2017deep}, they use an ensemble of networks for each JPEG quality level and manually pick the network at inference time; in~\cite{galteri2019deep} they use an auxiliary network to classify a JPEG to its quality level and automatically pick the artifact correction network at inference time. In both works, they show an improvement on object detection tasks. Katakol \etal~\cite{katakol2020distributed} tests semantic segmentation with several different compression algorithms using adversarial restoration.

\section{Methodology}
\label{sec:meth}

Our goal is to simulate a system receiving a JPEG compressed image compressed at some unknown quality level $q$\footnote{Most JPEG compression software specifies a scalar quality value in [0, 100] in lieu of a target bitrate, this quality value is not part of the JPEG standard.}. We assume that $q$ is chosen uniformly at random from the range $[10, 90]$, below quality $10$ there is little information preserved and above quality 90, the image is nearly identical to the uncompressed version. In all cases, the images are compressed before being put through any transformations that the target model prescribes\footnote{For example, cropping to $224 \times 224$ for ImageNet based models. Note that while this is more difficult to implement, the reverse process would incur an unrealistically greater loss of quality.}. For evaluation, the images are compressed at each value in the given range in steps of 10, and each model and each mitigation method are evaluated on these images. We use the Independent JPEG Group's libjpeg \cite{libjpeg} software for compression.

For any mitigation methods that require fine tuning, training images are randomly compressed from the same range as a form of data augmentation \cite{wang2019cnngenerated}. The images are always compressed at some quality factor, there are no uncompressed images used for fine tuning. Since the simulated system is assuming JPEG inputs, there is no need for non-JPEG images for fine tuning. The models are trained with a learning rate starting from $1 \times 10^{-3}$ and ending at $1 \times 10^{-6}$ for 200 epochs using a cosine annealing \cite{loshchilov2016sgdr} learning rate scheduler. We use stochastic gradient descent with momentum for the optimizer with the momentum set to $0.9$ and the weight decay set to $5 \times 10^{-4}$. As a rule, we only use validation sets for reporting final numbers as most datasets do not provide labeled test data. When training different mitigation techniques for the same model, the same batch size is used \footnote{\eg, fine tuning a ResNet 18 uses the same batch size as fine tuning artifact correction using resnet18 as the task supervision, but may not use the same batch size as COCO object detection experiments.}.

For methods requiring artifact correction, we use QGAC \cite{ehrlich2020quantization}. It is important to understand that correction-based mitigation has only recently become practical as the state-of-the-art has shifted from ``quality-aware'' models to ``quality blinded'' models. Quality aware models train a different model for each JPEG quality setting, \ie, there would be a unique model for quality 10 JPEGs, quality 20 JPEGs, \etc. This is a critical limitation because the JPEG quality is not stored in the JFIF file format, making quality-aware models impossible to use in a real setting. Theoretically, these models could be examined in the contrived scenerio of the study, however, this would require a combinatorially large number of models to be trained - one per quality level per AC model per task model. QGAC is quality-blinded meaning that a single model handles all JPEG quality settings. Additionally, it supports color images, achieves state-of-the-art performance on common benchmarks, and has public code and weights making it a prime candidate for use by other researchers. For completeness, we conducted a limited study using two common quality-aware models in Section \ref{sec:res:baseline} by limiting the compression quality settings and downstream tasks to make the training tractable.

We evaluate the models with no mitigation as a baseline to characterize the effect of JPEG compression when applied directly to machine learning models. We additionally evaluate two common mitigation methods. Finally, we propose a new mitigation method which requires no supervision to train, something which is crucial outside of the academic setting. We now briefly describe these methods.
\vspace{-0.4cm}
\paragraph{Baseline.} The baseline method simply passes the JPEG compressed evaluation images to the model unchanged. This serves as an important quantifier for how the JPEG quality affects pre-trained models. This method requires no fine tuning.
\vspace{-0.4cm}
\paragraph{Supervised Fine-Tuning.} In the Supervised Fine-Tuning mitigation, the task network is fine tuned using the protocol given above with JPEG images as input. In other words, we JPEG compress all images from the training set of the given task and fine tune the network from pretrained weights on the given task to improve performance on JPEG inputs. There are two major drawbacks to this method. The first is that it requires a training set of labeled data. This is not always easy to obtain in consumer applications. The second is that it sacrifices performance on uncompressed images in exchange for performance on compressed images. As we show in Section \ref{sec:results}, this method provides good results and a fast runtime, especially at training time (See Appendix E for throughput results).

\begin{figure}[t]
    \centering
    \includegraphics[width=0.4\textwidth]{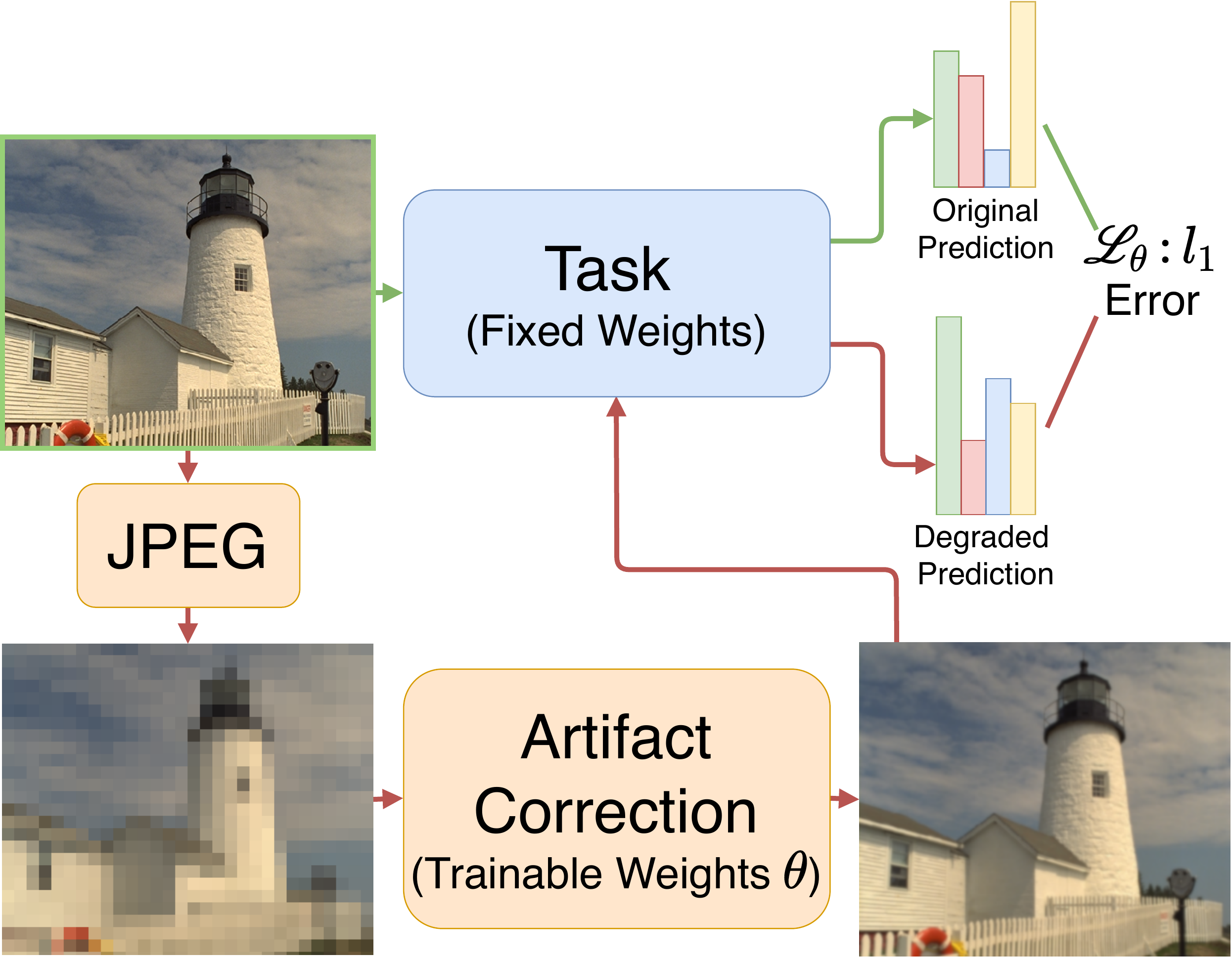}
    \caption{Task-Targeted Artifact Correction. The original image (green) is passed through the task network to obtain the unnormalized prediction logits. Then the image is JPEG compressed and artifact corrected using an artifact correction network with trainable weights. The corrected image is then passed through the task network. The $l_1$ error between the logits on the original image and the logits on the corrected image is used as loss to tune the artifact correction network weights.}
    \label{fig:ttac}
    \vspace{-0.12in}
\end{figure}

\begin{figure*}[ht!]
    \centering
    \begin{subfigure}[b]{0.33\textwidth}
        \centering
        \includegraphics[width=\textwidth]{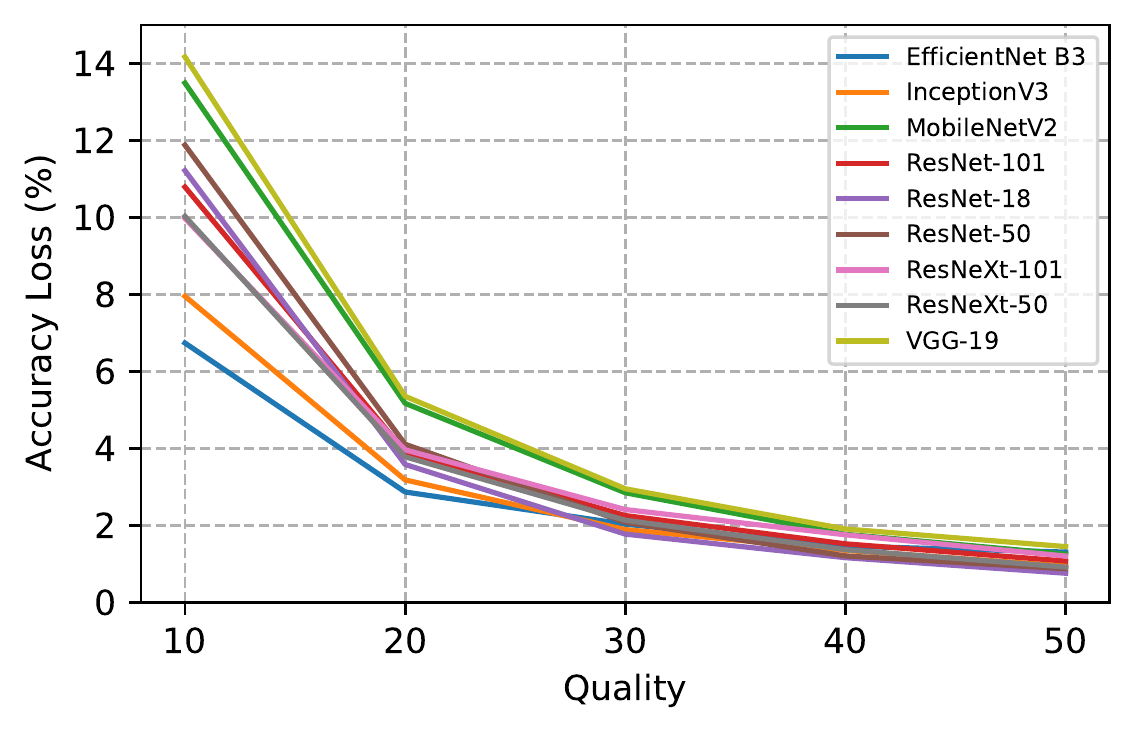}
        \caption{Classification}
    \end{subfigure}
    \begin{subfigure}[b]{0.33\textwidth}
        \centering
        \includegraphics[width=\textwidth]{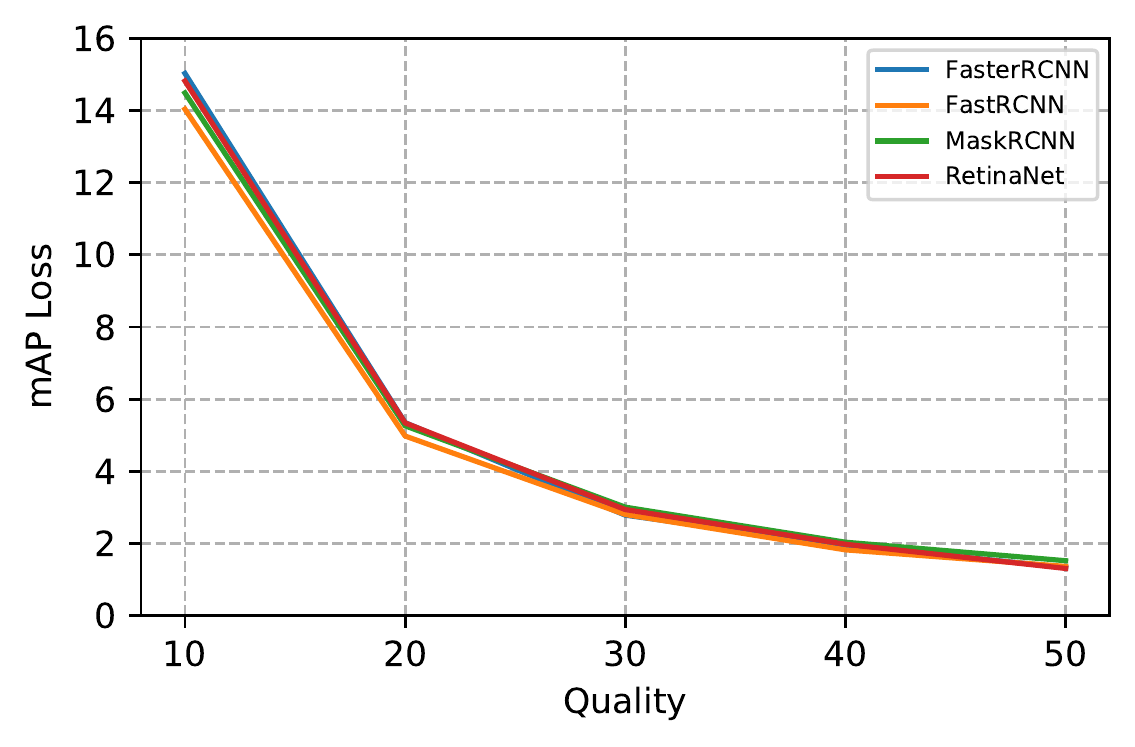}
        \caption{Detection and Instance Segmentation}
    \end{subfigure}
    \begin{subfigure}[b]{0.33\textwidth}
        \centering
        \includegraphics[width=\textwidth]{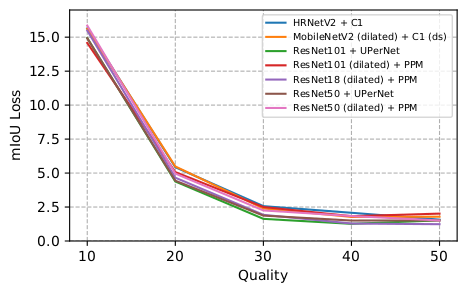}
        \caption{Semantic segmentation}
    \end{subfigure}
    \vspace{-0.2in}
    \caption{Performance loss due to JPEG compression by task. The plots show all models from a single task with no mitigation applied. For segmentation tasks, the format of the model name is \texttt{Encoder Model} + \texttt{Decoder Model} and ``ds'' indicates that the model was trained with deep supervision. Note that methods which use a Pyramid Pooling Module (PPM) decoder always use deep supervision.}
    \label{fig:nomitigation}
\end{figure*}
\begin{figure*}        \centering
    \begin{subfigure}[b]{0.33\textwidth}
        \centering
        \includegraphics[width=\textwidth]{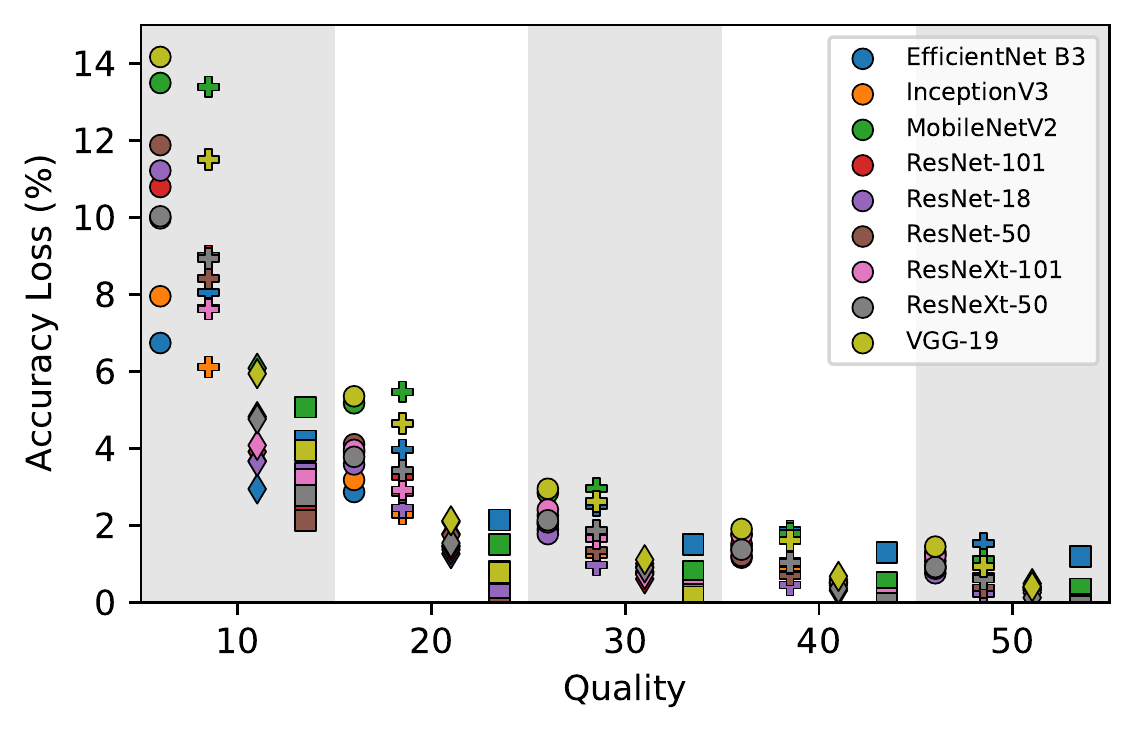}
        \caption{Classification}
    \end{subfigure}
    \begin{subfigure}[b]{0.33\textwidth}
        \centering
        \includegraphics[width=\textwidth]{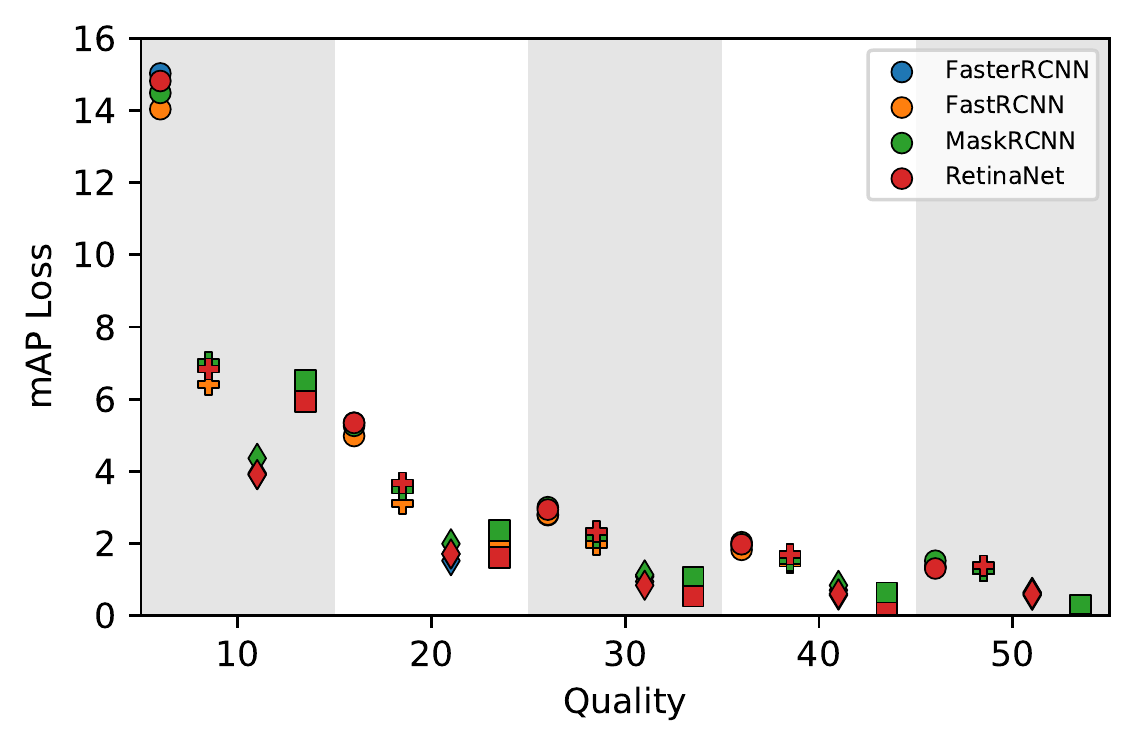}
        \caption{Detection and Instance Segmentation}
    \end{subfigure}
    \begin{subfigure}[b]{0.33\textwidth}
        \centering
        \includegraphics[width=\textwidth]{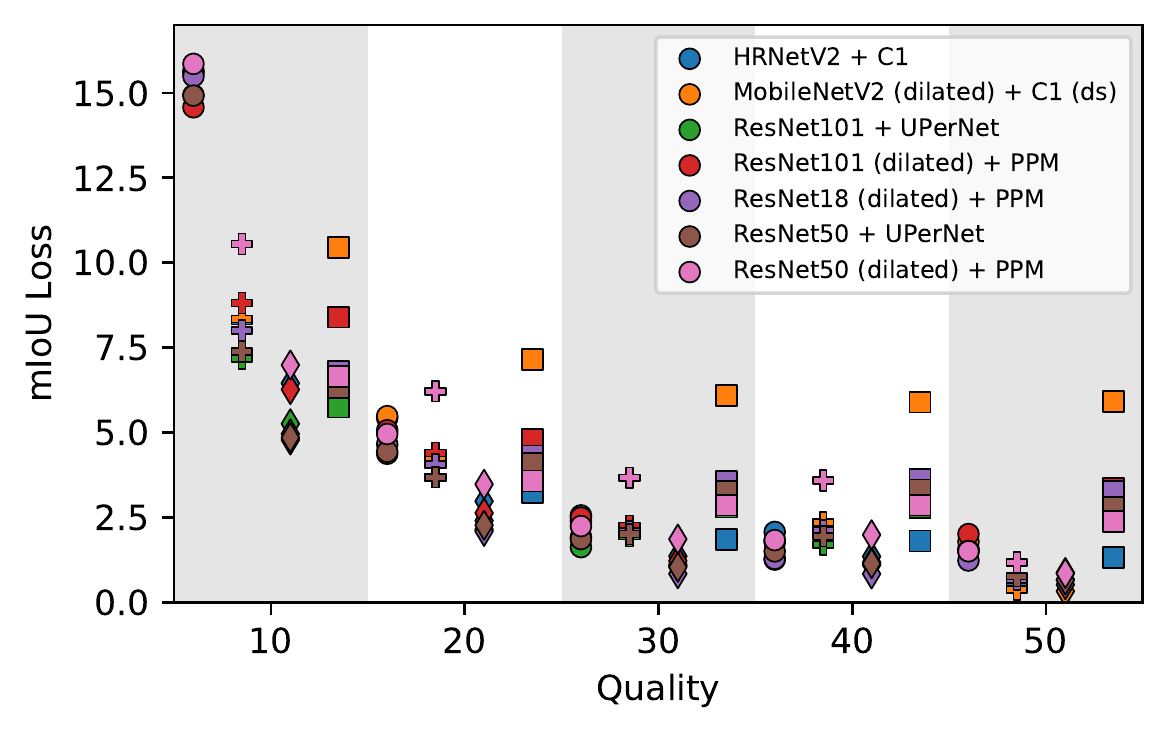}
        \caption{Semantic segmentation}
    \end{subfigure}
    \vspace{-0.2in}
    \caption{Performance loss of tested models with mitigation applied. \textbf{$\pmb{\bullet}$ Circle}: No Mitigation, \textbf{$\pmb{+}$ Cross}: Off-the-Shelf Artifact Correction, \textbf{$\blacklozenge$ Diamond}: Task-Targeted Artifact Correction, \textbf{$\blacksquare$ Square}: Supervised Fine-Tuning.  The models in this figure correspond to those shown in Figure~\ref{fig:nomitigation}. These plots are best viewed digitally and are intended to show the trend of models for each task. For plots with individual models, please see Appendix D.}
    \label{fig:withmitigation}
    \vspace{-0.2in}
\end{figure*}

\vspace{-0.4cm}
\paragraph{Artifact Correction.} For the artifact correction method, we JPEG compress the evaluation images and then perform artifact correction on them before passing them to the task model. This method requires no fine tuning, and we use pretrained weights for the artifact correction model and for the task model. This method does not sacrifice performance on clean images, since these can be detected (by file extension or mime type) and artifact correction can be skipped in this case; however, this method is not trainable and we will show in the results section that it gives worse performance than fine tuning methods.
\paragraph{Task-Targeted Artifact Correction.} In this novel mitigation technique, we fine tune the artifact correction networking using error on the task network logits. To do this, we minimize the $l_1$ distance between the task network logits computed on an uncompressed and compressed version of the same image. Formally, given a mini-batch of images $B$, we minimize
\begin{equation}
    \Lagr_\theta = \lVert \text{Task}(B) - \text{Task}(\text{AC}(\text{JPEG}_q(B) ; \theta))\rVert_1,
    \label{eq:loss}
\end{equation}
where Task is any task network, AC is the artifact correction network with parameters $\theta$, and $q$ is the JPEG quality level. This is shown schematically in Figure \ref{fig:ttac}.
In this way, the artifact correction network learns to correct in a way that maximizes the performance of the downstream network. Note that this method is entirely self-supervised, no ground truth labels are used (the only loss comes from the difference in behavior on the compressed \vs uncompressed versions of the images). Not only is this easier to deploy in a consumer setting where labeled data is hard to obtain, but it also alleviates any concern that the artifact correction network may be learning to perform the task for the task network, artificially increasing the number of parameters and making for an unfair comparison. By using $l_1$ error on the network output, the method is applicable to many different tasks with little or no modification. This method again does not sacrifice performance on uncompressed images since the task network weights are unchanged and an uncompressed input can be detected and the artifact correction step skipped. We show in Section \ref{sec:results} that this method approaches fine tuning the task network directly, even exceeding it in some tasks, despite having no access to ground truth labels.

In addition to this, Task-Targeted Artifact correction supports flexible training scenarios. We show in Section \ref{sec:results:transfer} that the networks weights are \textbf{transferrable} \eg, an artifact correction network which was trained for one network can be used for other networks. This allows a lightweight training setup where a ``fast-to-train'' network is used to create the artifact correction network weights and then reused for models which would be cumbersome to train. Similarly, Task-Targeted Artifact Correction supports \textbf{multihead} training, where multiple downstream tasks are used at the same time during training. While this method bears a superficial similarity to stability training \cite{zheng2016improving}, both transfer and multihead are impossible with stability training and the exact loss formulation we use (Equation \ref{eq:loss}) is universal: it can be easily applied to many tasks.

\section{Results}
\label{sec:results}

\begin{figure*}[ht]
    \centering
    \footnotesize
    \renewcommand{\tabcolsep}{2pt}
    \resizebox{\textwidth}{!}{
        \begin{tabular}{@{}ccccc@{}}
            \includegraphics[width=0.2\textwidth]{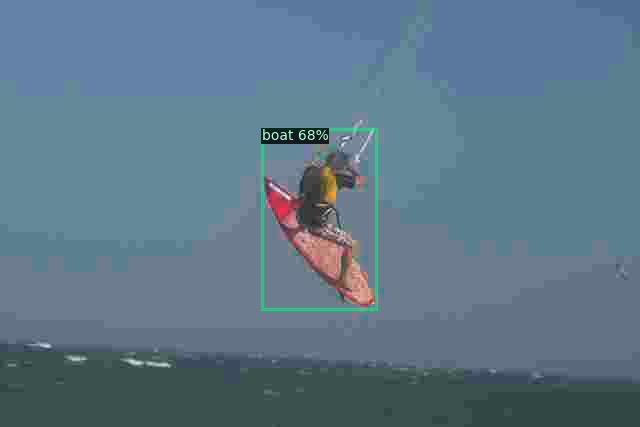}             &
            \includegraphics[width=0.2\textwidth]{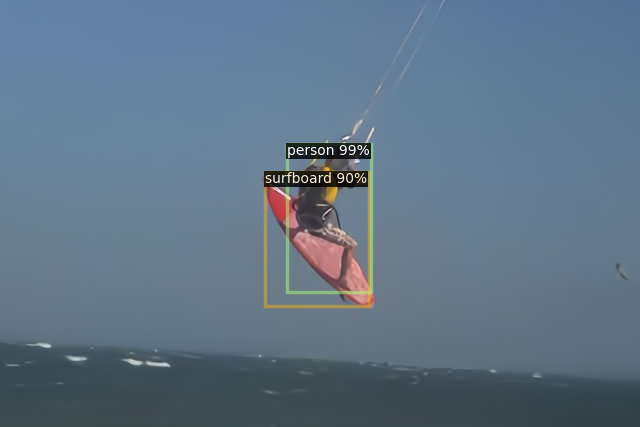}            &
            \includegraphics[width=0.2\textwidth]{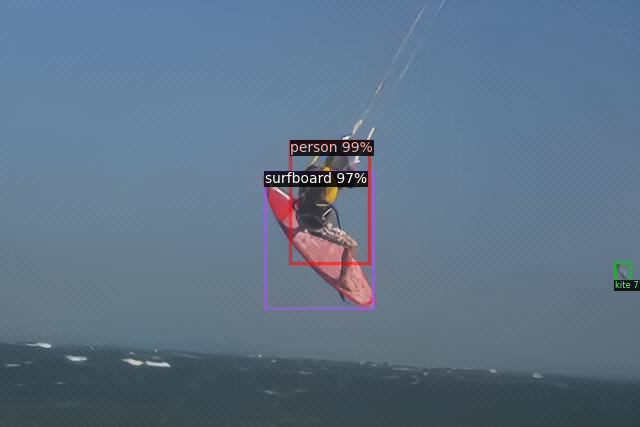}                   &
            \includegraphics[width=0.2\textwidth]{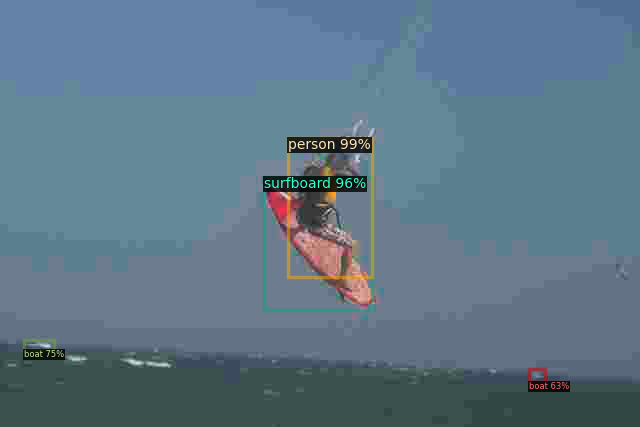}                   &
            \includegraphics[width=0.2\textwidth]{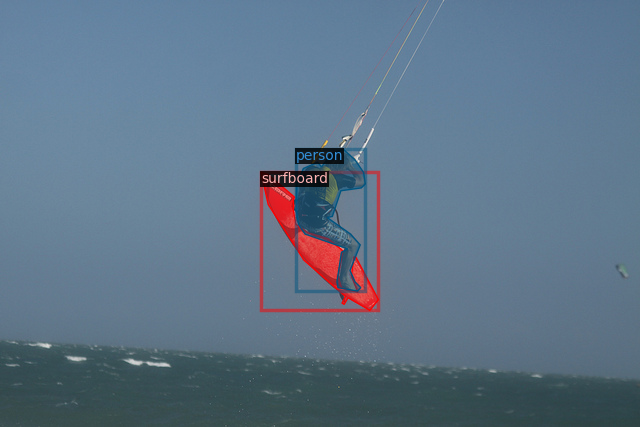}                     \\
            \includegraphics[width=0.2\textwidth]{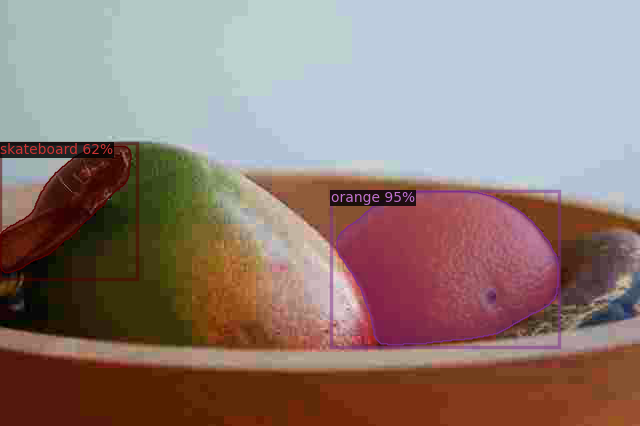}                &
            \includegraphics[width=0.2\textwidth]{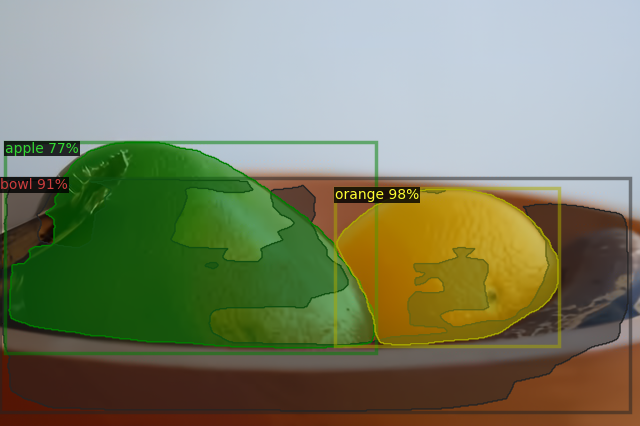}           &
            \includegraphics[width=0.2\textwidth]{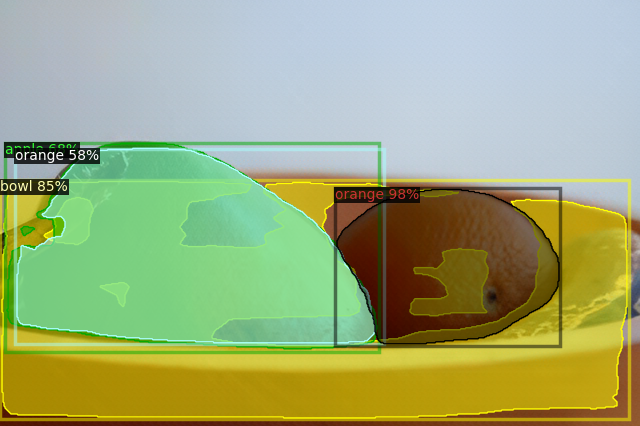}                  &
            \includegraphics[width=0.2\textwidth]{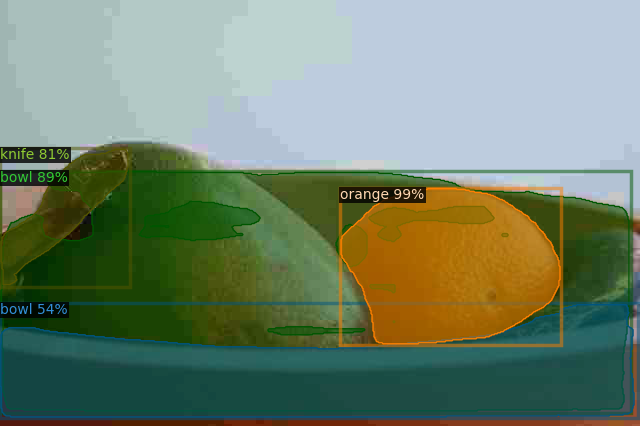}                  &
            \includegraphics[width=0.2\textwidth]{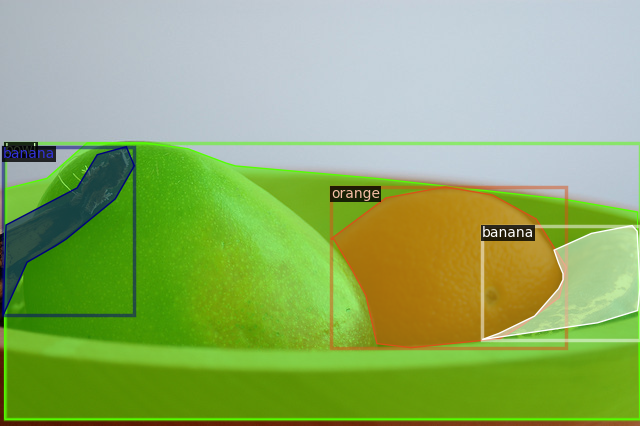}          \\
            \includegraphics[width=0.2\textwidth]{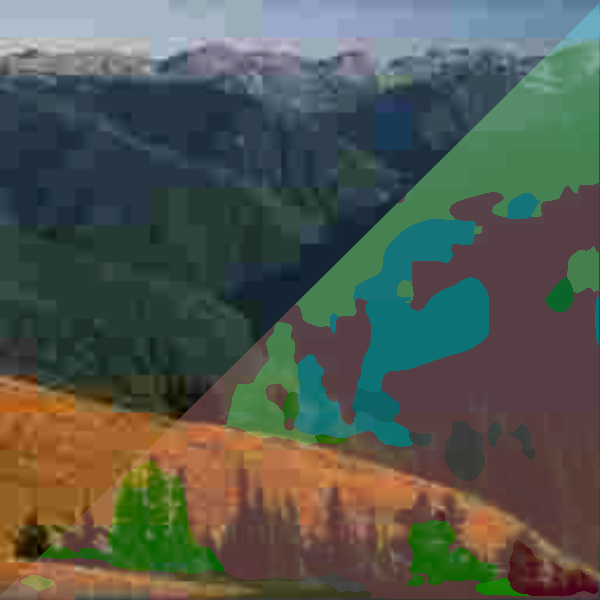}  &
            \includegraphics[width=0.2\textwidth]{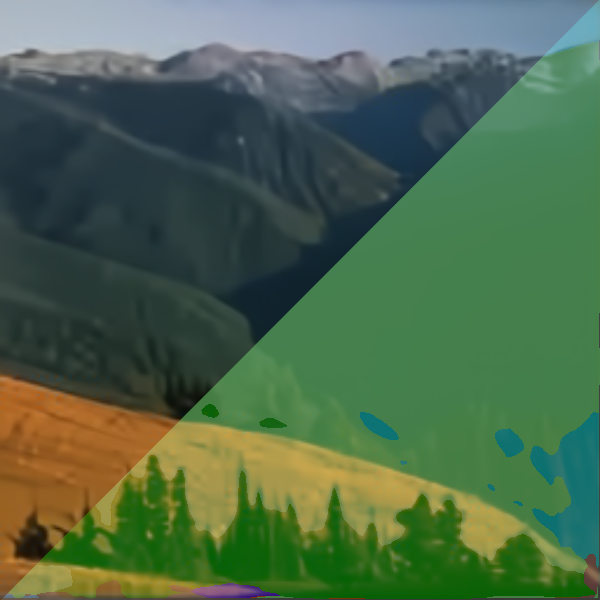} &
            \includegraphics[width=0.2\textwidth]{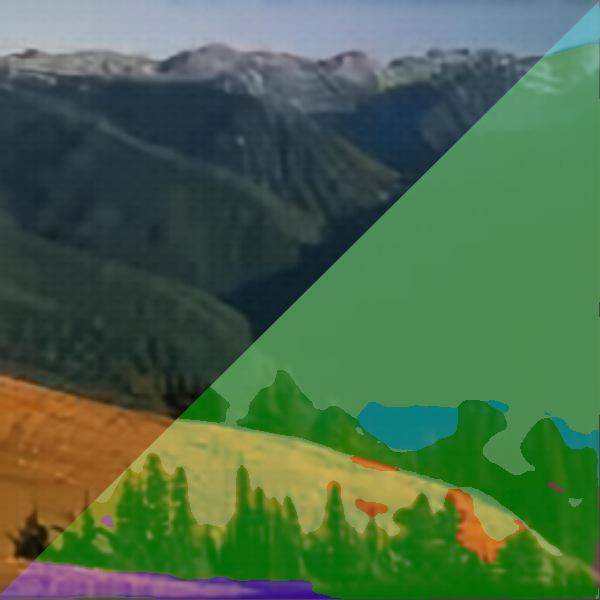}        &
            \includegraphics[width=0.2\textwidth]{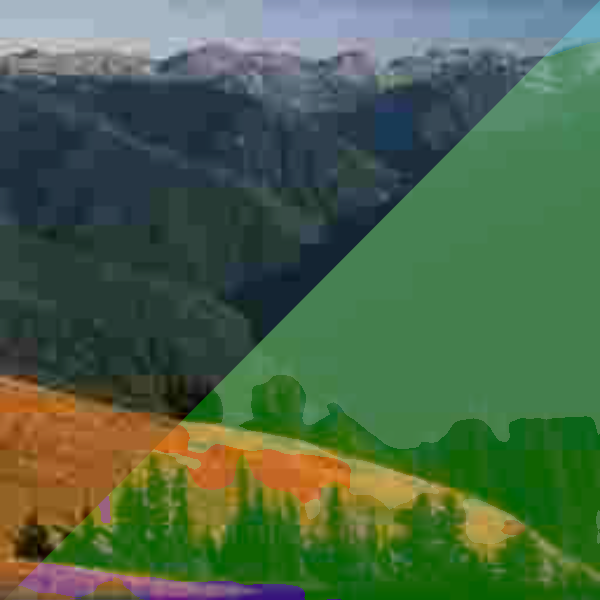}        &
            \includegraphics[width=0.2\textwidth]{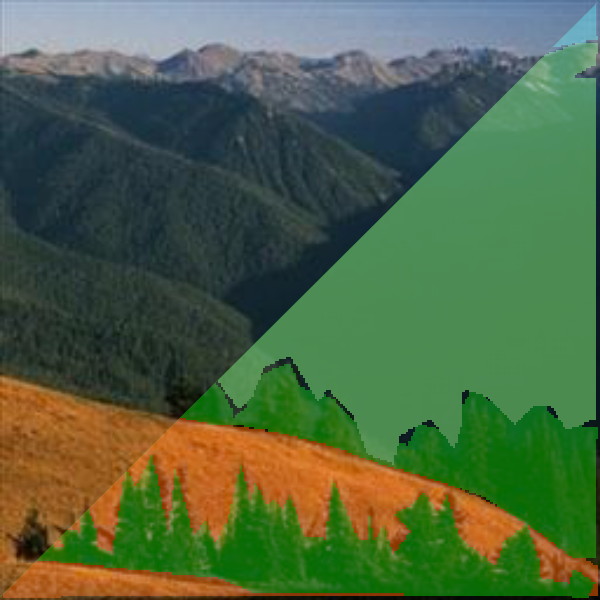}          \\
            \makecell{\textbf{No Mitigation}                                                                           \\\;}                                                                   &
            \makecell{\textbf{Off-the-Shelf}                                                                           \\\textbf{Artifact Correction}} & \makecell{\textbf{Task-Targeted}\\\textbf{Artifact Correction}} & \makecell{\textbf{Supervised Fine-Tuning}\\\;} & \makecell{\textbf{Ground Truth}\\\;} \\
        \end{tabular}
    }
    \vspace{-0.05in}
    \caption{Qualitative results. Top: FasterRCNN Detection, Middle: MaskRCNN Instance Segmentation, Bottom: HRNetV2 + C1 Semantic Segmentation. All inputs were compressed at quality 10. Note the poor quality of the results with no mitigation. In particular, the detection prediction of ``boat'', while incorrect, is reasonable based on the deformed shape and lack of texture of the compressed image. Despite the compression, the network is displaying some understanding that there is an ocean scene. Similarly, for MaskRCNN, skateboard is predicted seemingly on the shape of the region alone. As mitigations are applied, the confidence of the correct detections for both detection and instance segmentation increases. For semantic segmentation, the result is mostly incorrect with seemingly random classification regions which mitigations are able to clean up significantly.}
    \vspace{-0.2in}
    \label{fig:qual}
\end{figure*}

Since our study is focused on both \textbf{analyzing} (Sections \ref{sec:res:study} and \ref{sec:res:for}) and \textbf{mitigating} (Sections \ref{sec:results:transfer}, and \ref{sec:res:baseline}), our results are organized into one of these two categories. First, we show an abridged version of the study on JPEG compression, the full study results are available in Appendix D. After this, we present a preliminary study of two recent forensics models which were tested using the same methodology as the main study. We then examine the transferability and multihead scenarios mentioned in Section \ref{sec:meth} in detail, a unique property of Task-Targeted Artifact Correction that makes it a flexible mitigation option.

\subsection{Analyzing: Abridged Study Results}
\label{sec:res:study}

We show a subset of our results highlighting interesting, unusual, or edge cases as well as discussing overall findings for each major task. We aim to provide a good covering of common tasks and datasets and picked several models and a single representative dataset per task. While our work is the most comprehensive study of its kind to date, it is not exhaustive: there exist models and datasets which we did not test on. In most cases, we started from pretrained weights available in commodity deep learning libraries.

The results presented in the body of the paper are restricted to qualities [10, 50] (heavy to moderate compression) for brevity and because these are the most interesting results. Results on the full range of qualities we considered ([10, 90]) along with tables of results are available in our appendices. In each case, the results are shown as plots giving loss in performance on a task appropriate metric \vs JPEG quality level. We studied Classification, Object Detection, Instance and Semantic Segmentation as a set of computer vision tasks.

The abbreviated results are shown in Figure \ref{fig:nomitigation}, which shows the results of all models with no mitigation separated by task, and Figure \ref{fig:withmitigation}, which shows the behavior of individual models with mitigations applied compared to no mitigation in condensed form. Please see Appendix D for full plots and tables for individual models. All performance measures show the absolute drop in performance, \eg, a 14\% drop indicates that the model performs 14\% worse on JPEG images at the given quality than on uncompressed images. For the reference numbers that we used for this comparison, see Appendix D.4. As a general rule, we observed that more complex tasks incurred a larger performance penalty on JPEG compressed inputs. Additionally, the more complex the task, the more likely it was to be aided by Task-Targeted Artifact Correction, and the less likely it was to be aided by Supervised Fine-Tuning. In Figure \ref{fig:qual}, we show qualitative results for detection, instance and semantic segmentation. Please see Appendix C to view these results in more detail. We now briefly discuss the details of the study for each task.

\vspace{-0.4cm}
\paragraph{Classification.}
We tested classification models using the ImageNet \cite{imagenet_cvpr09} dataset. We tested the following models: MobileNetV2 \cite{sandler2018mobilenetv2}, ResNet 18, 50, and 101 \cite{he2016deep}, ResNeXt 50 and 101 \cite{xie2017aggregated}, VGG 19 \cite{simonyan2014very}, InceptionV3 \cite{szegedy2016rethinking}, and EfficientNet B3 \cite{tan2019efficientnet}. The pretrained weights for the task networks in this section come from the torchvision library \cite{marcel2010torchvision}. The evaluation metric used was Top-1 Accuracy. Models in this task generally responded better to Supervised Fine-Tuning than to Task-Targeted Artifact Correction with the notable exception of MobileNetV2 and EfficientNet which responded better to Task-Targeted Artifact Correction. We ran GradCAM \cite{selvaraju2017grad} to examine Class Activation Maps for JPEG inputs as well as all mitigations. We found that JPEG degrades the gradient quality as well as induces localization errors. Please see Appendix A for this analysis.
\vspace{-0.4cm}
\paragraph{Detection and Instance Segmentation.}
Next, we show results on object detection and instance segmentation. These models were tested using the MS-COCO dataset \cite{lin2014microsoft}. We tested three detection models: Fast R-CNN \cite{girshick2015fast}, Faster R-CNN \cite{ren2016faster}, and RetinaNet \cite{lin2017focal}, and we used Mask R-CNN \cite{he2017mask} for instance segmentation. The pretrained weights come from the Detectron2 library \cite{wu2019detectron2}. In all cases, we use a model with a ResNet 50 \cite{he2016deep} backbone, and for the R-CNNs we use a Feature Pyramid Network \cite{lin2017feature} for the detector. For Task-Targeted Artifact Correction training, we use loss on only the backbone features rather than the detection logits. The evaluation metric used was mean average precision (mAP). Models in this section responded well to all mitigation techniques, with Task-Targeted Artifact Correction helping the most for low-quality settings. The moderate quality settings, however, responded better to Supervised Fine-Tuning For a further analysis of detection, we used TIDE \cite{tide-eccv2020} and determined that missed detections make up the bulk of errors caused by JPEG compression, please see Appendix B for this analysis.
\vspace{-0.5cm}
\paragraph{Semantic Segmentation.}
For semantic segmentation, we show results on ADE20K  using the accompanying code \cite{zhou2016semantic,zhou2017scene}. This code is organized into pluggable encoders and decoders. We test the following encoders: MobileNetV2 \cite{sandler2018mobilenetv2}, ResNet 18, 50, and 101 \cite{he2016deep}, and HRNet \cite{sun2019high}. We test the following decoders: C1 with deep supervision \cite{zhao2017pyramid}, PSPNet (the Pyramid Pooling Module) with and without deep supervision \cite{zhao2017pyramid}, and UPerNet \cite{xiao2018unified}. Similar to the object detection experiments, we train Task-Targeted Artifact Correction using loss from the encoder features only. The metric used is mean of per-class intersection-over-union of classified pixels (mIoU). In general, segmentation models were greatly affected by compressed inputs and Supervised Fine-Tuning was not an effective mitigation, with some models performing worse after fine tuning than with no mitigation at all. Conversely, Task-Targeted Artifact Correction was able to effectively mitigate performance loss for low and moderate qualities.

\vspace{-0.3cm}
\subsection{Analyzing: Limited Study on Forensics}
\label{sec:res:for}
\begin{figure}
    \centering
    \begin{subfigure}[t]{0.23\textwidth}
        \includegraphics[width=\textwidth]{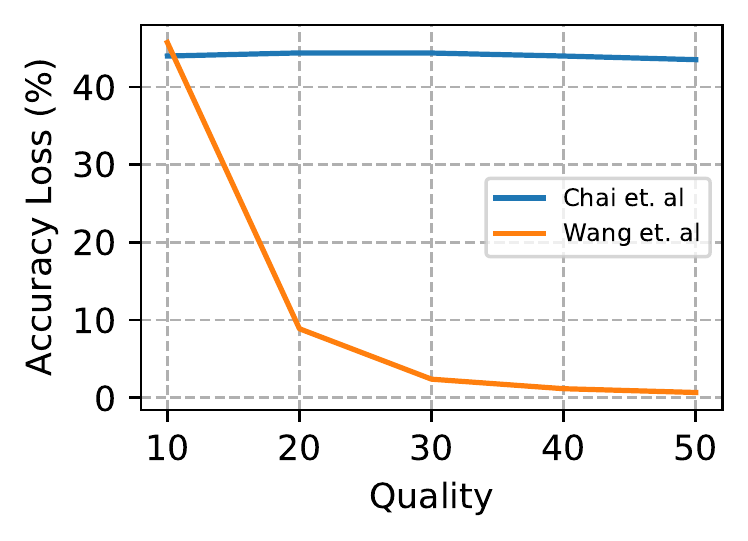}
        \caption{Forensic task performance with no mitigation.}
    \end{subfigure}
    \hfill
    \begin{subfigure}[t]{0.23\textwidth}
        \includegraphics[width=\textwidth]{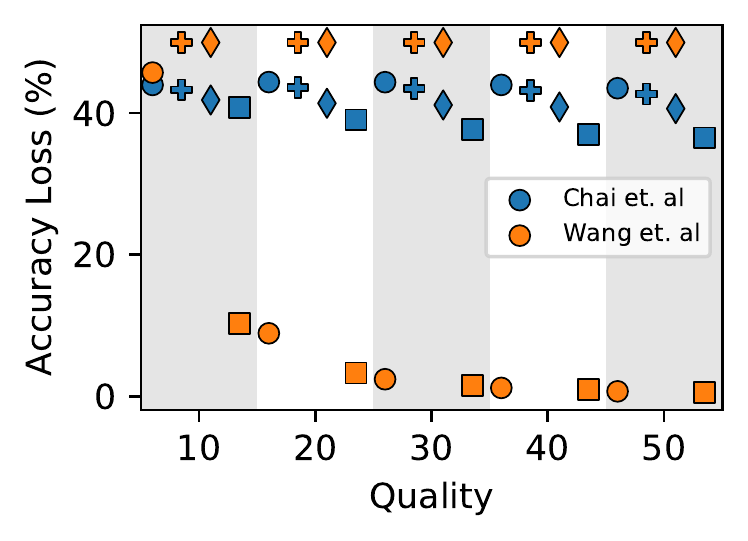}
        \caption{Forensics task performance with mitigation}
    \end{subfigure}
    \vspace{-0.1in}
    \caption{Forensic model results. \textbf{$\pmb{\bullet}$ Circle}: No Mitigation, \textbf{$\pmb{+}$ Cross}: Off-the-Shelf Artifact Correction, \textbf{$\blacklozenge$ Diamond}: Task-Targeted Artifact Correction, \textbf{$\blacksquare$ Square}: Supervised Fine-Tuning.}
    \label{fig:for}
    \vspace{-0.2in}
\end{figure}

In addition to the above computer vision tasks, we conducted a limited study on two recent forensics models: Wang \etal \cite{wang2019cnngenerated} and Chai \etal \cite{chai2020makes}. The goal of both of these models is to detect an image which, in whole or in part, was generated by a GAN \cite{goodfellow2014generative}, and can be expressed as a two-class classification problem with classes ``real'' and ``fake''. Wang \etal uses a ResNet 50 architecture \cite{he2016deep} to classify the images. Chai \etal uses a slightly modified Xception \cite{chollet2017xception} model to classify patches, forming an image label by majority vote among the patches. Both methods use their own bespoke datasets and provide pretrained models which we obtained and used for our experiments. The evaluation metric we used was intended to match the one presented in each paper. For Wang \etal that is accuracy of the real/fake prediction, while for Chai \etal it is accuracy of the predicted patches. The results are shown in Figure \ref{fig:for}. Despite the similar formulation, both networks have completely different behavior on compressed images than traditional ImageNet \cite{imagenet_cvpr09} classification and furthermore have completely different behavior to each other. We note that both models have nearly 100\% accuracy on clean images they were trained on. Under JPEG compression, even up to the quality 90 (the maximum that we tested) performance of Chai \etal stays around 50\% indicating the equivalent to a random guess while Wang \etal is able to match its performance by quality 50. While both models incorporated slight JPEG-as-data-augmentation into their pretrained weights, they are greatly aided by additional Fine-Tuning although artifact correction of any kind provided little to no benefit. Since these methods are intended to spot potentially malicious behavior, we find it troubling that a simple perturbation like JPEG is capable of fooling the networks to such a degree and we hope that this study will provide an impetus for further investigation.

\begin{figure*}
    \centering
    \begin{subfigure}[t]{0.33\textwidth}
        \includegraphics[width=\textwidth]{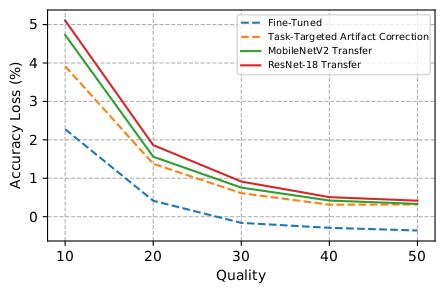}
        \caption{ResNet101 Results with Transfer.}
        \label{fig:transfer:intra}
    \end{subfigure}
    \begin{subfigure}[t]{0.33\textwidth}
        \includegraphics[width=\textwidth]{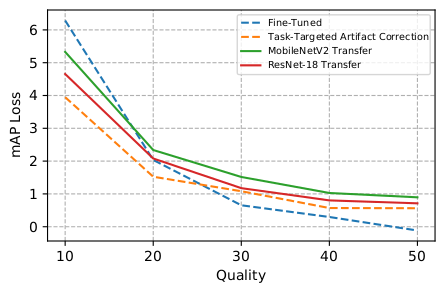}
        \caption{FasterRCNN Results with Transfer.}
        \label{fig:transfer:inter-d}
    \end{subfigure}
    \begin{subfigure}[t]{0.33\textwidth}
        \includegraphics[width=\textwidth]{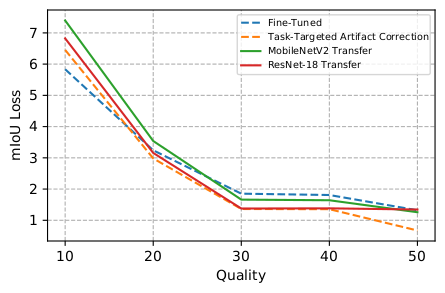}
        \caption{HRNetV2 + C1 Results with Transfer.}
        \label{fig:transfer:inter-s}
    \end{subfigure}
    \caption{Transfer Results. In all plots, we add an evaluation using artifact correction weights that were trained on ResNet-18 and MobileNetV2, our lightest weight models. Note that ``Fine-Tuned'' and ``Task-Targeted Artifact Correction'' methods are both trained using their respective task network directly \eg in (a) they use a ResNet 101. \textbf{- -} dashed lines indicate results shown in Section \ref{sec:res:study}.}
    \label{fig:transfer}
\end{figure*}

\begin{figure*}
    \centering
    \begin{subfigure}[t]{0.33\textwidth}
        \includegraphics[width=\textwidth]{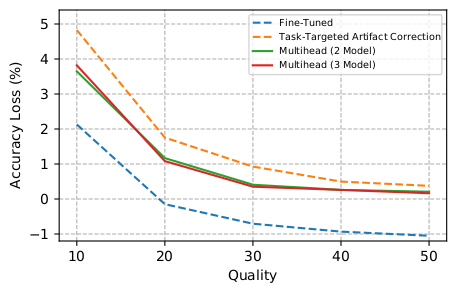}
        \caption{ResNet50 Results with Multihead.}
    \end{subfigure}
    \begin{subfigure}[t]{0.33\textwidth}
        \includegraphics[width=\textwidth]{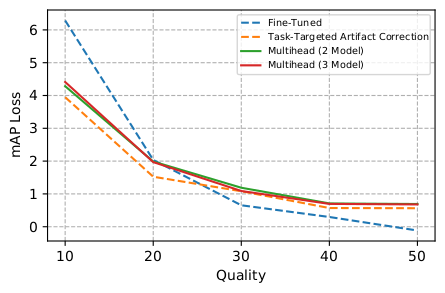}
        \caption{FasterRCNN Results with Multihead.}
    \end{subfigure}
    \begin{subfigure}[t]{0.33\textwidth}
        \includegraphics[width=\textwidth]{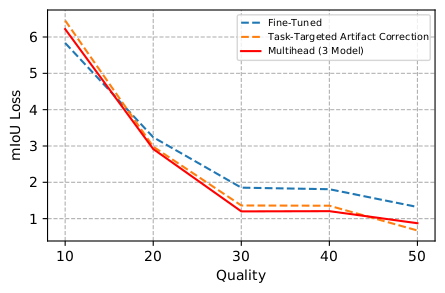}
        \caption{HRNetV2 + C1 Results with Multihead.}
    \end{subfigure}
    \caption{Multihead Results. In all plots, we add an evaluation using artifact correction weights that were trained using multiple task networks. For the two task setup, we used ResNet-50 and FasterRCNN. For the three task setup, we used ResNet-50, FasterRCNN, and HRNetV2 + C1. Note that HRNetV2 + C1 has no two-task multihead model. \textbf{- -} dashed lines indicate results shown in Section \ref{sec:res:study}.}
    \label{fig:mh}
    \vspace{-0.2in}
\end{figure*}

\subsection{Mitigating: Transferability and Multihead}
In this section, we show results which are intended to give a better understanding of our proposed Task-Targeted Artifact Correction. In particular, we examine the transferability of the targeted models, \eg, if a model which was trained for one network and task, like a lightweight MobileNetV2 for ImageNet classification, for example, can be used to mitigate performance loss of a more complex and harder to train network like a ResNet-101, or even used in a wholly different task like detection. We also examine if this performance mitigation holds when a correction network is targeted to multiple models simultaneously, in other words, when multiple task networks are providing supervisory signal. These scenarios improve the practical application of the method by allowing for more flexible training and model reuse in addition to the stated advantage of not requiring labels. A table of results for all plots in this section is given in Appendix F.

\label{sec:results:transfer}

\begin{table}[t]
    \centering
    \footnotesize
    \renewcommand{\arraystretch}{1.2}
    \renewcommand{\tabcolsep}{2.4mm}
    \caption{Comparison with Common AC Baselines for Task-Targeted Artifact Correction. ARCNN and IDCN are both ``quality aware'' models. Metrics are classification: top-1 accuracy, detection: mAP, segmentation: mIoU. QGAC outperforms both.}
    \label{tab:baseline}
    \vspace{-0.05in}
    \resizebox{0.47\textwidth}{!}{
        \setlength{\cmidrulewidth}{0.01em}
\begin{tabular}{@{}llrr@{}}
    \toprule
    Task (Network)                                        & AC Network & Quality 10     & Quality 20     \\
    \midrule
    \multirow{3}{*}{MobileNetV2 (Classification)}         & ARCNN      & 61.19          & 66.89          \\
                                                          & IDCN       & 62.62          & 65.72          \\
                                                          & QGAC       & \textbf{64.64} & \textbf{68.63} \\
    \midrule
    \multirow{3}{*}{FasterRCNN (Detection)}               & ARCNN      & 24.99          & 27.02          \\
                                                          & IDCN       & 25.99          & 28.23          \\
                                                          & QGAC       & \textbf{31.43} & \textbf{33.85} \\
    \midrule
    \multirow{3}{*}{HRNetV2 + C1 (Semantic Segmentation)} & ARCNN      & 29.35          & 35.90          \\
                                                          & IDCN       & 32.62          & 37.00          \\
                                                          & QGAC       & \textbf{34.14} & \textbf{37.61} \\
    \bottomrule
\end{tabular}
    }
    \vspace{-0.2in}
\end{table}
\vspace{-0.4cm}
\paragraph{Intra-Task Transferability.}
For intra-task transferability, we tested two scenarios. We first trained Task-Targeted Artifact Correction models on ResNet 18 \cite{he2016deep} and MobileNet V2 \cite{sandler2018mobilenetv2}, two of our smallest models which are fast to train. We then tested these models on ImageNet \cite{imagenet_cvpr09} classification using ResNet 101 \cite{he2016deep} as the downstream network, one of our largest and slowest to train models. The result shown in Figure \ref{fig:transfer:intra} is comparable with the task-targeted model trained on the ResNet 101 itself, indicating good transfer.
\vspace{-0.4cm}
\paragraph{Inter-Task Transferability.}
For inter-task transferability, we used the same artifact correction models trained in the previous section but now used them to test using COCO \cite{lin2014microsoft} object detection on Faster R-CNN \cite{ren2015faster} and ADE20K \cite{zhou2016semantic,zhou2017scene} semantic segmentation using the HRNetV2 \cite{sun2019high} encoder with C1 decoder. These results are shown in Figures \ref{fig:transfer:inter-d} and \ref{fig:transfer:inter-s}, and again show good transfer with comparable performance to task-targeted networks trained for each of the networks.
\vspace{-0.4cm}
\paragraph{Multihead.}
We tested two setups for multihead training: a two task setup and a three task setup. For the two task setup, we train the artifact correction network using a ResNet 18 \cite{he2016deep} as well as a Faster R-CNN \cite{ren2015faster} providing downstream loss. For the three task setup we add in semantic segmentation using HRNetV2 \cite{sun2019high} + C1. To train this, we alternate batches, taking one batch from ImageNet, performing a full forward pass using the Resnet 18, then taking a batch from COCO and performing another full forward pass using the Faster R-CNN backbone and in three task case following the same procedure with ADE20K. The $l_1$ loss of the features for each separate network is taken and the backpropagated loss is summed. The result, shown in Figure \ref{fig:mh} is on-par with artifact correction models trained with a single model, and in the classification case, both multi-head models perform better than the single task corrector indicating better generalization.

\subsection{Mitigating: Limited Comparison with Other Artifact Correction Models}
\label{sec:res:baseline}

As discussed in Section \ref{sec:meth}, artifact-correction based mitigation has only recently become viable using quality-blind methods. Using quality-aware methods, which train a unique model for each quality setting, is impossible in real scenarios because the quality setting is not stored in the JPEG file, and in the context of the study, leads to an intractable training protocol which would require models to be trained per-quality level and per-model. Nevertheless, since quality aware models dominated the literature for many years, there is some interest in their behavior in this context. To make this tractable we restrict this part of the study to two models: ARCNN \cite{dong2015compression}, a standard baseline in artifact correction, and IDCN \cite{zheng2019implicit} a recent model. ARCNN was modified to handle color images and IDCN handles them natively. Both models were ported to our framework and retrained to within 0.5dB PSNR of the published numbers. We additionally restrict the quality settings to 10 and 20 only (the quality settings reported by IDCN). Table \ref{tab:baseline} shows that both models perform worse than QGAC, further motivating our use of it for the main study.

\vspace{-0.4cm}
\section{Conclusion}
\vspace{-0.2cm}
In this paper we conducted a large scale study of JPEG compression on common computer vision tasks and datasets. Our study shows that JPEG compression has a steep penalty across the board for heavy to moderate compression settings. We also tested several strategies for mitigating this performance penalty including a novel method which requires no labels. Our proposed mitigation strategy achieves better results than other unlabeled mitigations and since it requires no labels it is ideal for consumer facing applications where labels are often hard to obtain. For complex tasks, our method outperforms the supervised method despite having no access to ground truth labels. Our method promotes model reuse by allowing transfer of weights between tasks and multihead training. For further results, including throughput for each tested model (Appendix E) we invite readers to our appendices.

We hope to extend this work by considering more compression methods. Despite their relative disuse, there are other still image compression techniques, such as JPEG 2000 \cite{marcellin2000jpeg}, HEIF \cite{hannuksela2015overview}, WebP \cite{webp2020new}, and BPG \cite{bpg} that should be considered. Consideration of learned-image-compression algorithms is also a useful addition. Finally, video processing models are becoming commonplace, and video compression is almost exclusively lossy. An extended study should consider these models over different video compression settings.

It is our hope that this will be a living study that it will be updated as new models, compression algorithms, and mitigation techniques are developed. Our benchmarking code is pluggable to allow for experimentation with different models, and will be made freely available. Futher, since our Task-Targeted Artifact Correction method shows good transferability, we plan to make the weights we trained for this study available for general use. Many applications will benefit from using our TTAC weights with no modification.

\medskip
\noindent\textbf{Acknowledgement} \medspace This project was partially supported by independent grants from Facebook AI, DARPA SemaFor (HR001119S0085) and DARPA SAIL-ON (W911NF2020009) programs.

\clearpage
{\small
    \printbibliography
}

\clearpage

\onecolumn

\appendix

\section{GradCam Visualizations}

\begin{figure}[H]
    \begin{subfigure}{0.32\textwidth}
        \includegraphics[width=\textwidth]{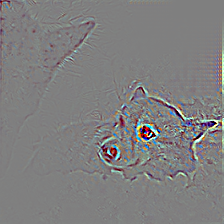}
        \caption{Fine tuned Model Gradient}
    \end{subfigure}
    \hfill
    \begin{subfigure}{0.32\textwidth}
        \includegraphics[width=\textwidth]{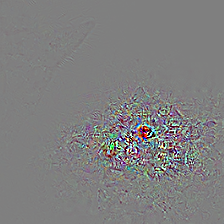}
        \caption{Original Model Gradient with JPEG Input}
    \end{subfigure}
    \hfill
    \begin{subfigure}{0.32\textwidth}
        \includegraphics[width=\textwidth]{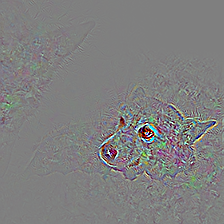}
        \caption{Original Model Gradient with Original Input}
    \end{subfigure}
    \begin{subfigure}{0.32\textwidth}
        \includegraphics[width=\textwidth]{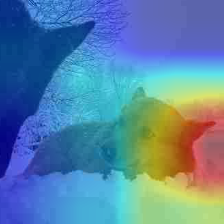}
        \caption{Fine tuned Model CAM}
    \end{subfigure}
    \hfill
    \begin{subfigure}{0.32\textwidth}
        \includegraphics[width=\textwidth]{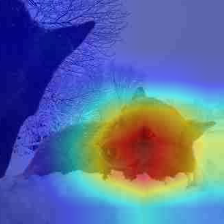}
        \caption{Original Model CAM with JPEG Input}
    \end{subfigure}
    \hfill
    \begin{subfigure}{0.32\textwidth}
        \includegraphics[width=\textwidth]{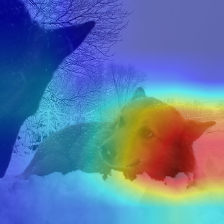}
        \caption{Original Model CAM with Original Input}
    \end{subfigure}
    \caption{Fine Tuned Model Comparison}
    \label{fig:ftmc}
\end{figure}

\begin{figure}[H]
    \begin{subfigure}{0.32\textwidth}
        \includegraphics[width=\textwidth]{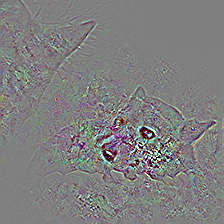}
        \caption{Off-the-Shelf AC Gradient}
    \end{subfigure}
    \hfill
    \begin{subfigure}{0.32\textwidth}
        \includegraphics[width=\textwidth]{figures/gradcam/jpeg/cam_gb.png}
        \caption{Original Model Gradient with JPEG Input}
    \end{subfigure}
    \hfill
    \begin{subfigure}{0.32\textwidth}
        \includegraphics[width=\textwidth]{figures/gradcam/original/cam_gb.png}
        \caption{Original Model Gradient with Original Input}
    \end{subfigure}
    \begin{subfigure}{0.32\textwidth}
        \includegraphics[width=\textwidth]{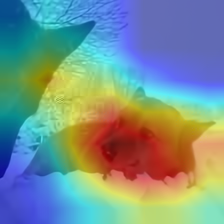}
        \caption{Off-the-Shelf AC CAM}
    \end{subfigure}
    \hfill
    \begin{subfigure}{0.32\textwidth}
        \includegraphics[width=\textwidth]{figures/gradcam/jpeg/cam.png}
        \caption{Original Model CAM with JPEG Input}
    \end{subfigure}
    \hfill
    \begin{subfigure}{0.32\textwidth}
        \includegraphics[width=\textwidth]{figures/gradcam/original/cam.png}
        \caption{Original Model CAM with Original Input}
    \end{subfigure}
    \caption{Off-the-Shelf Artifact Correction Comparison}
    \label{fig:otsc}
\end{figure}

\begin{figure}[H]
    \begin{subfigure}{0.32\textwidth}
        \includegraphics[width=\textwidth]{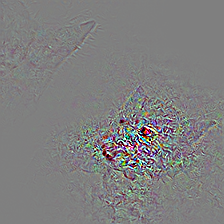}
        \caption{Task-Targeted AC Gradient}
    \end{subfigure}
    \hfill
    \begin{subfigure}{0.32\textwidth}
        \includegraphics[width=\textwidth]{figures/gradcam/jpeg/cam_gb.png}
        \caption{Original Model Gradient with JPEG Input}
    \end{subfigure}
    \hfill
    \begin{subfigure}{0.32\textwidth}
        \includegraphics[width=\textwidth]{figures/gradcam/original/cam_gb.png}
        \caption{Original Model Gradient with Original Input}
    \end{subfigure}
    \begin{subfigure}{0.32\textwidth}
        \includegraphics[width=\textwidth]{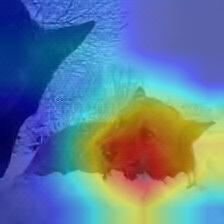}
        \caption{Task-Targeted AC CAM}
    \end{subfigure}
    \hfill
    \begin{subfigure}{0.32\textwidth}
        \includegraphics[width=\textwidth]{figures/gradcam/jpeg/cam.png}
        \caption{Original Model CAM with JPEG Input}
    \end{subfigure}
    \hfill
    \begin{subfigure}{0.32\textwidth}
        \includegraphics[width=\textwidth]{figures/gradcam/original/cam.png}
        \caption{Original Model CAM with Original Input}
    \end{subfigure}
    \caption{Task-Targeted Artifact Correction Comparison}
    \label{fig:ttacc}
\end{figure}

\begin{figure}[H]
    \begin{subfigure}[t]{0.47\textwidth}
        \centering
        \includegraphics[width=\textwidth]{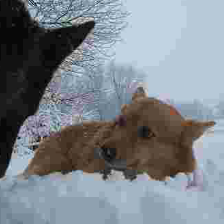}
        \caption{JPEG Q=10, Prediction: ``Norwich terrier'', Fine-Tuned Prediction: ``Pembroke, Pembroke Welsh corgi''}
    \end{subfigure}
    \hfill
    \begin{subfigure}[t]{0.47\textwidth}
        \centering
        \includegraphics[width=\textwidth]{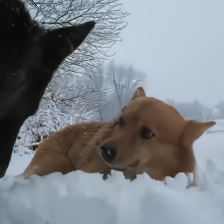}
        \caption{Off-the-Shelf Artifact Correction, Prediction: ``basenji''}
    \end{subfigure}
    \begin{subfigure}[t]{0.47\textwidth}
        \centering
        \includegraphics[width=\textwidth]{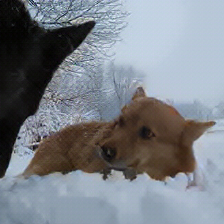}
        \caption{Task-Targeted Artifact Correction, Prediction: ``Pembroke, Pembroke Welsh corgi''}
    \end{subfigure}
    \hfill
    \begin{subfigure}[t]{0.47\textwidth}
        \centering
        \includegraphics[width=\textwidth]{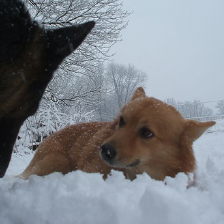}
        \caption{Original, Prediction: ``Pembroke, Pembroke Welsh corgi''}
    \end{subfigure}
    \caption{MobileNetV2, Ground Truth: ``Pembroke, Pembroke Welsh corgi''}
    \label{fig:mnv2_qual}
\end{figure}

In order to provide more insight into the potential cause of misclassifications on JPEG images, besides the obvious answer of ``quality degredation'', we provide visualizations using GradCam \cite{selvaraju2017grad}. The visualizations are shown in Figures \ref{fig:ftmc} - \ref{fig:ttacc}, see Figure \ref{fig:mnv2_qual} for enlarged inputs as well as predicted and ground-truth classes (the fine tuned model also predicts the correct class on this image). The visualizations compare the gradient and class-activation-maps (CAMs) for each mitigation technique to passing the JPEG directly (``no mitigation'') and to passing the original image directly. What we see is quite telling. The gradient with no mitigation is degraded significantly with respect to the original input, however the CAM indicates that it is still focusing in the correct location in the image. Fine-tuning the model greatly improves the quality of the gradient, however the CAM localization is now off. Off-the-shelf artifact correction improves the gradient quality, however the localization is now less constrained and the network appears unsure of where in the image to focus. Finally, Task-Targeted artifact correction seems to make an improvement in the gradient while preserving the CAM localization.

\clearpage

\section{Detection Errors}

Here we look deeper at the detection errors produced by JPEG compressed inputs using TIDE \cite{tide-eccv2020}. TIDE computes a breakdown of exactly which errors contributed to mAP loss during evaluation of detection and instance segmentation and shows the breakdown graphically in a condensed yet informative format. We ran TIDE evaluation on FasterRCNN for box detection and MaskRCNN for instance segmentation with no mitigations applied to understand how JPEG effects specific detection errors.

The results show similar behavior for both methods. On low quality JPEGs, the bulk of the errors are missed detections. This can be seen in the pie chart showing the relative proportions of missed detections, which is roughly 50\% for quality 10, and in the high number of false negatives in the bar chart on the lower right. As the quality increases, the proportion of missed detections gradually decreases and at high quality, localization errors make up a larger proportion of the errors. It should be noted that although the proportion of error attributed to localization increases, the detections overall are much more reliable on high quality JPEGs as expected. This can be seen in the significantly lower false negative rate as well as the scale of the x-axis of the bar chart in the bottom left of the images.

\begin{figure}[H]
    \begin{subfigure}{0.28\textwidth}
        \includegraphics[width=\textwidth]{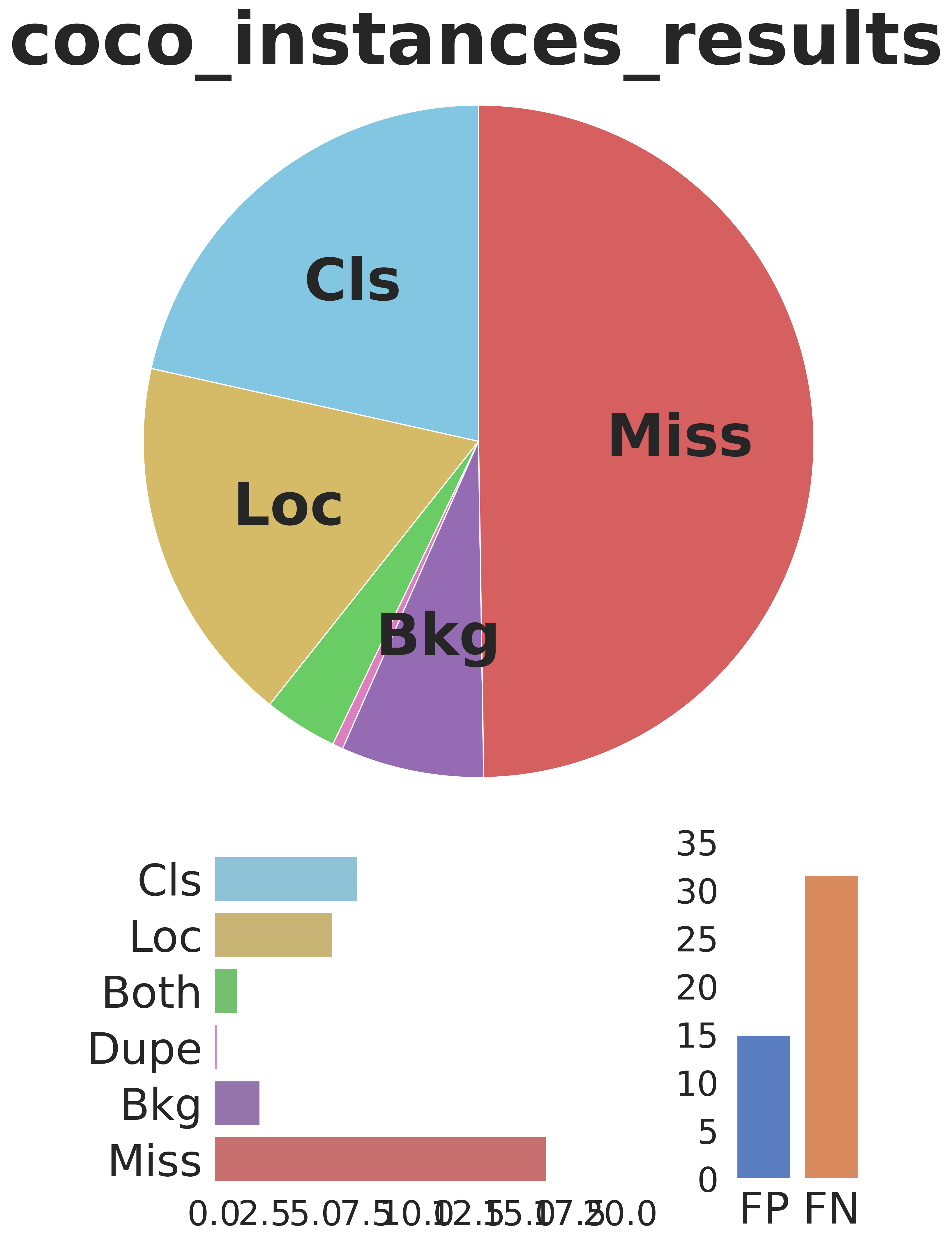}
        \caption{Q=10}
    \end{subfigure}
    \hfill
    \begin{subfigure}{0.28\textwidth}
        \includegraphics[width=\textwidth]{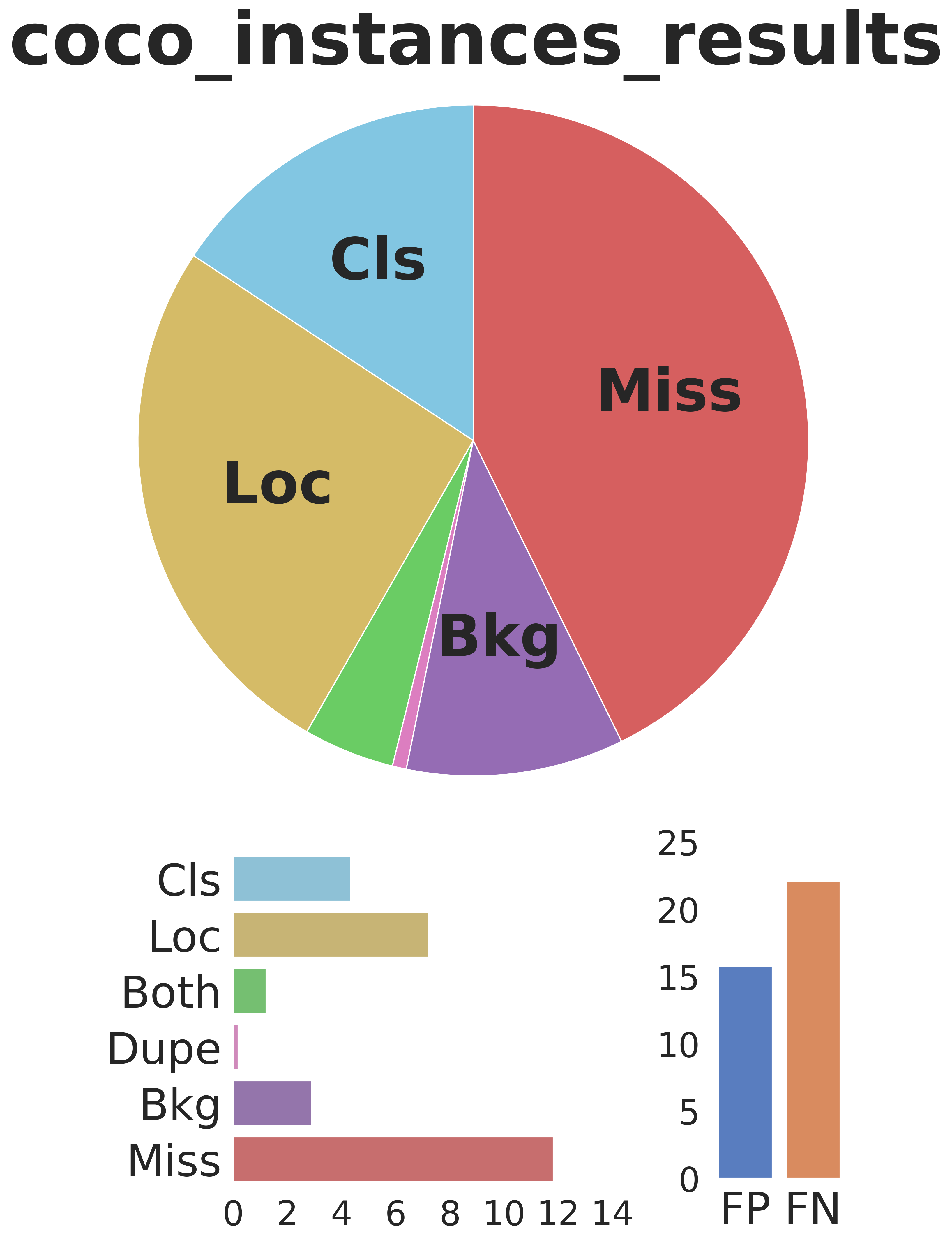}
        \caption{Q=50}
    \end{subfigure}
    \hfill
    \begin{subfigure}{0.28\textwidth}
        \includegraphics[width=\textwidth]{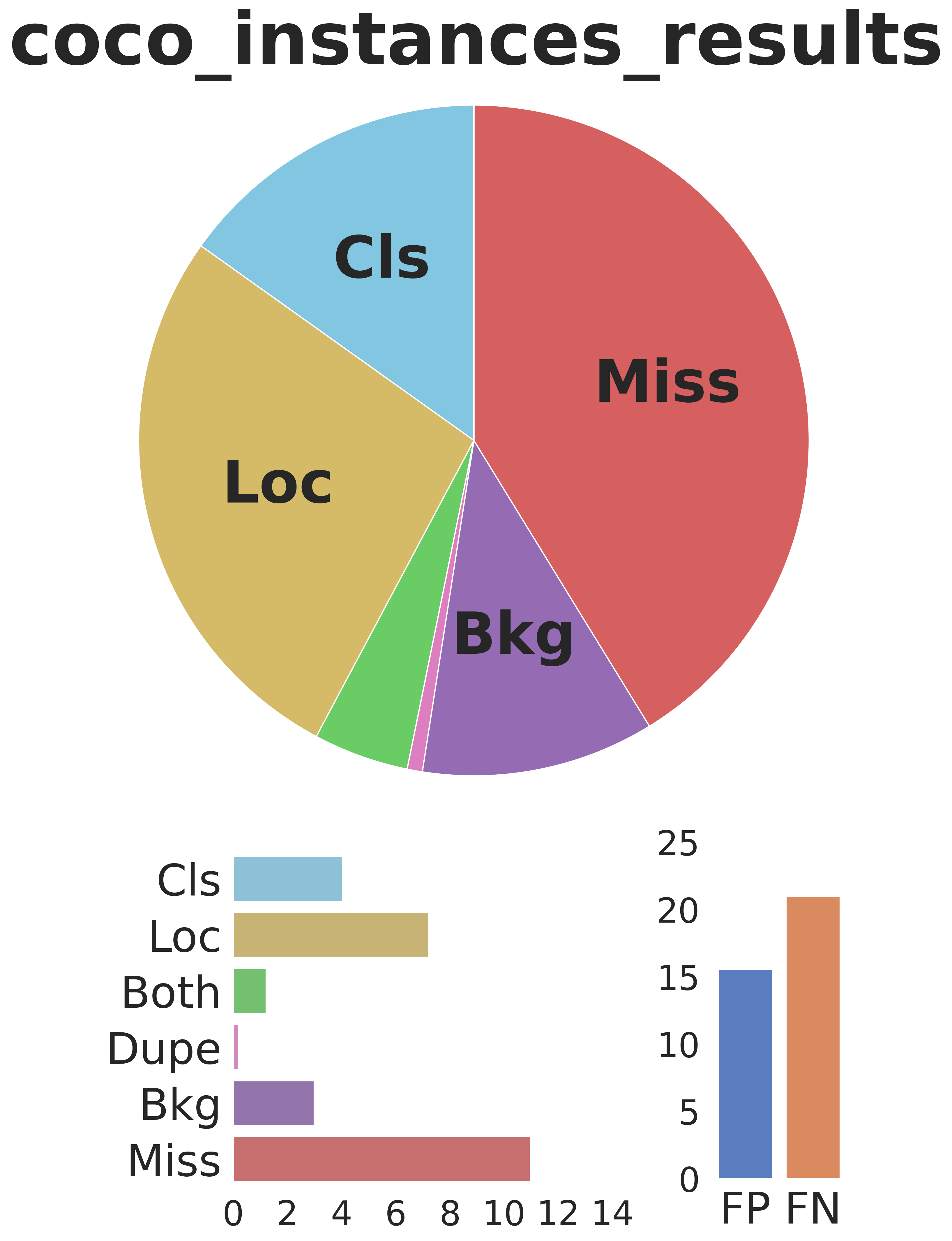}
        \caption{Uncompressed}
    \end{subfigure}
    \caption{FasterRCNN TIDE Plots}
\end{figure}

\begin{figure}[H]
    \begin{subfigure}{0.28\textwidth}
        \includegraphics[width=\textwidth]{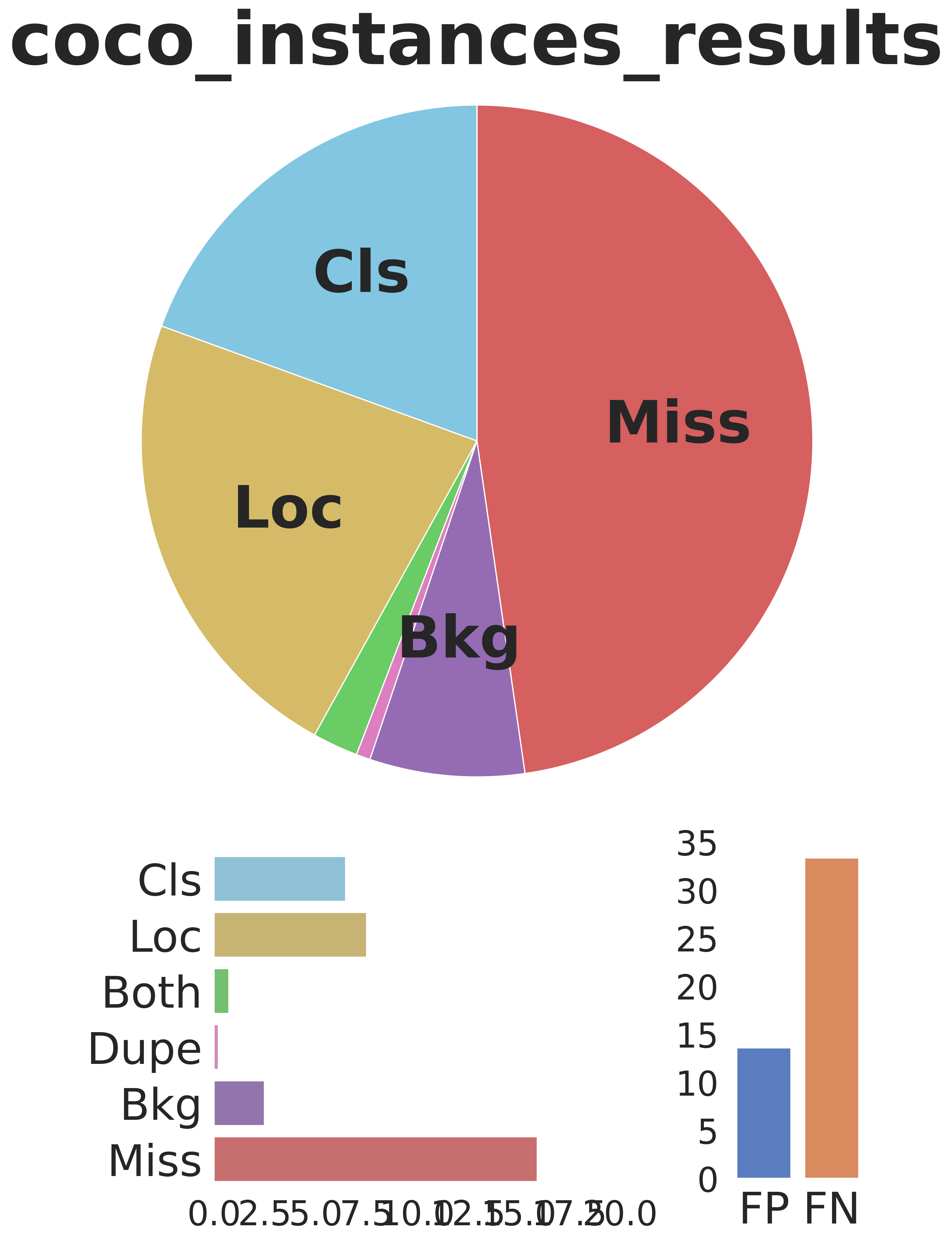}
        \caption{Q=10}
    \end{subfigure}
    \hfill
    \begin{subfigure}{0.28\textwidth}
        \includegraphics[width=\textwidth]{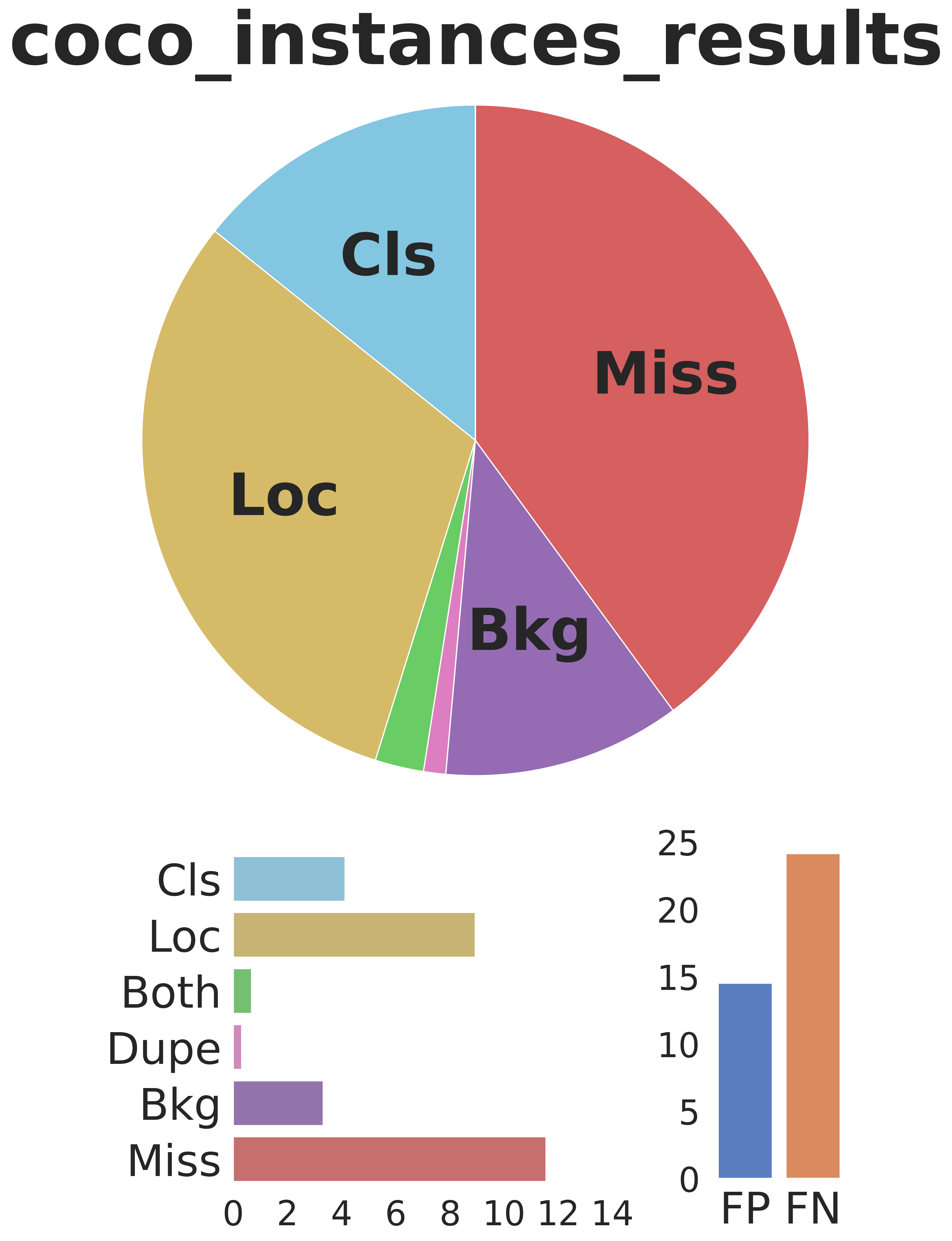}
        \caption{Q=50}
    \end{subfigure}
    \hfill
    \begin{subfigure}{0.28\textwidth}
        \includegraphics[width=\textwidth]{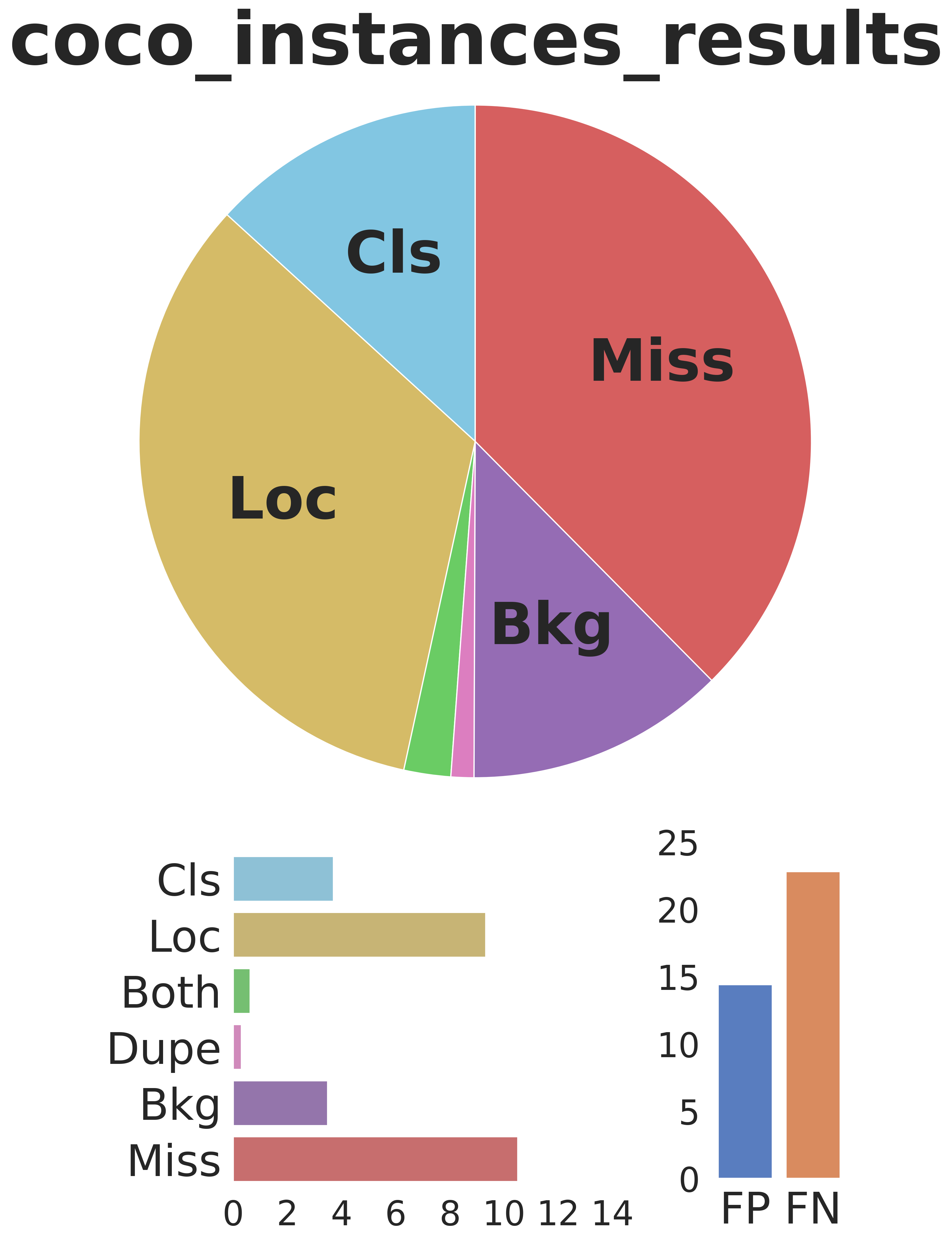}
        \caption{Uncompressed}
    \end{subfigure}
    \caption{MaskRCNN TIDE Plots}
\end{figure}

\section{Qualitative Results}

Since the proposed Task-Targeted Artifact Correction is at its core an image-to-image regression technique, we provide some qualitative results here that show images with their downstream task network behavior. All of the images in this section were compressed at quality 10 before being corrected. Where appropriate we also visualize the result of the Supervised Fine-Tuning method for comparison.

\begin{figure}[H]
    \begin{subfigure}{0.47\textwidth}
        \centering
        \includegraphics[width=\textwidth]{figures/qualitative/fasterrcnn-detection/degraded.png}
        \caption{JPEG Q=10}
    \end{subfigure}
    \hfill
    \begin{subfigure}{0.47\textwidth}
        \centering
        \includegraphics[width=\textwidth]{figures/qualitative/fasterrcnn-detection/corrected.png}
        \caption{Off-the-Shelf Artifact Correction}
    \end{subfigure}

    \begin{subfigure}{0.47\textwidth}
        \centering
        \includegraphics[width=\textwidth]{figures/qualitative/fasterrcnn-detection/tt.png}
        \caption{Task-Targeted Artifact Correction}
    \end{subfigure}
    \hfill
    \begin{subfigure}{0.47\textwidth}
        \centering
        \includegraphics[width=\textwidth]{figures/qualitative/fasterrcnn-detection/ft.png}
        \caption{Supervised Fine-Tuning}
    \end{subfigure}
    \begin{subfigure}{0.47\textwidth}
        \centering
        \includegraphics[width=\textwidth]{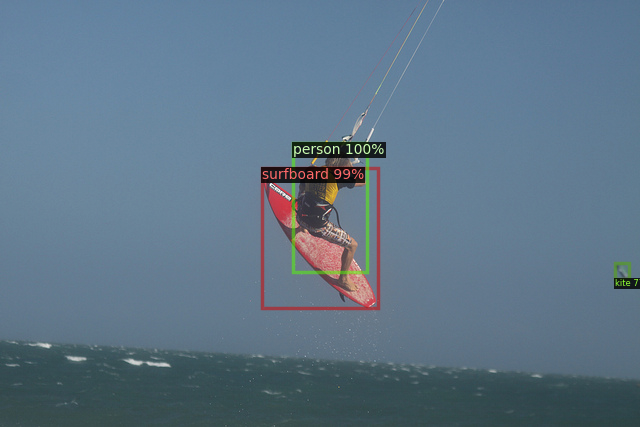}
        \caption{Original}
    \end{subfigure}
    \hfill
    \begin{subfigure}{0.47\textwidth}
        \centering
        \includegraphics[width=\textwidth]{figures/qualitative/fasterrcnn-detection/gt.png}
        \caption{Ground Truth}
    \end{subfigure}
    \caption{FasterRCNN}
\end{figure}

\begin{figure}[H]
    \begin{subfigure}{0.47\textwidth}
        \centering
        \includegraphics[width=\textwidth]{figures/qualitative/maskrccn_segmentation/jpeg.png}
        \caption{JPEG Q=10}
    \end{subfigure}
    \hfill
    \begin{subfigure}{0.47\textwidth}
        \centering
        \includegraphics[width=\textwidth]{figures/qualitative/maskrccn_segmentation/corrected.png}
        \caption{Off-the-Shelf Artifact Correction}
    \end{subfigure}

    \begin{subfigure}{0.47\textwidth}
        \centering
        \includegraphics[width=\textwidth]{figures/qualitative/maskrccn_segmentation/tt.png}
        \caption{Task-Targeted Artifact Correction}
    \end{subfigure}
    \hfill
    \begin{subfigure}{0.47\textwidth}
        \centering
        \includegraphics[width=\textwidth]{figures/qualitative/maskrccn_segmentation/ft.png}
        \caption{Supervised Fine-Tuning}
    \end{subfigure}
    \begin{subfigure}{0.47\textwidth}
        \centering
        \includegraphics[width=\textwidth]{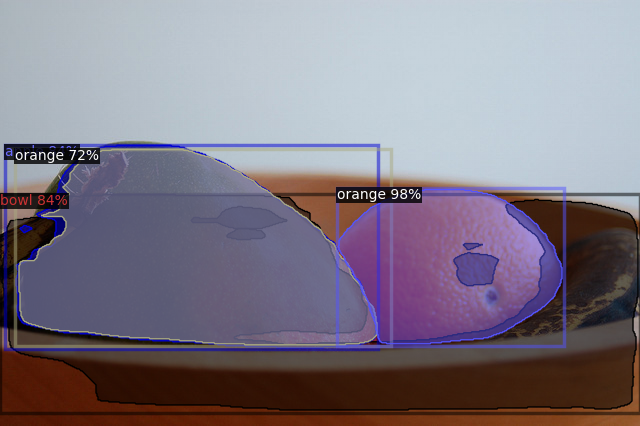}
        \caption{Original}
    \end{subfigure}
    \hfill
    \begin{subfigure}{0.47\textwidth}
        \centering
        \includegraphics[width=\textwidth]{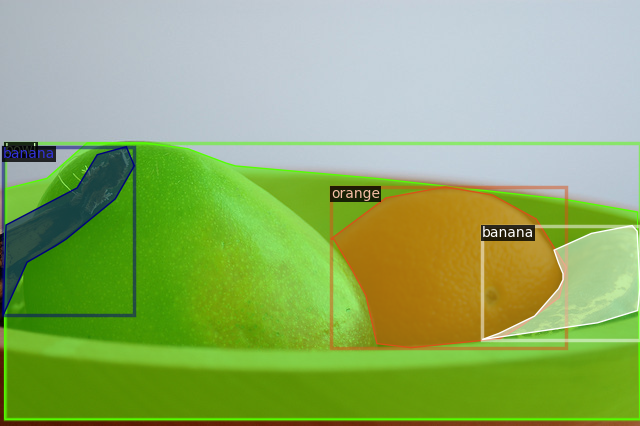}
        \caption{Ground Truth}
    \end{subfigure}
    \captionof{figure}{MaskRCNN}
\end{figure}
\begin{figure}[H]
    \begin{subfigure}{0.27\textwidth}
        \centering
        \includegraphics[width=\textwidth]{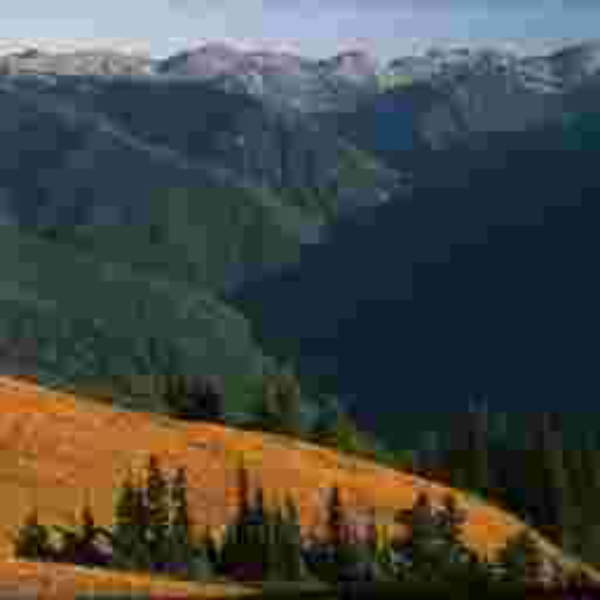}
        \caption{JPEG Q=10}
    \end{subfigure}
    \hfill
    \begin{subfigure}{0.27\textwidth}
        \centering
        \includegraphics[width=\textwidth]{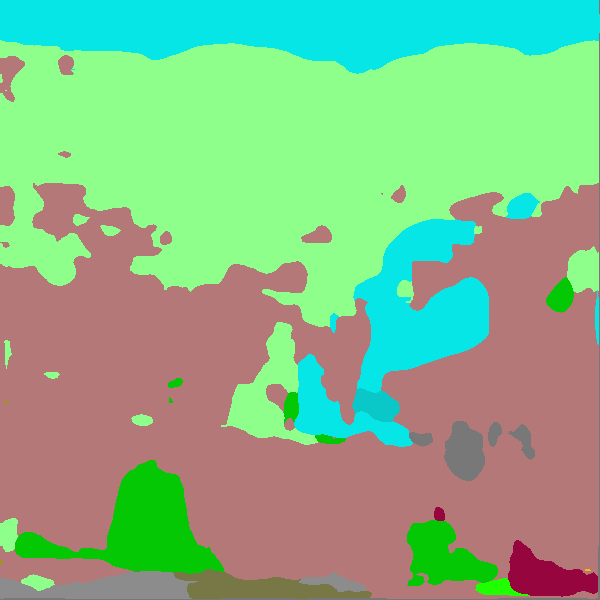}
        \caption{Degraded Prediction}
    \end{subfigure}
    \hfill
    \begin{subfigure}{0.27\textwidth}
        \centering
        \includegraphics[width=\textwidth]{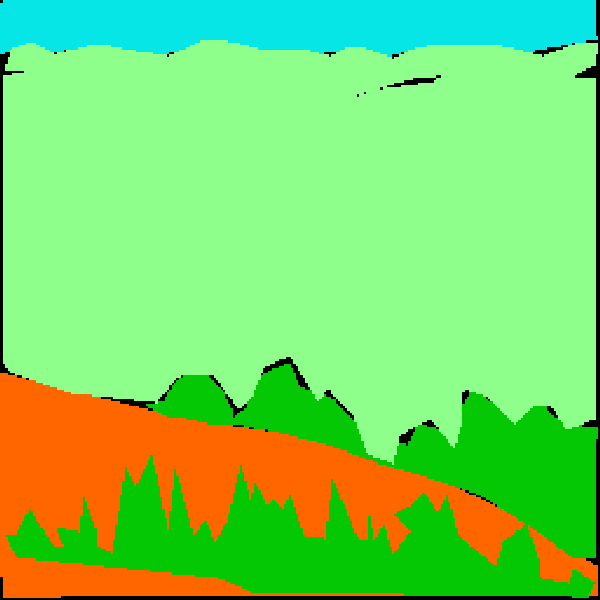}
        \caption{Ground Truth}
    \end{subfigure}

    \begin{subfigure}{0.27\textwidth}
        \centering
        \includegraphics[width=\textwidth]{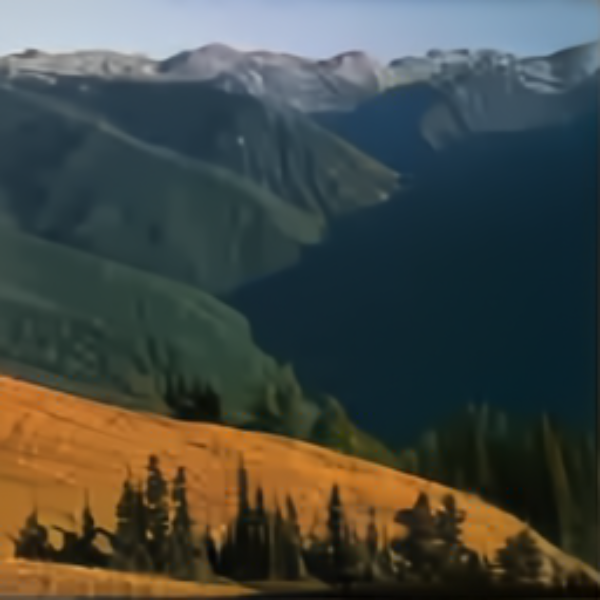}
        \caption{Off-the-Shelf Artifact Correction}
    \end{subfigure}
    \hfill
    \begin{subfigure}{0.27\textwidth}
        \centering
        \includegraphics[width=\textwidth]{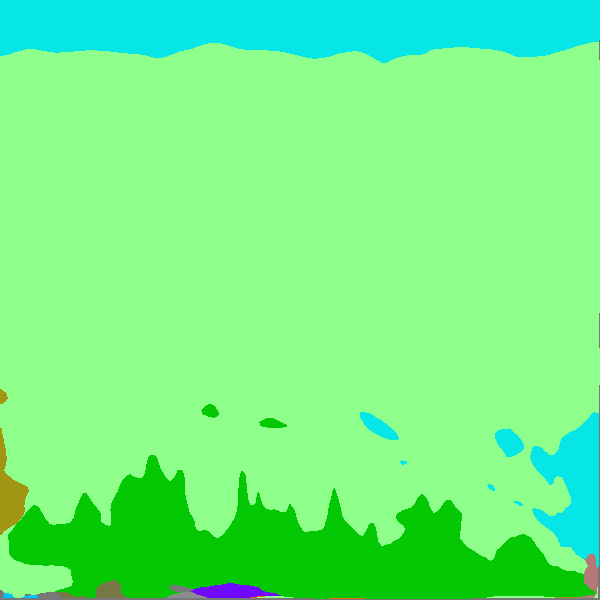}
        \caption{Off-the-Shelf Artifact Correction Prediction}
    \end{subfigure}
    \hfill
    \begin{subfigure}{0.27\textwidth}
        \centering
        \includegraphics[width=\textwidth]{figures/qualitative/hrnet_segmentation/gt.png}
        \caption{Ground Truth}
    \end{subfigure}

    \begin{subfigure}{0.27\textwidth}
        \centering
        \includegraphics[width=\textwidth]{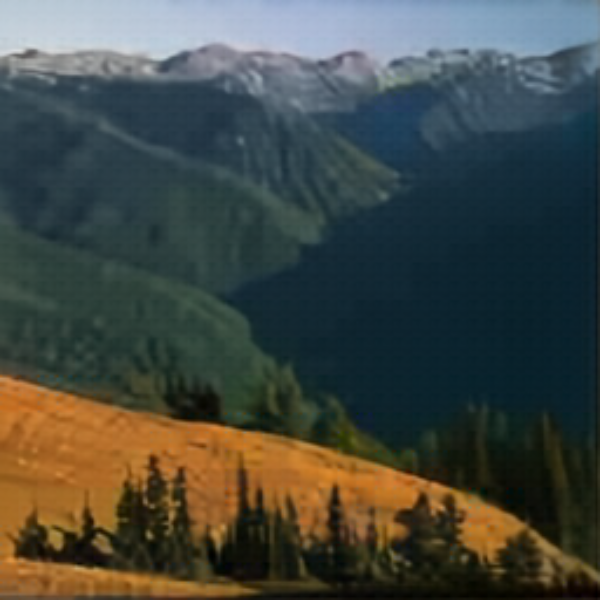}
        \caption{Task-Targeted Artifact Correction}
    \end{subfigure}
    \hfill
    \begin{subfigure}{0.27\textwidth}
        \centering
        \includegraphics[width=\textwidth]{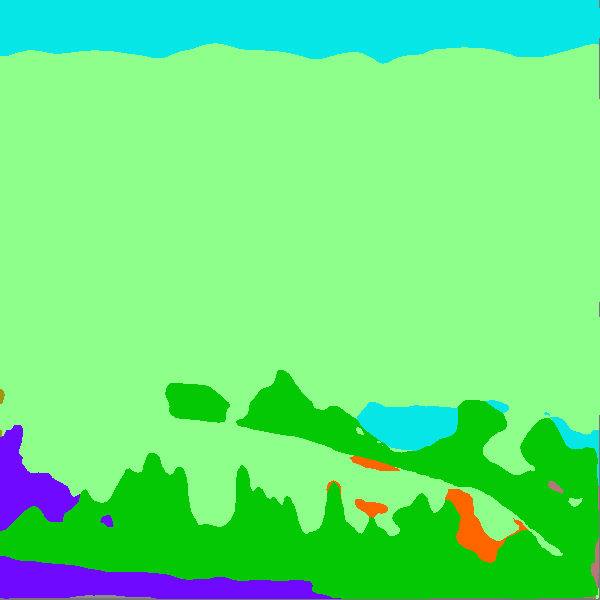}
        \caption{Task-Targeted Artifact Correction Prediction}
    \end{subfigure}
    \hfill
    \begin{subfigure}{0.27\textwidth}
        \centering
        \includegraphics[width=\textwidth]{figures/qualitative/hrnet_segmentation/gt.png}
        \caption{Ground Truth}
    \end{subfigure}

    \begin{subfigure}{0.27\textwidth}
        \centering
        \includegraphics[width=\textwidth]{figures/qualitative/hrnet_segmentation/degraded.png}
        \caption{JPEG Q=10}
    \end{subfigure}
    \hfill
    \begin{subfigure}{0.27\textwidth}
        \centering
        \includegraphics[width=\textwidth]{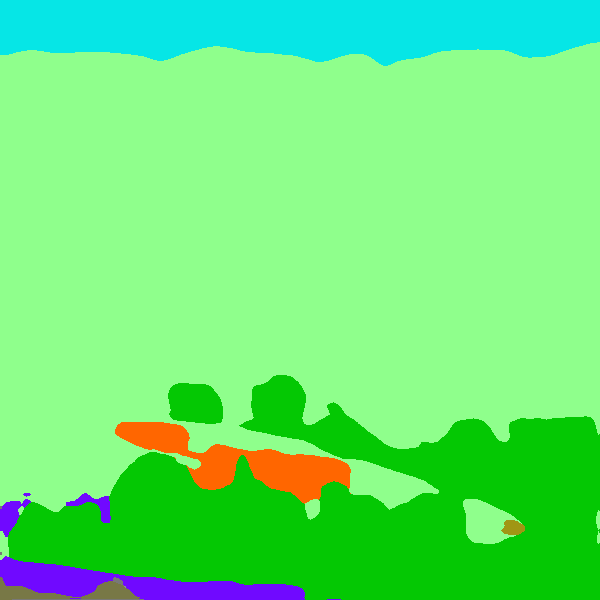}
        \caption{Supervised Fine-Tuning Prediction}
    \end{subfigure}
    \hfill
    \begin{subfigure}{0.27\textwidth}
        \centering
        \includegraphics[width=\textwidth]{figures/qualitative/hrnet_segmentation/gt.png}
        \caption{Ground Truth}
    \end{subfigure}
    \caption{HRNetV2 + C1}
\end{figure}

\section{Full Study Results}

Here we give the full results of the study including plots and tables of results for JPEG quality levels [10, 90]. The results are shown visually in plots similar to those given in the body of the paper and the raw numbers are provided in tables.

\subsection{Plots of Results}

\begin{figure}[H]
    \centering
    \includegraphics[width=\textwidth]{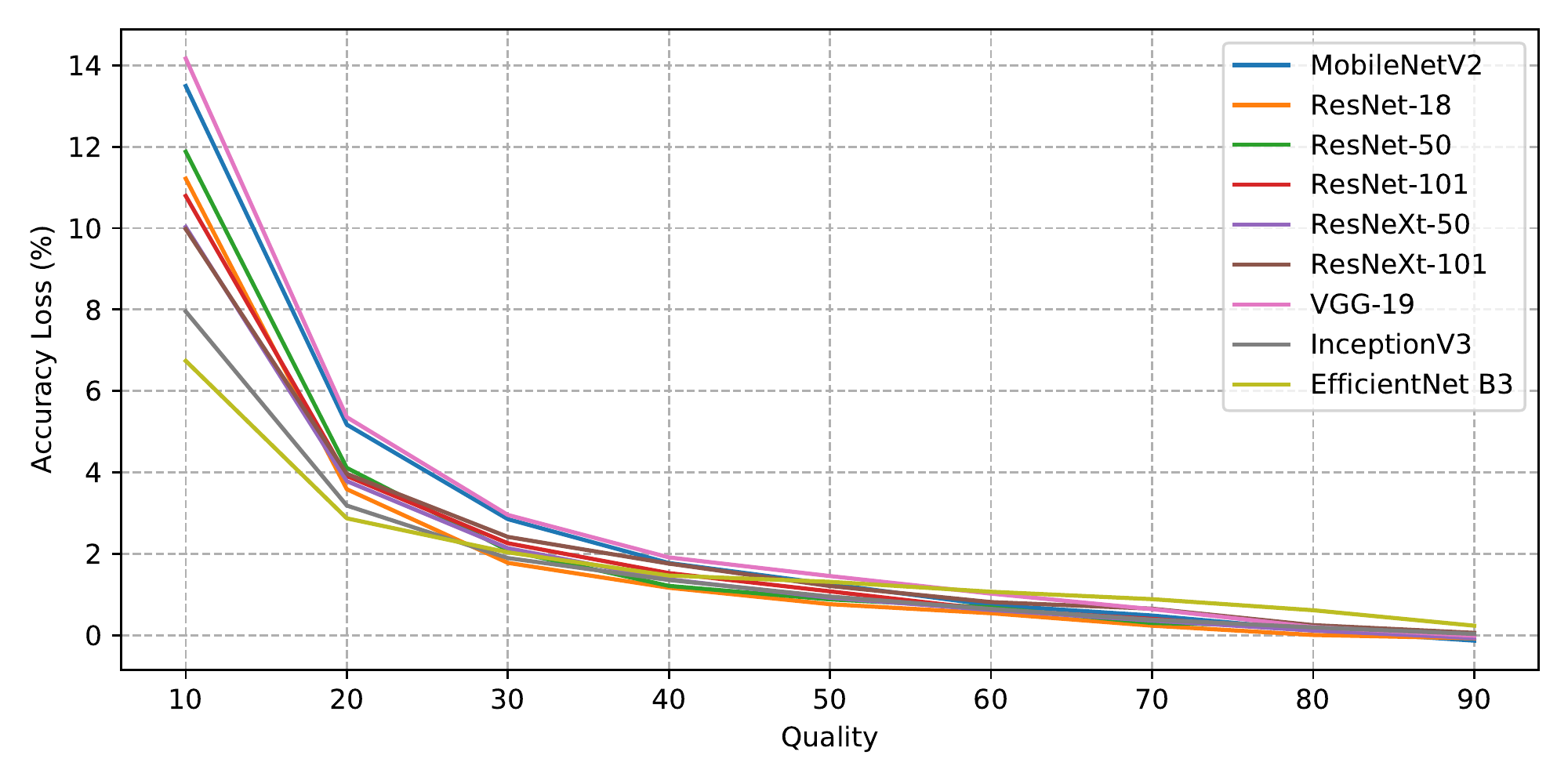}
    \caption{Classification}
\end{figure}

\begin{figure}[H]
    \begin{subfigure}[t]{0.48\textwidth}
        \centering
        \includegraphics[width=\textwidth]{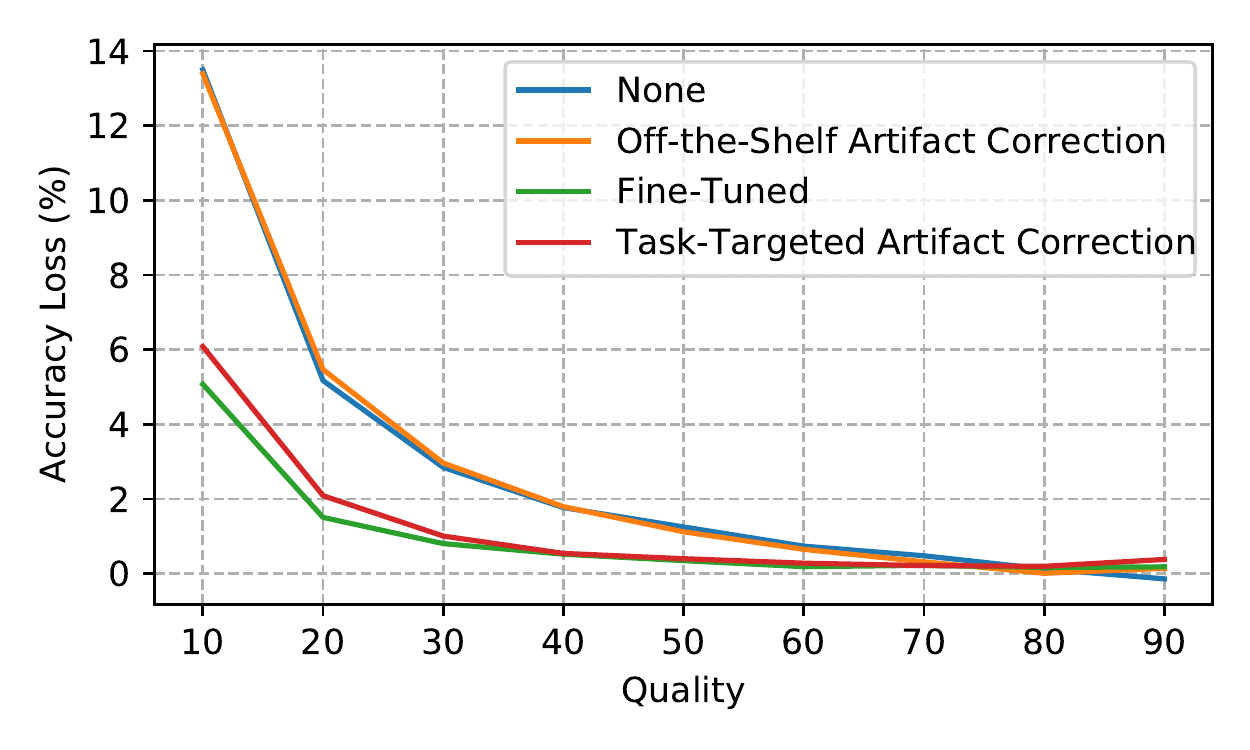}
        \caption{MobileNetV2}
    \end{subfigure}
    \hfill
    \begin{subfigure}[t]{0.48\textwidth}
        \centering
        \includegraphics[width=\textwidth]{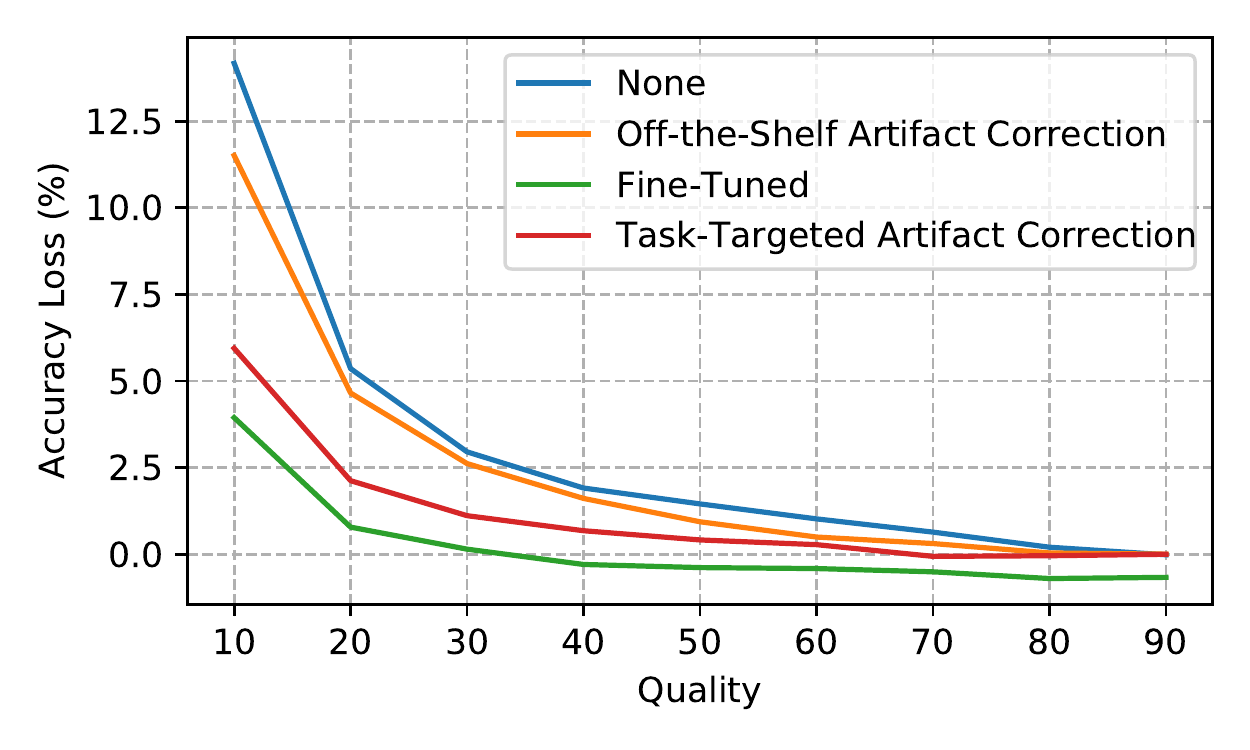}
        \caption{VGG-19}
    \end{subfigure}
\end{figure}

\begin{figure}[H]
    \ContinuedFloat
    \begin{subfigure}[t]{0.48\textwidth}
        \centering
        \includegraphics[width=\textwidth]{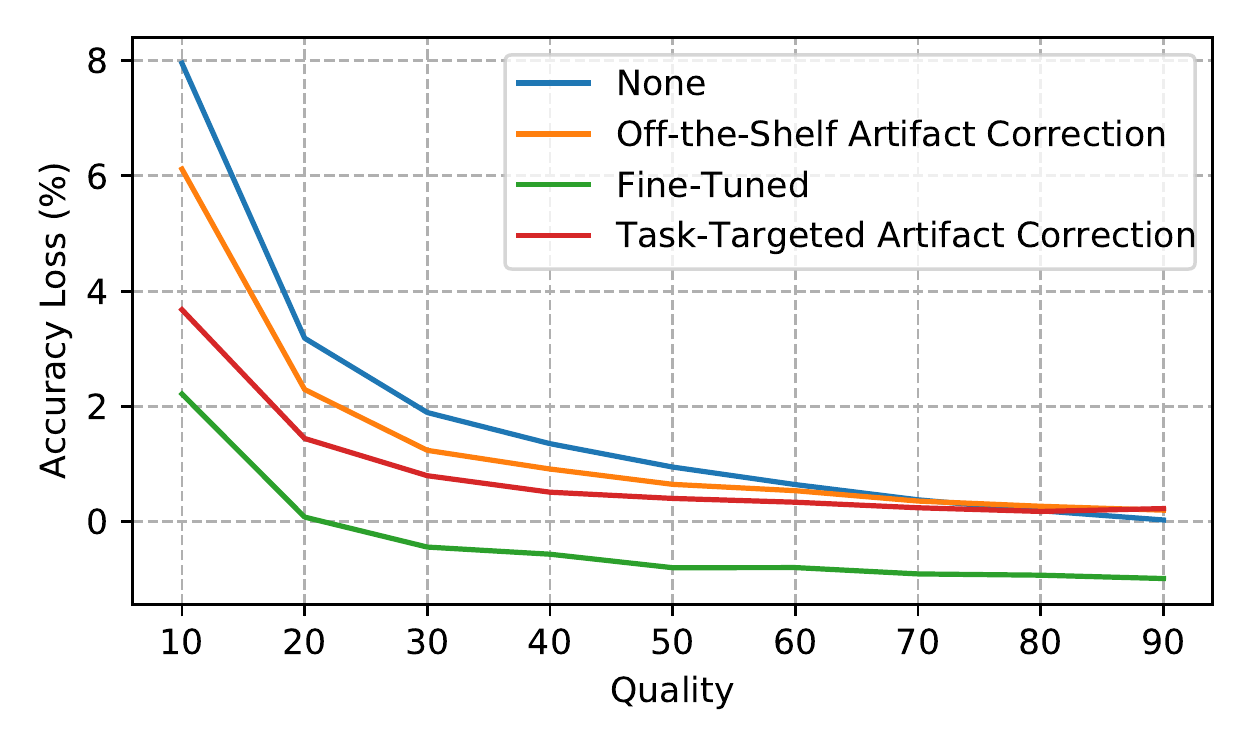}
        \caption{InceptionV3}
    \end{subfigure}
    \hfill
    \begin{subfigure}[t]{0.48\textwidth}
        \centering
        \includegraphics[width=\textwidth]{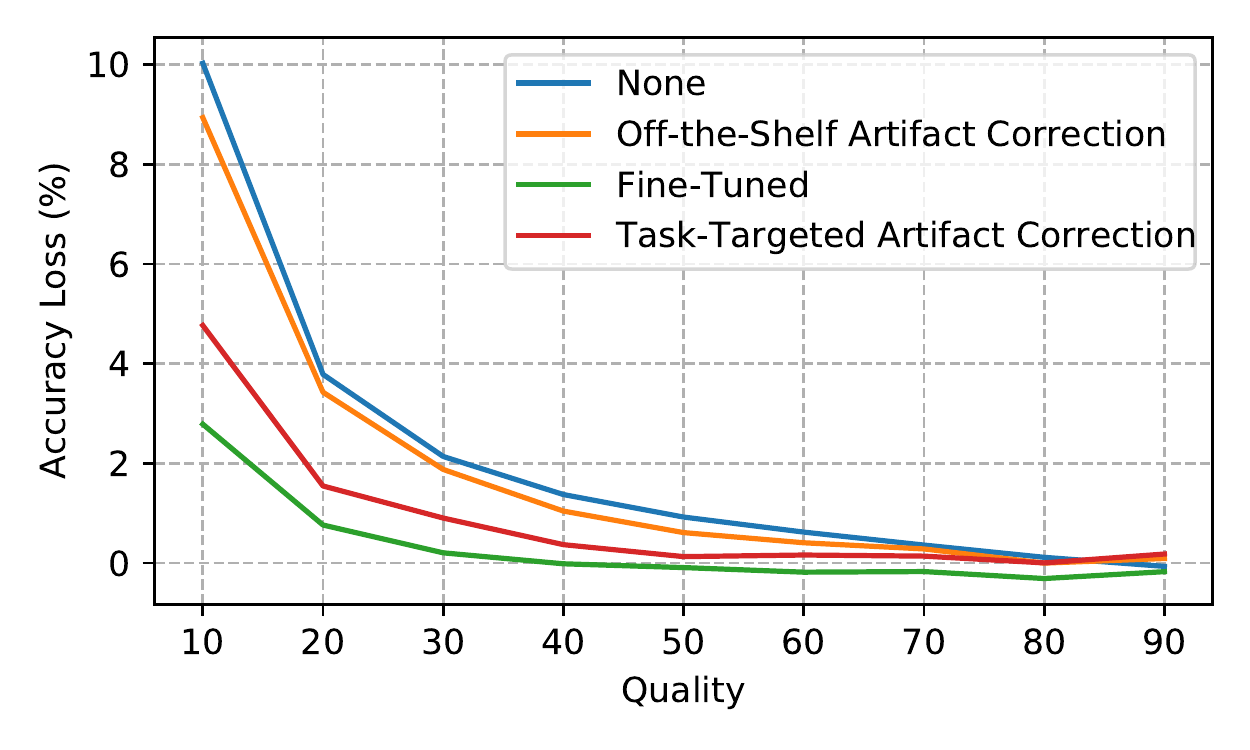}
        \caption{ResNeXt 50}
    \end{subfigure}
    \begin{subfigure}[t]{0.48\textwidth}
        \centering
        \includegraphics[width=\textwidth]{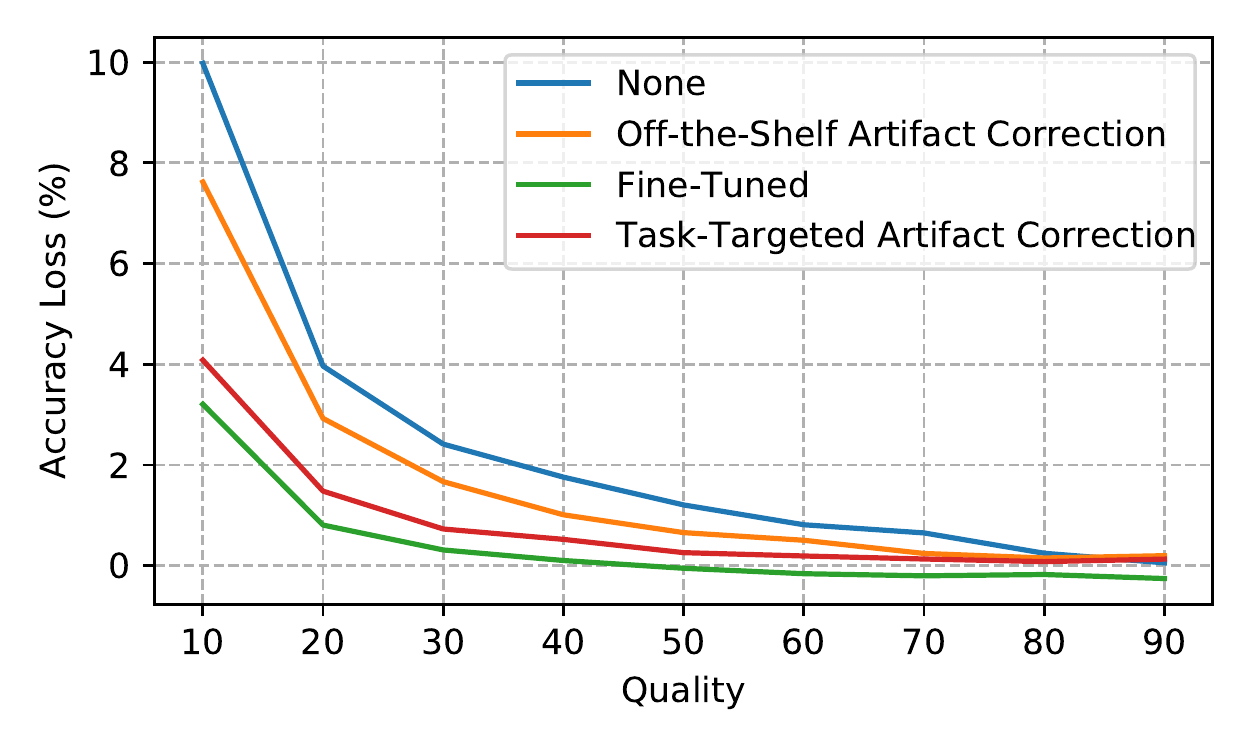}
        \caption{ResNeXt 101}
    \end{subfigure}
    \hfill
    \begin{subfigure}[t]{0.48\textwidth}
        \centering
        \includegraphics[width=\textwidth]{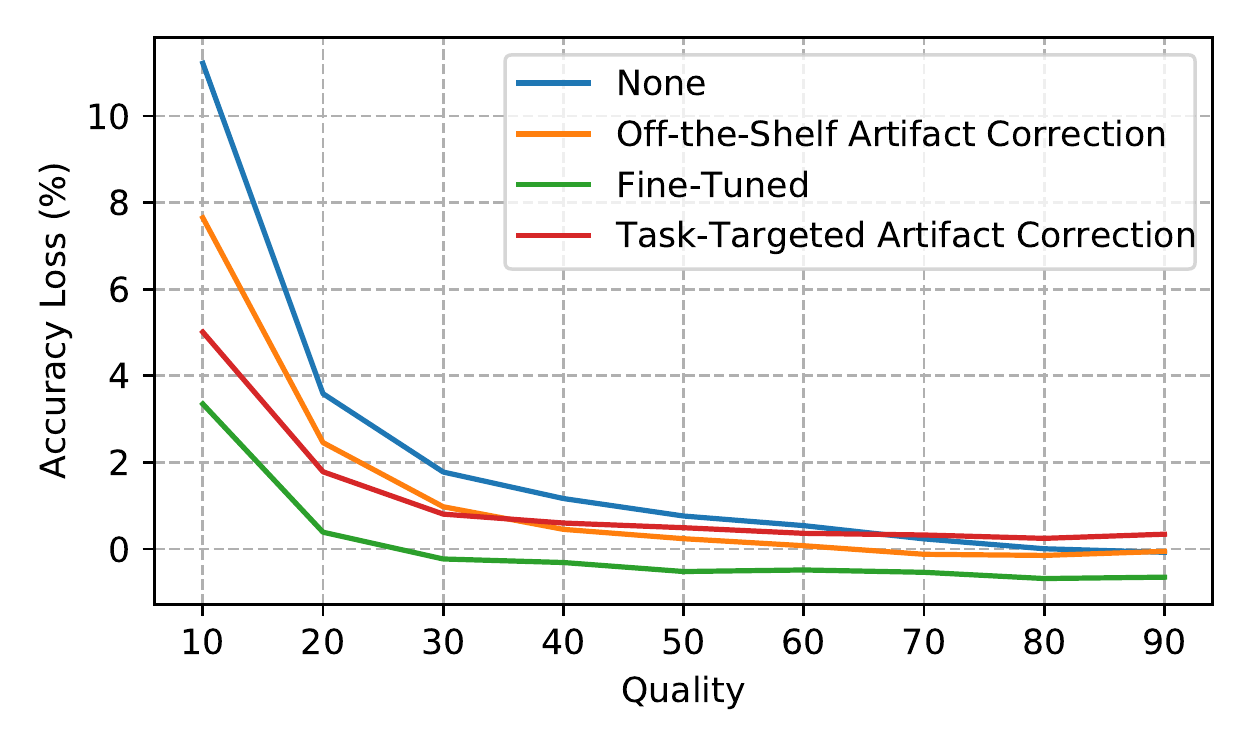}
        \caption{ResNet 18}
    \end{subfigure}
    \begin{subfigure}[t]{0.48\textwidth}
        \centering
        \includegraphics[width=\textwidth]{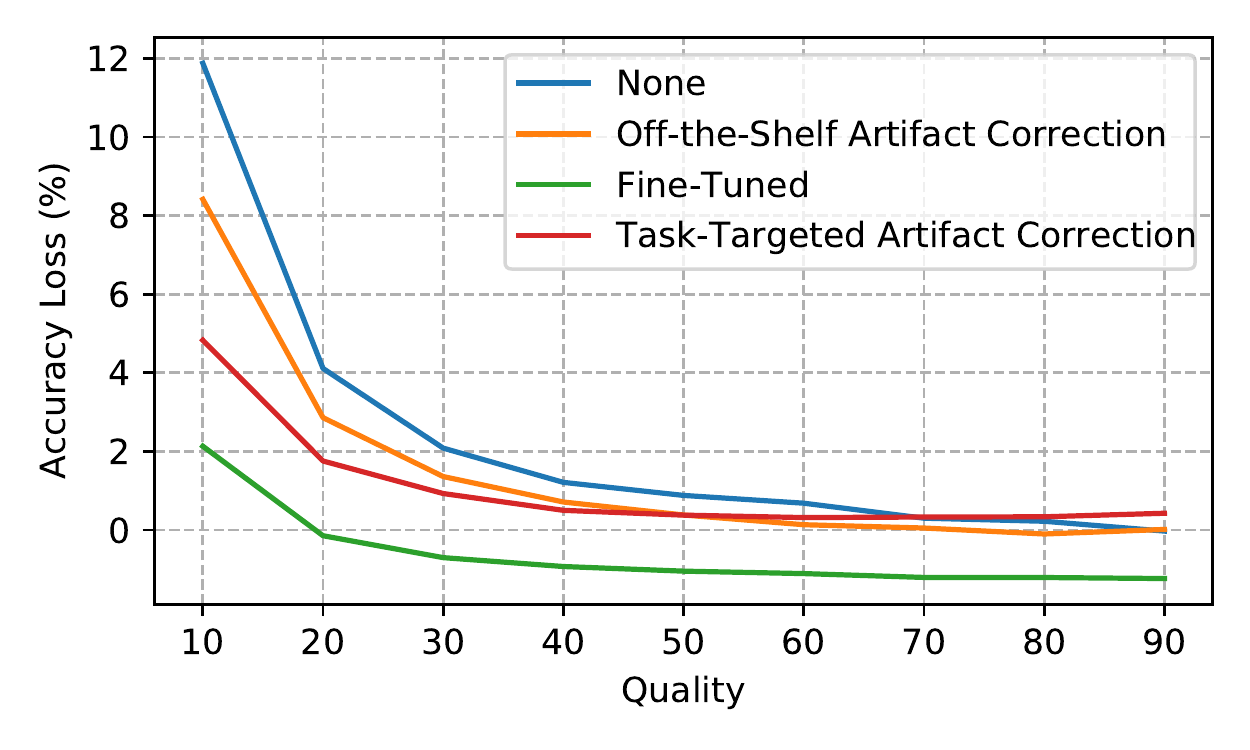}
        \caption{ResNet 50}
    \end{subfigure}
    \hfill
    \begin{subfigure}[t]{0.48\textwidth}
        \centering
        \includegraphics[width=\textwidth]{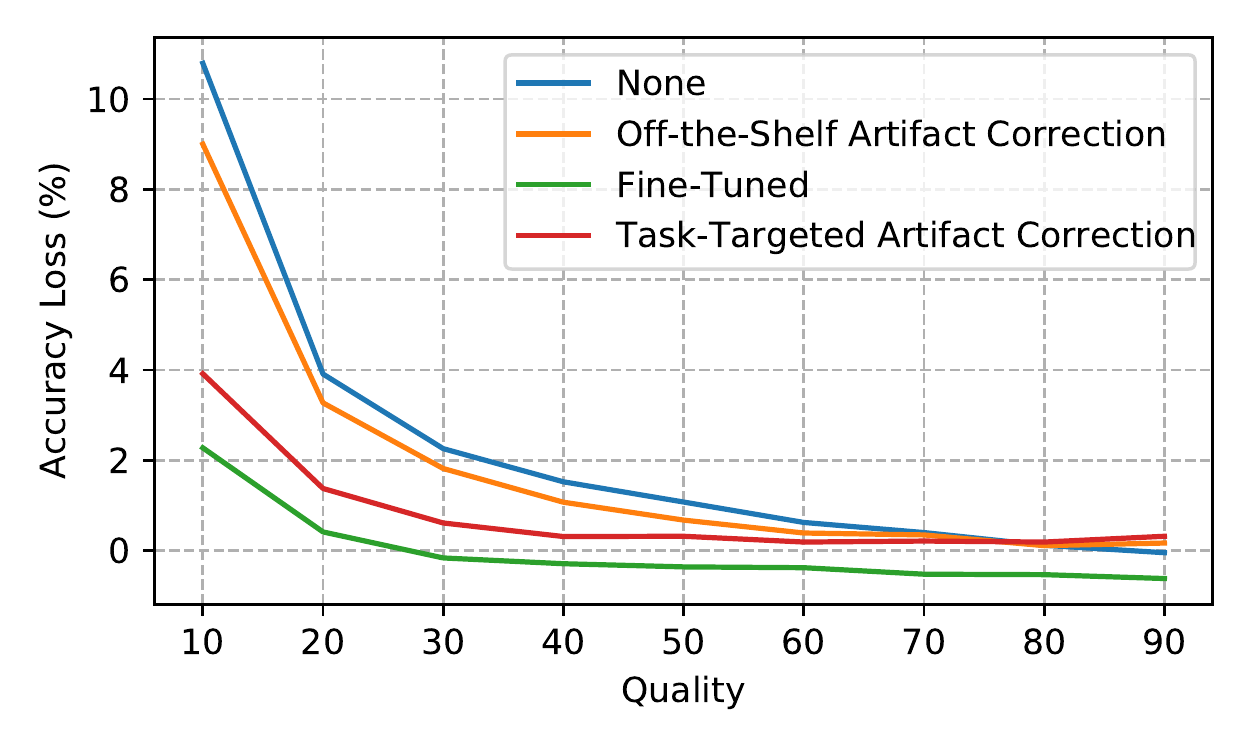}
        \caption{ResNet 101}
    \end{subfigure}
    \begin{subfigure}[t]{0.48\textwidth}
        \centering
        \includegraphics[width=\textwidth]{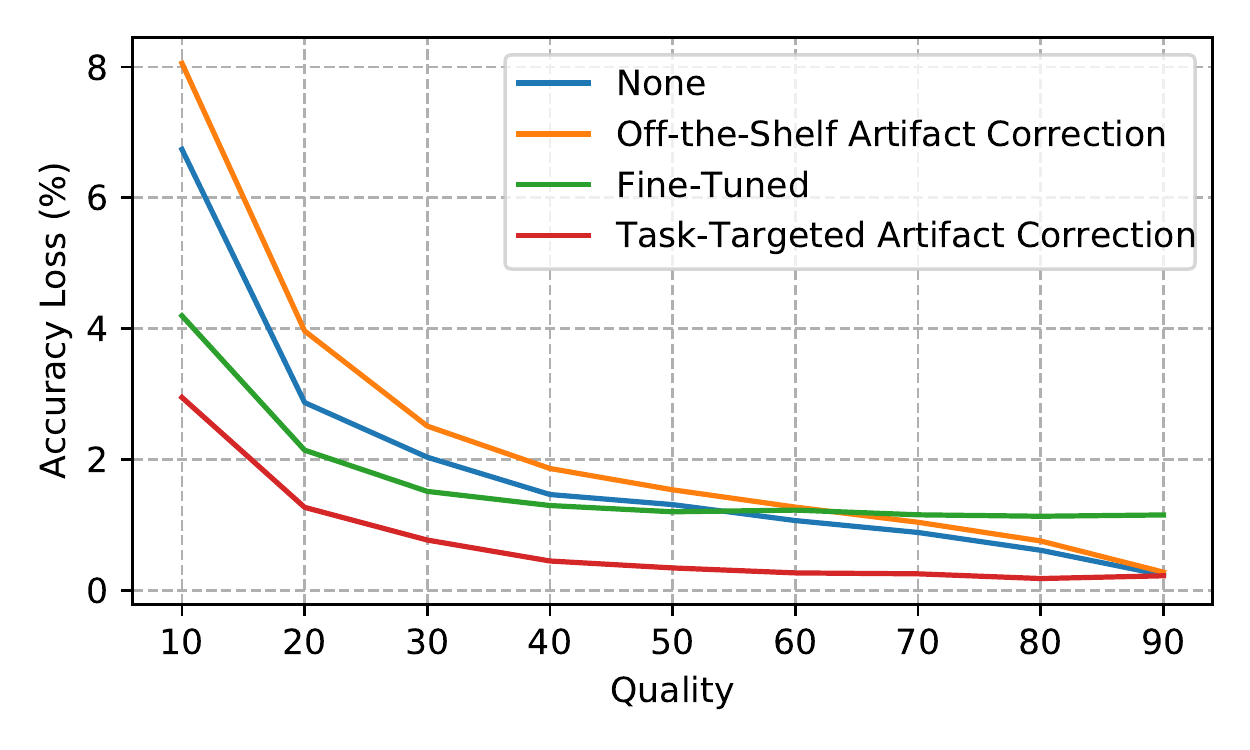}
        \caption{EfficientNet B3}
    \end{subfigure}
\end{figure}

\begin{figure}[H]
    \centering
    \includegraphics[width=\textwidth]{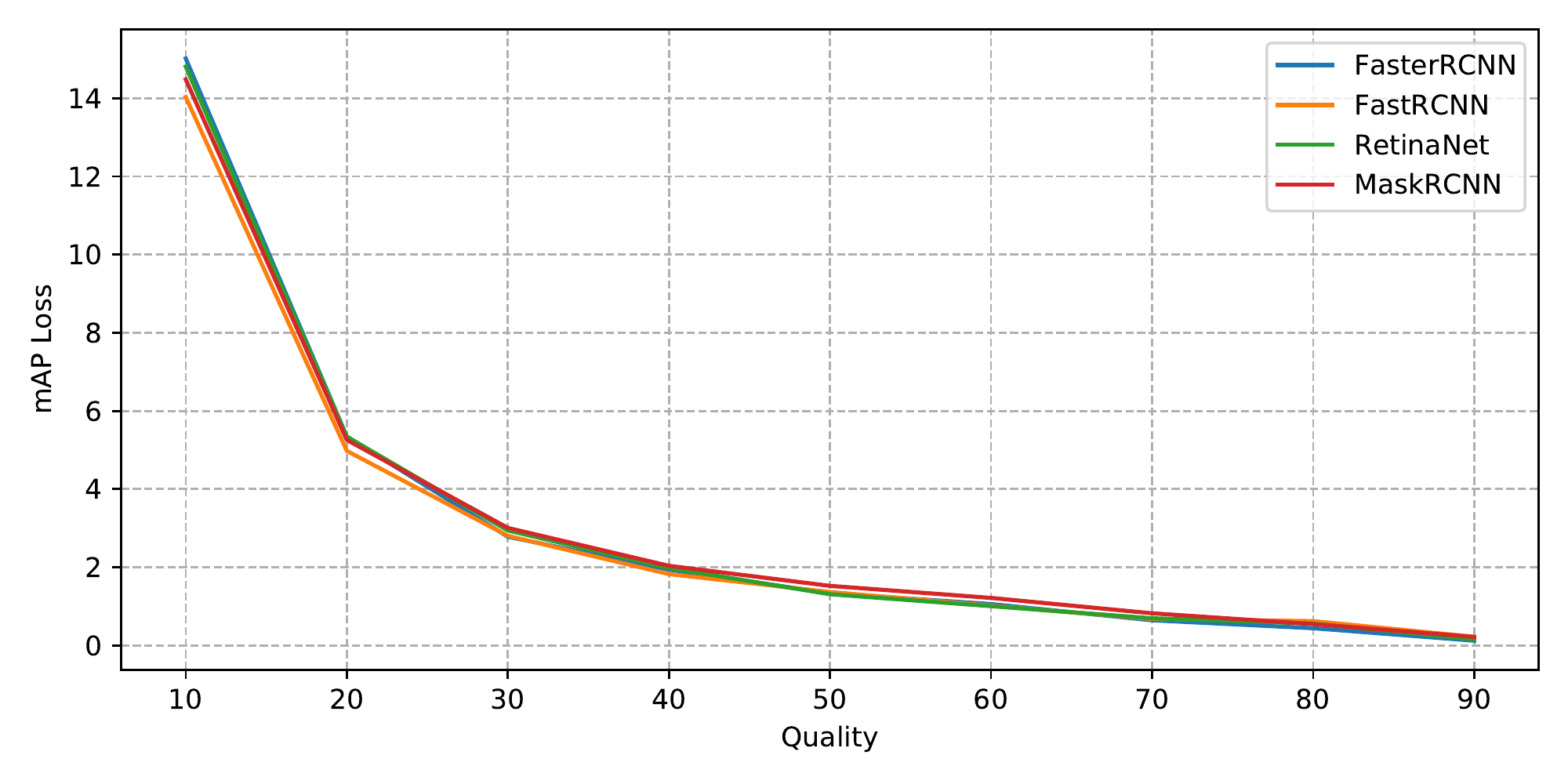}
    \caption{Detection and Instance Segmentation}
\end{figure}

\begin{figure}[H]
    \begin{subfigure}[t]{0.48\textwidth}
        \centering
        \includegraphics[width=\textwidth]{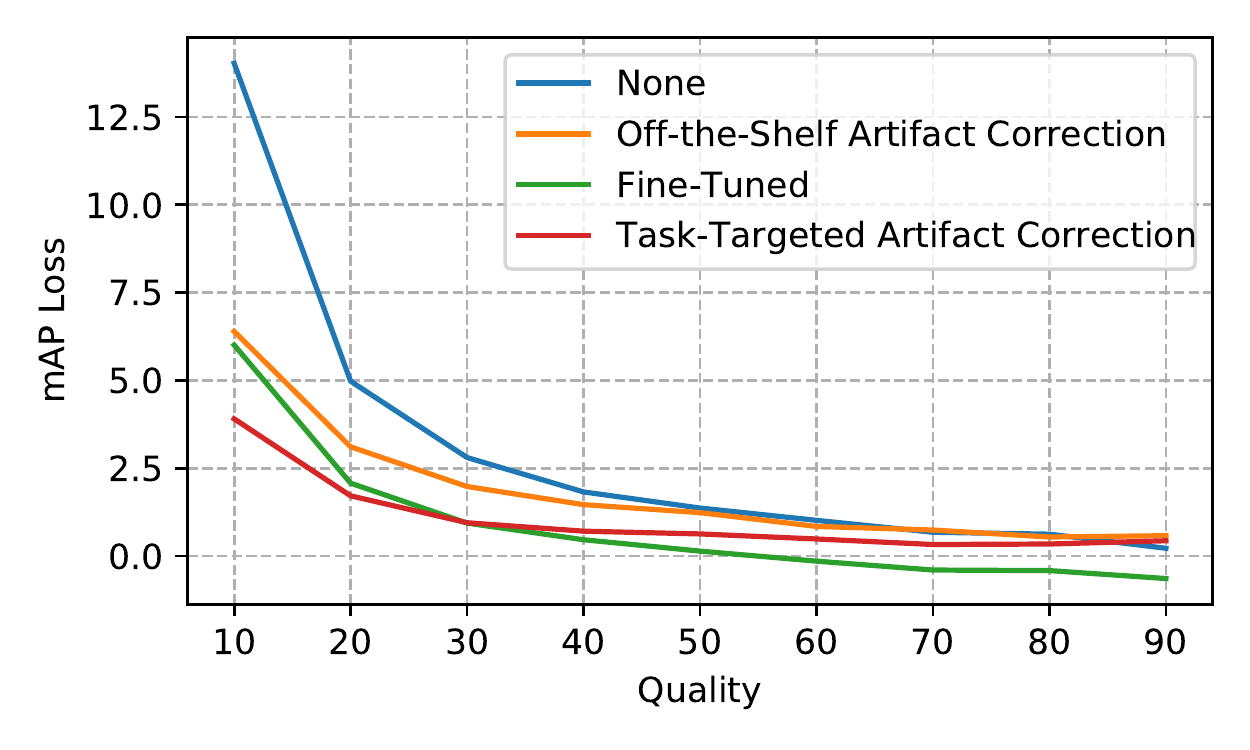}
        \vspace{-0.7cm}
        \caption{FastRCNN}
    \end{subfigure}
    \hfill
    \begin{subfigure}[t]{0.48\textwidth}
        \centering
        \includegraphics[width=\textwidth]{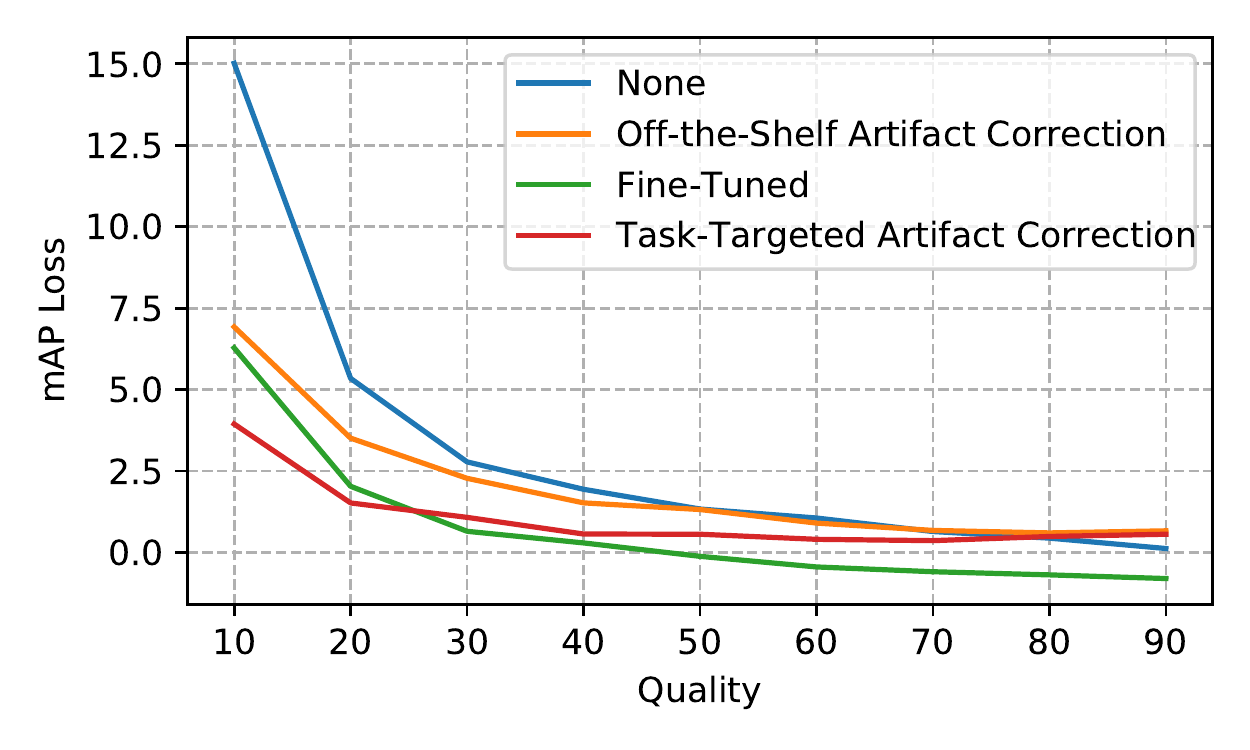}
        \vspace{-0.7cm}
        \caption{FasterRCNN}
    \end{subfigure}
    \begin{subfigure}[t]{0.48\textwidth}
        \centering
        \includegraphics[width=\textwidth]{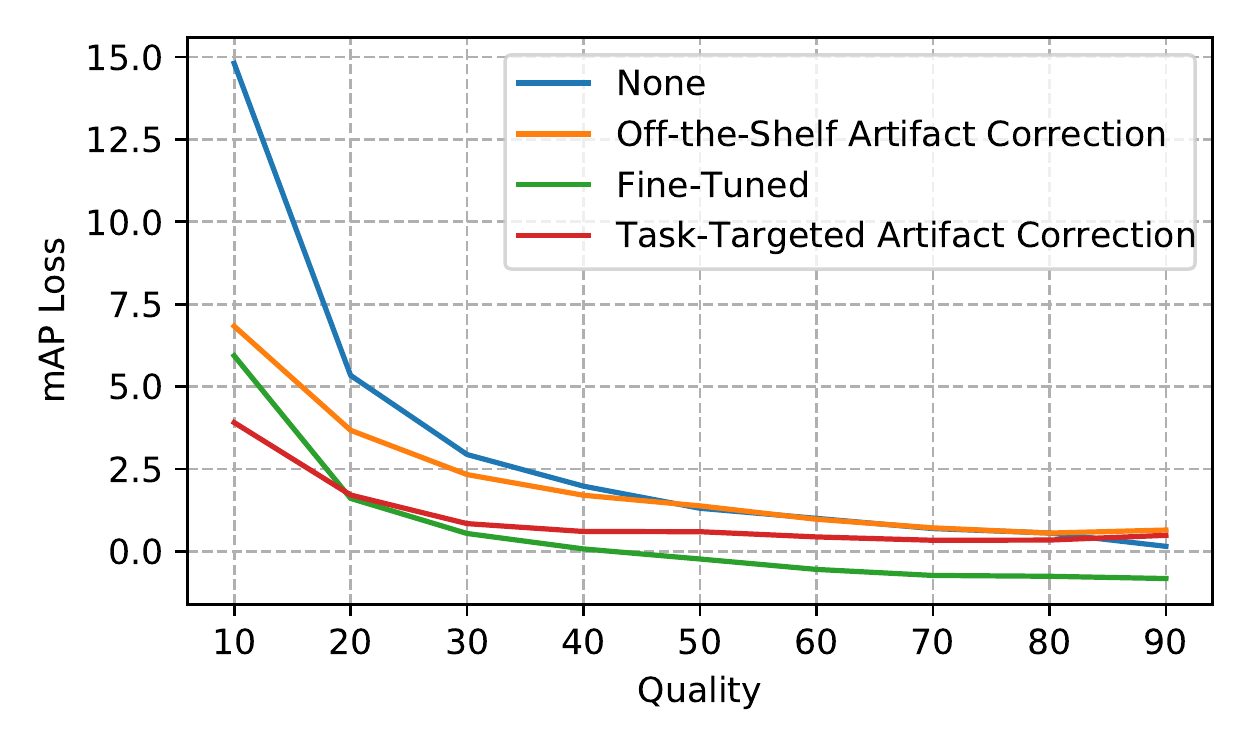}
        \vspace{-0.7cm}
        \caption{RetinaNet}
    \end{subfigure}
    \hfill
    \begin{subfigure}[t]{0.48\textwidth}
        \centering
        \includegraphics[width=\textwidth]{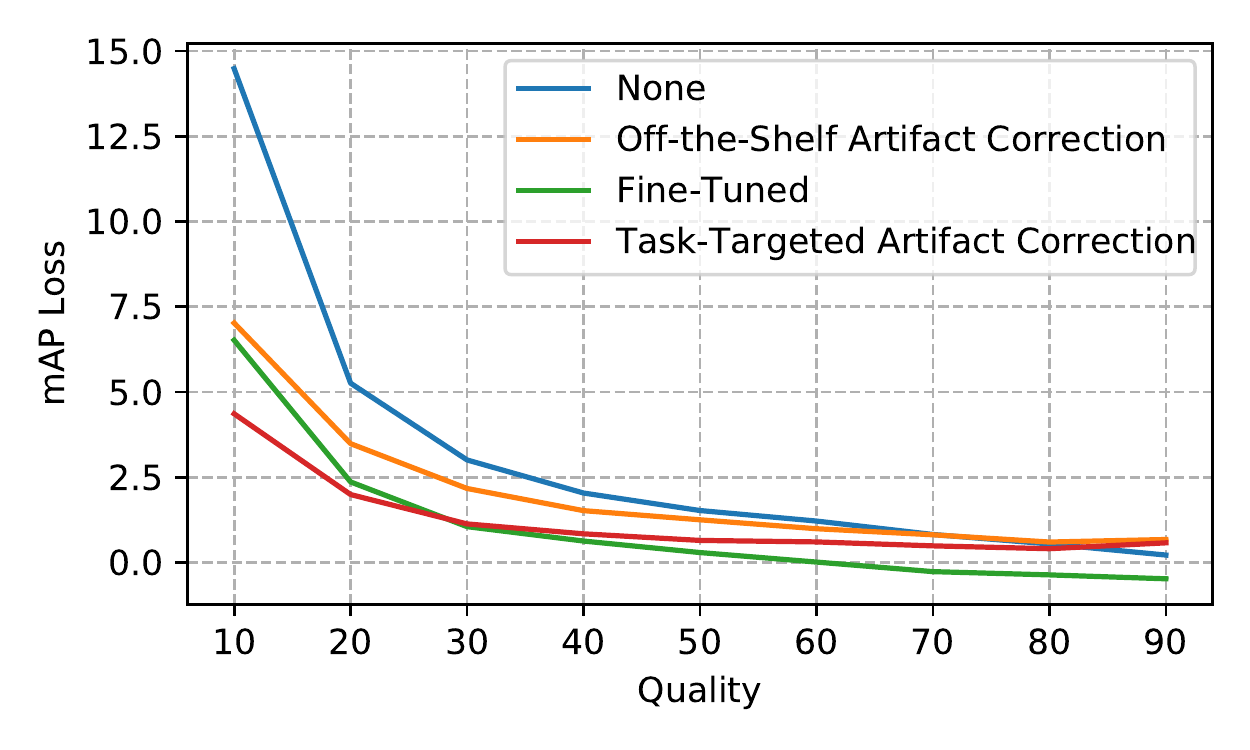}
        \vspace{-0.7cm}
        \caption{MaskRCNN}
    \end{subfigure}
\end{figure}

\begin{figure}[H]
    \centering
    \includegraphics[width=\textwidth]{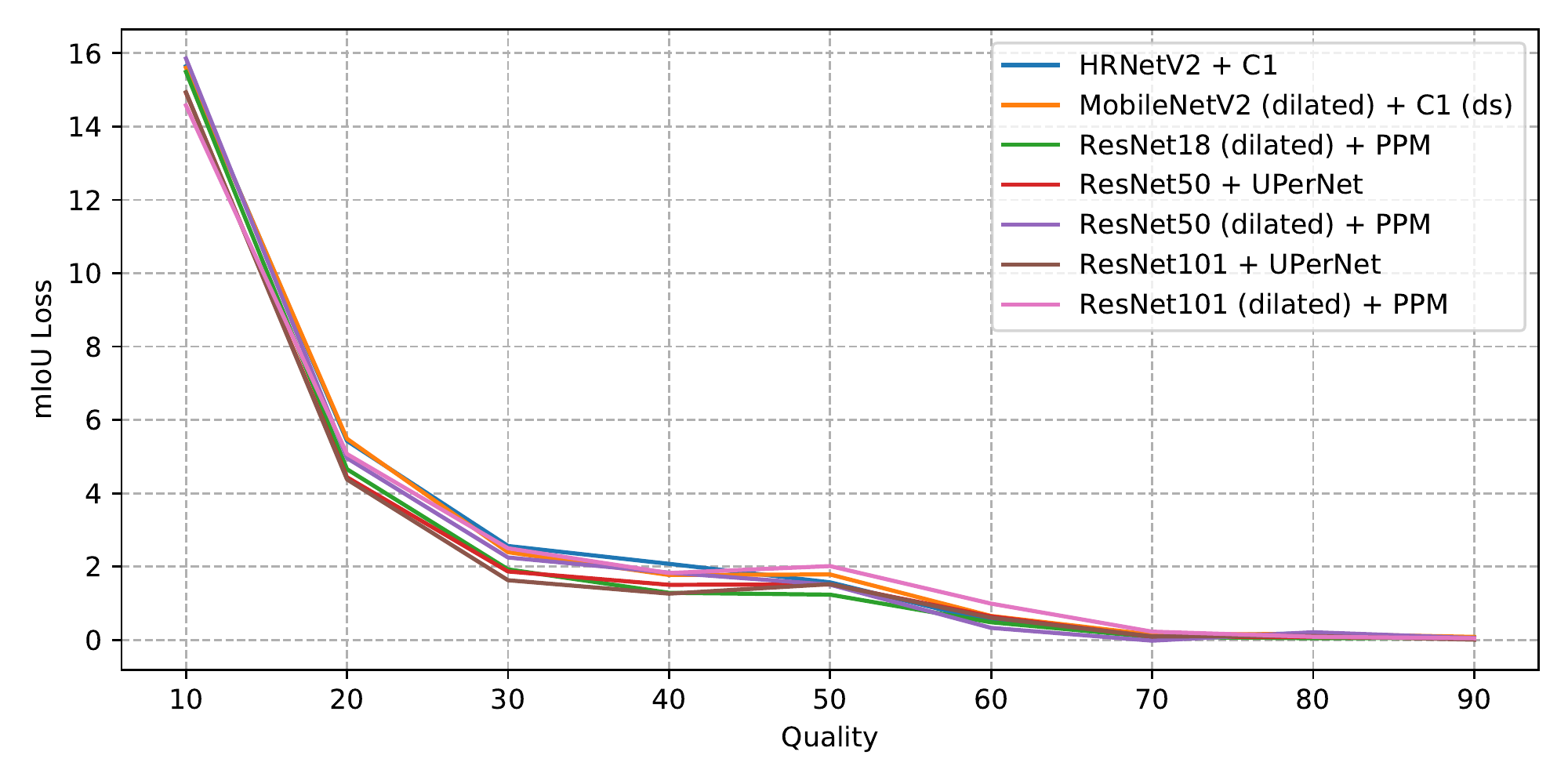}
    \caption{Semantic segmentation}
\end{figure}

\begin{figure}[H]
    \begin{subfigure}[t]{0.48\textwidth}
        \centering
        \includegraphics[width=\textwidth]{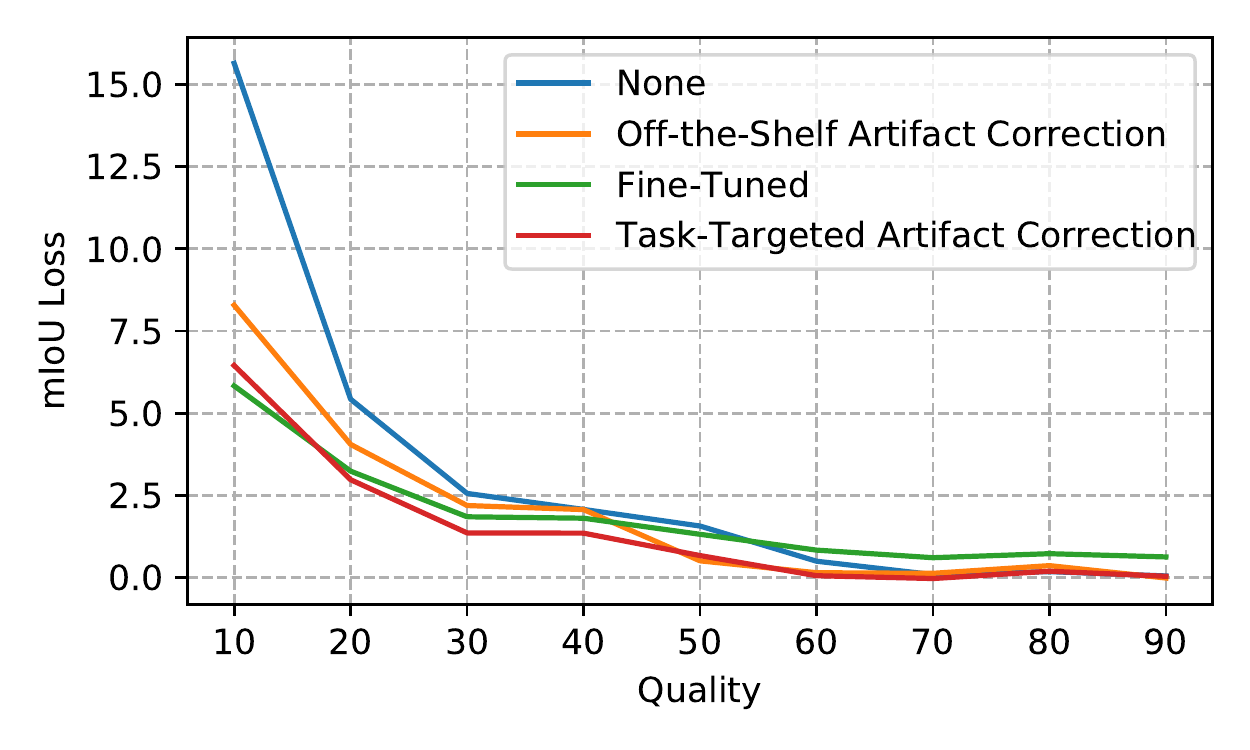}
        \vspace{-0.7cm}
        \caption{HRNetV2 + C1}
    \end{subfigure}
    \hfill
    \begin{subfigure}[t]{0.48\textwidth}
        \centering
        \includegraphics[width=\textwidth]{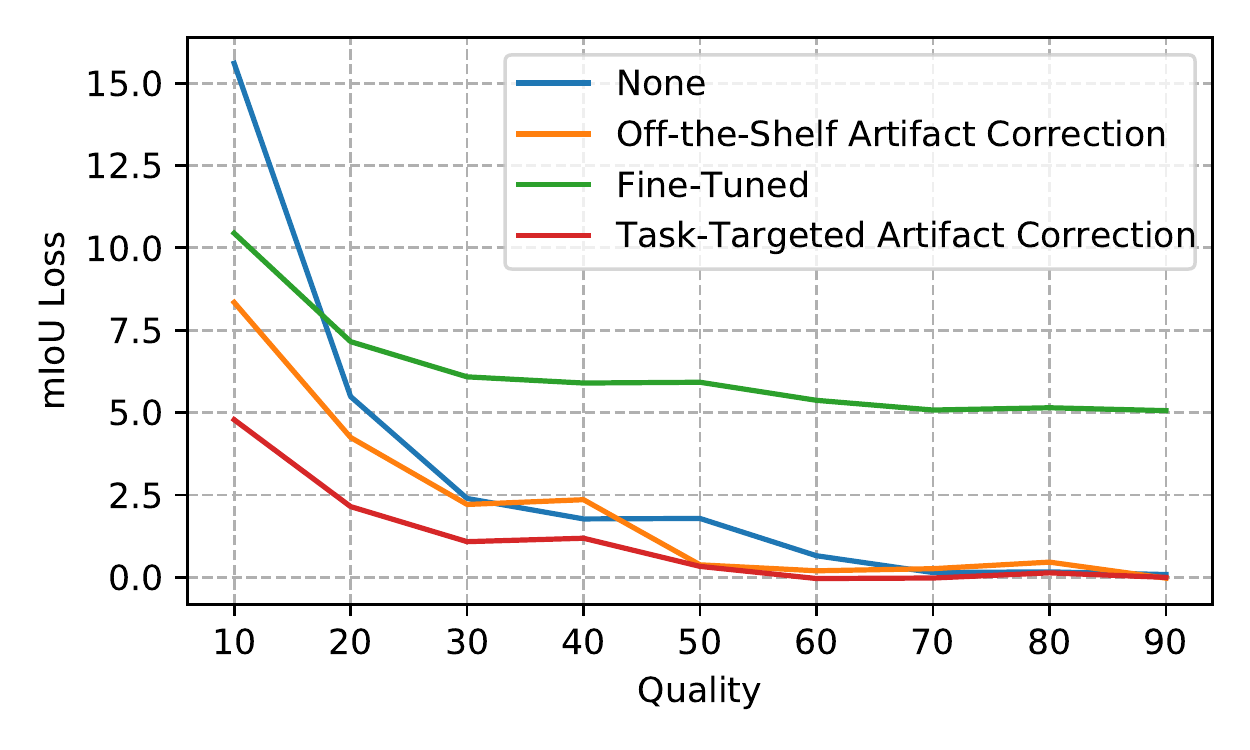}
        \vspace{-0.7cm}
        \caption{MobileNetV2 + C1}
    \end{subfigure}
    \begin{subfigure}[t]{0.48\textwidth}
        \centering
        \includegraphics[width=\textwidth]{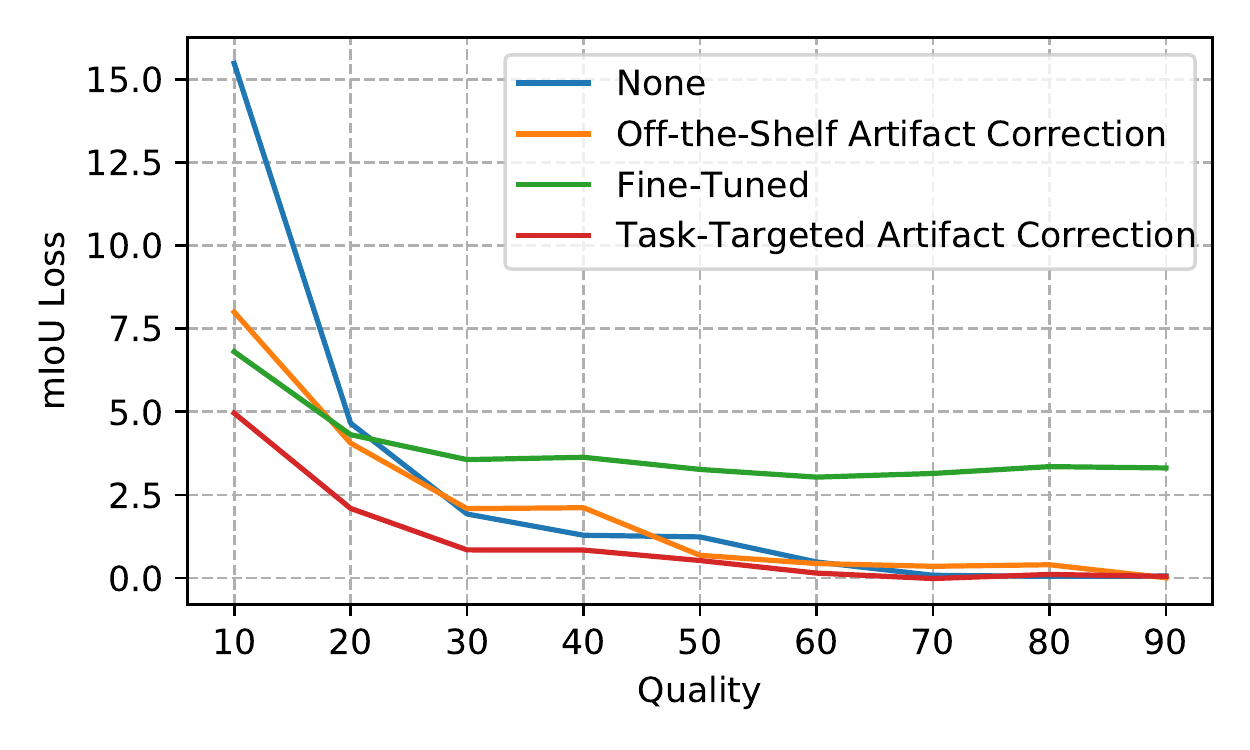}
        \vspace{-0.7cm}
        \caption{ResNet 18 + PPM}
    \end{subfigure}
    \hfill
    \begin{subfigure}[t]{0.48\textwidth}
        \centering
        \includegraphics[width=\textwidth]{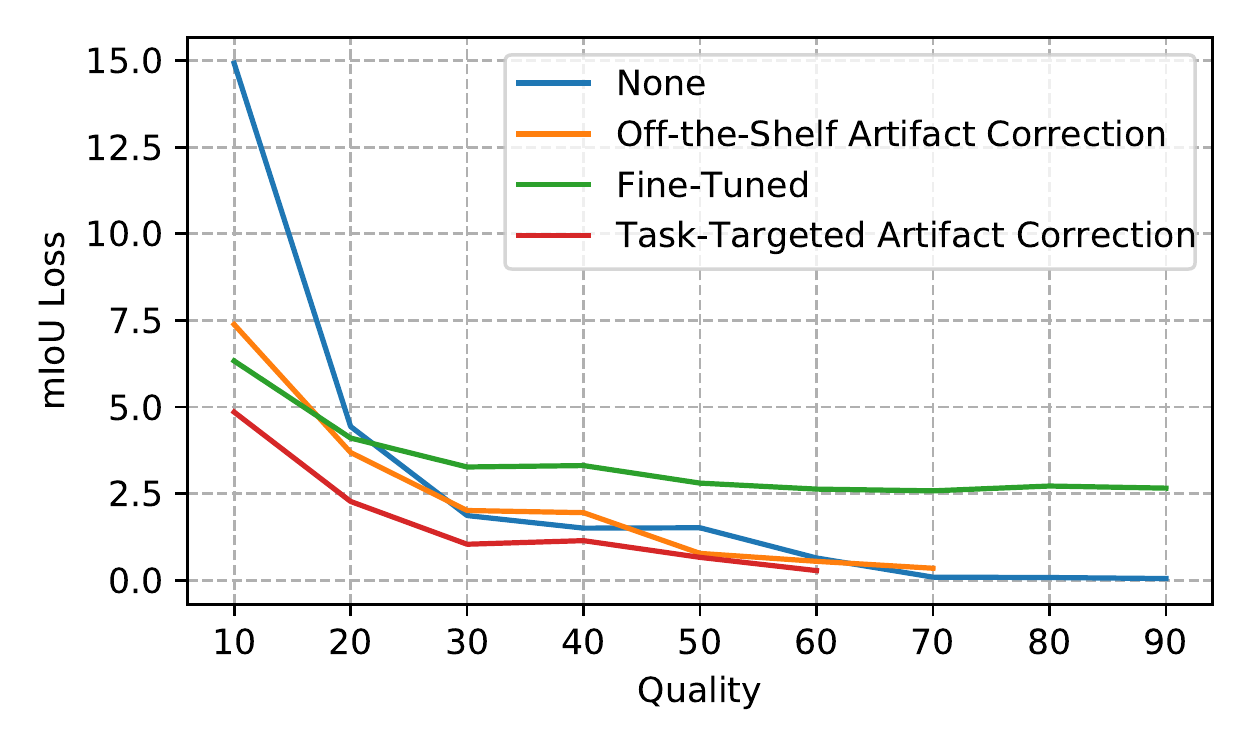}
        \vspace{-0.7cm}
        \caption{ResNet 50 + UPerNet}
    \end{subfigure}
\end{figure}
\begin{figure}[H]
    \ContinuedFloat
    \begin{subfigure}[t]{0.48\textwidth}
        \centering
        \includegraphics[width=\textwidth]{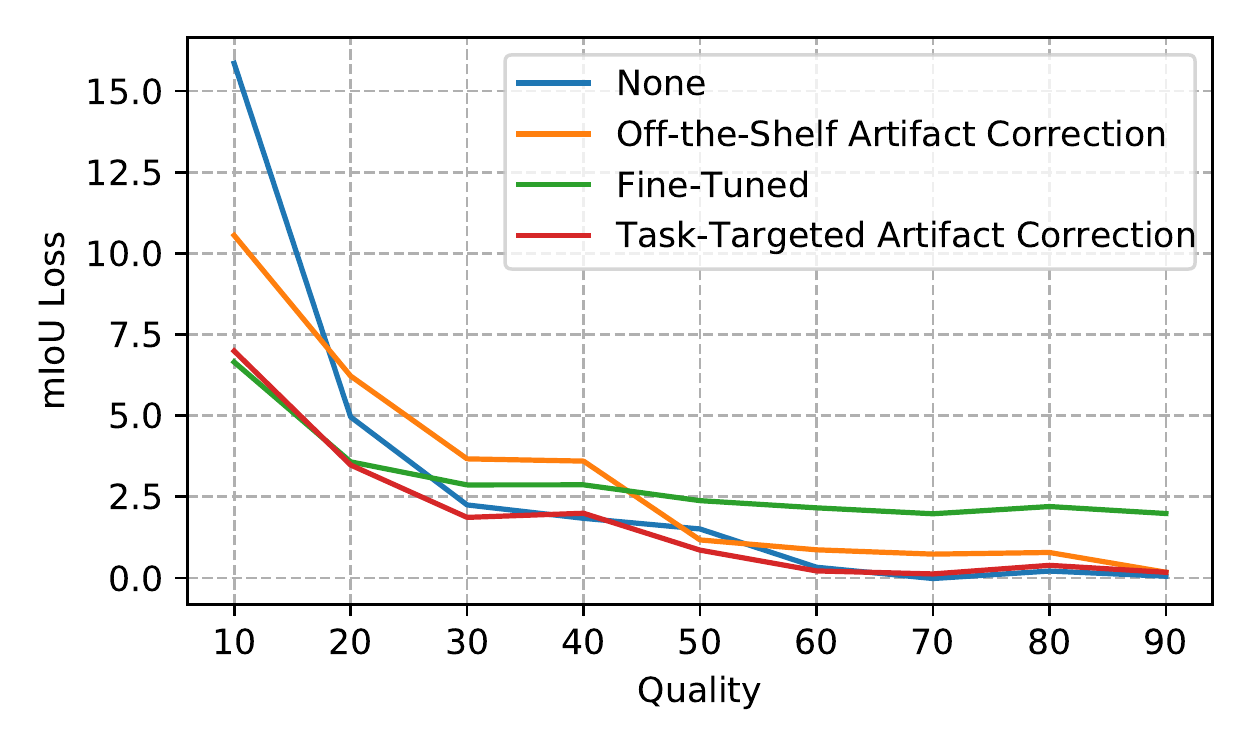}
        \vspace{-0.7cm}
        \caption{ResNet 50 + PPM}
    \end{subfigure}
    \hfill
    \begin{subfigure}[t]{0.48\textwidth}
        \centering
        \includegraphics[width=\textwidth]{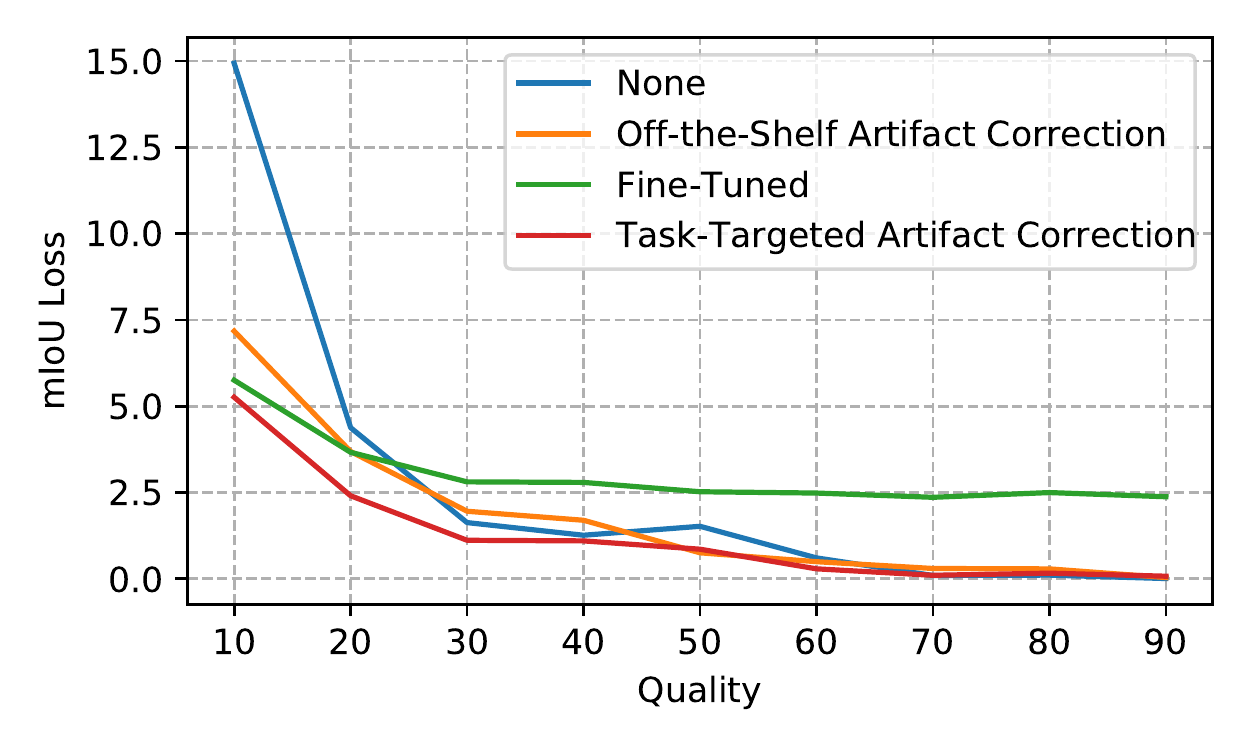}
        \vspace{-0.7cm}
        \caption{ResNet 101 + UPerNet}
    \end{subfigure}
    \begin{subfigure}[t]{0.48\textwidth}
        \centering
        \includegraphics[width=\textwidth]{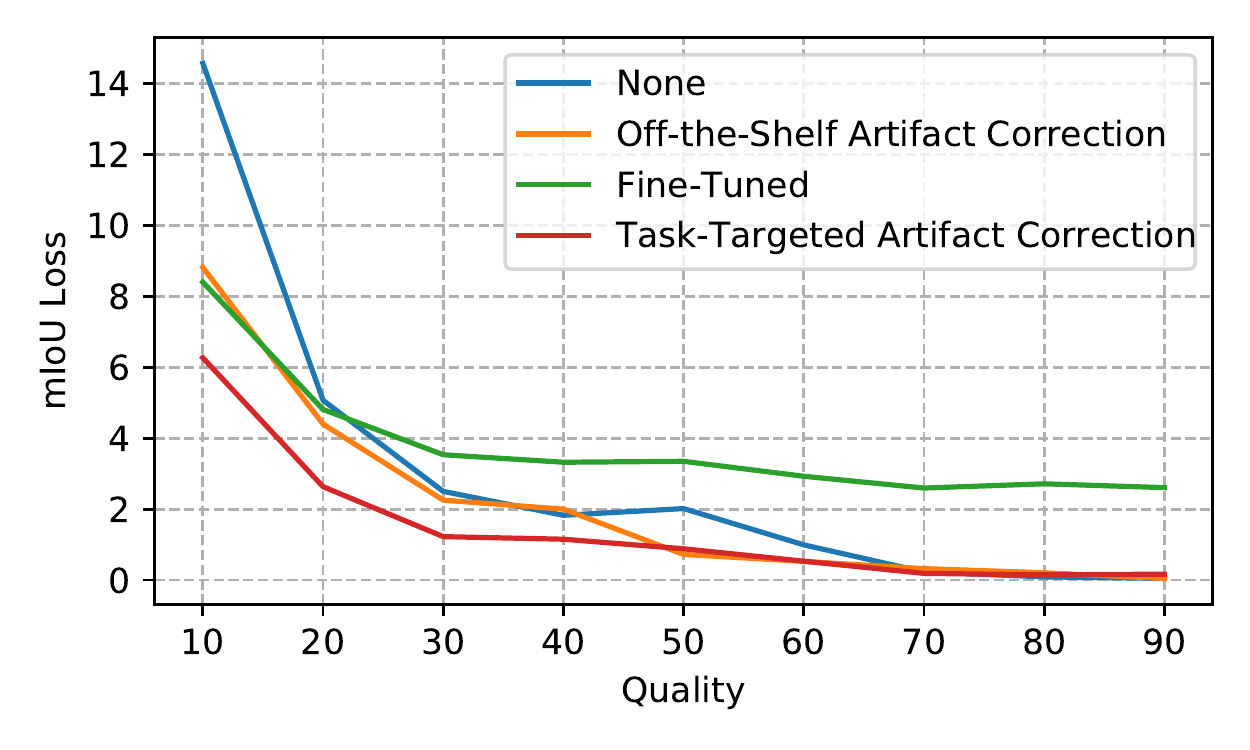}
        \vspace{-0.7cm}
        \caption{ResNet 101 + PPM}
    \end{subfigure}
\end{figure}

\begin{figure}[H]
    \centering
    \includegraphics[width=\textwidth]{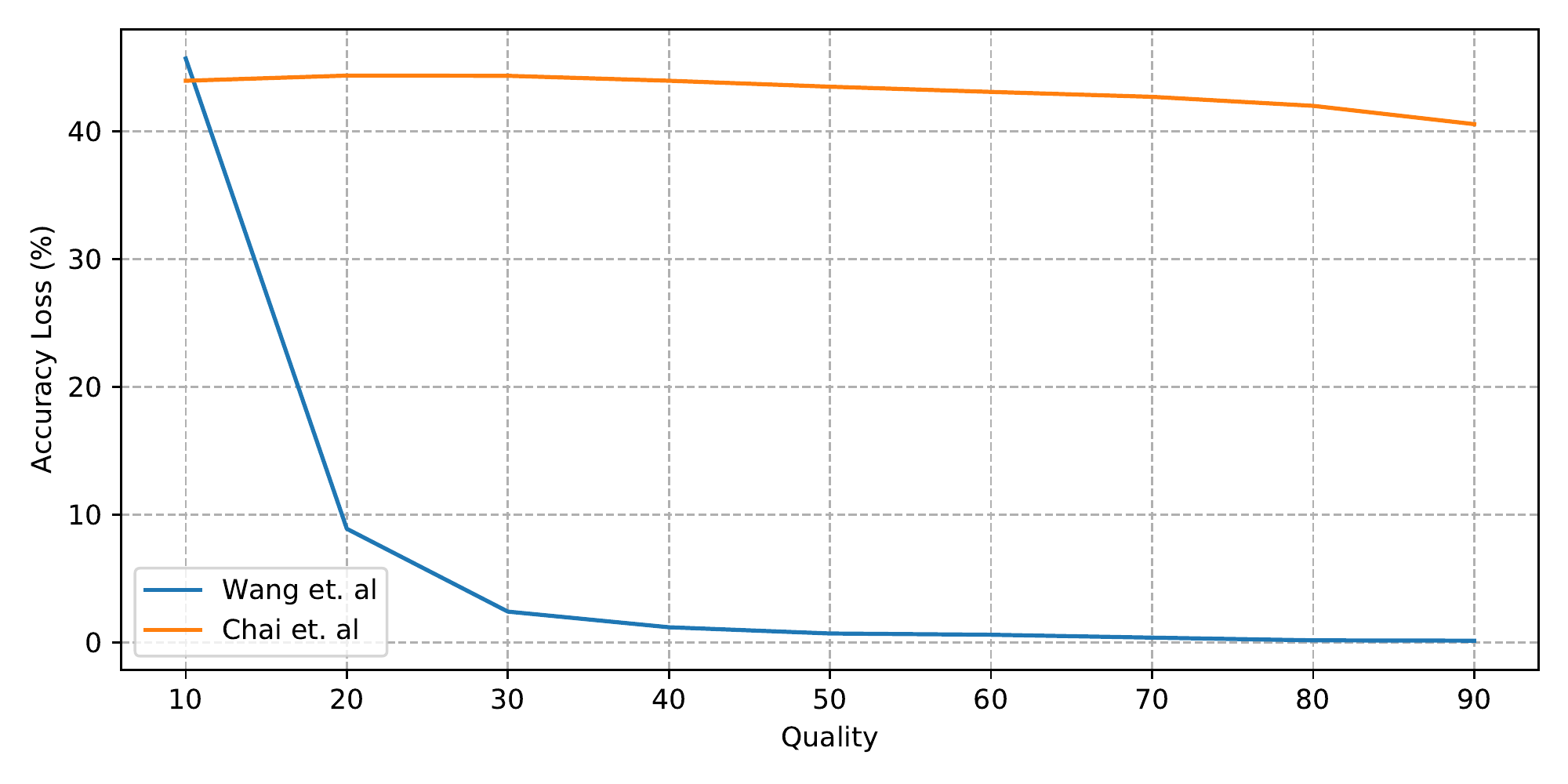}
    \caption{Forensics}
\end{figure}

\begin{figure}[H]
    \begin{subfigure}[t]{0.48\textwidth}
        \centering
        \includegraphics[width=\textwidth]{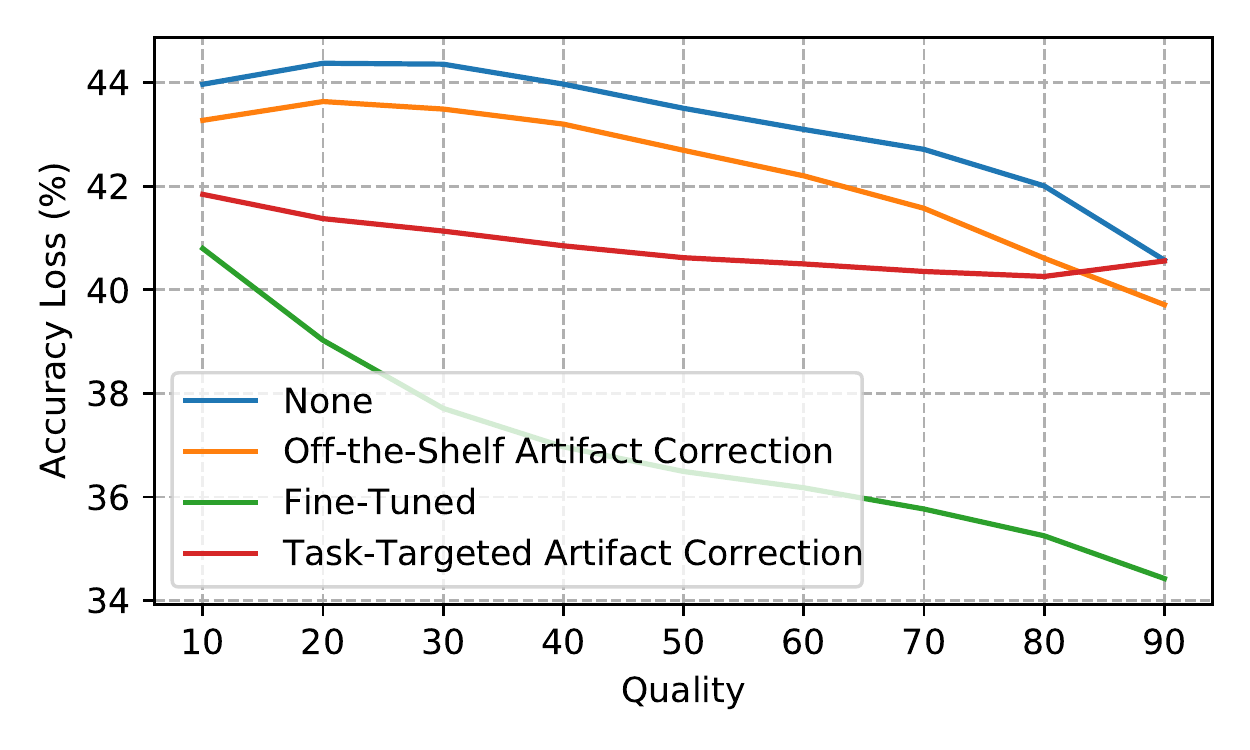}
        \vspace{-0.7cm}
        \caption{Chai \etal}
    \end{subfigure}
    \begin{subfigure}[t]{0.48\textwidth}
        \centering
        \includegraphics[width=\textwidth]{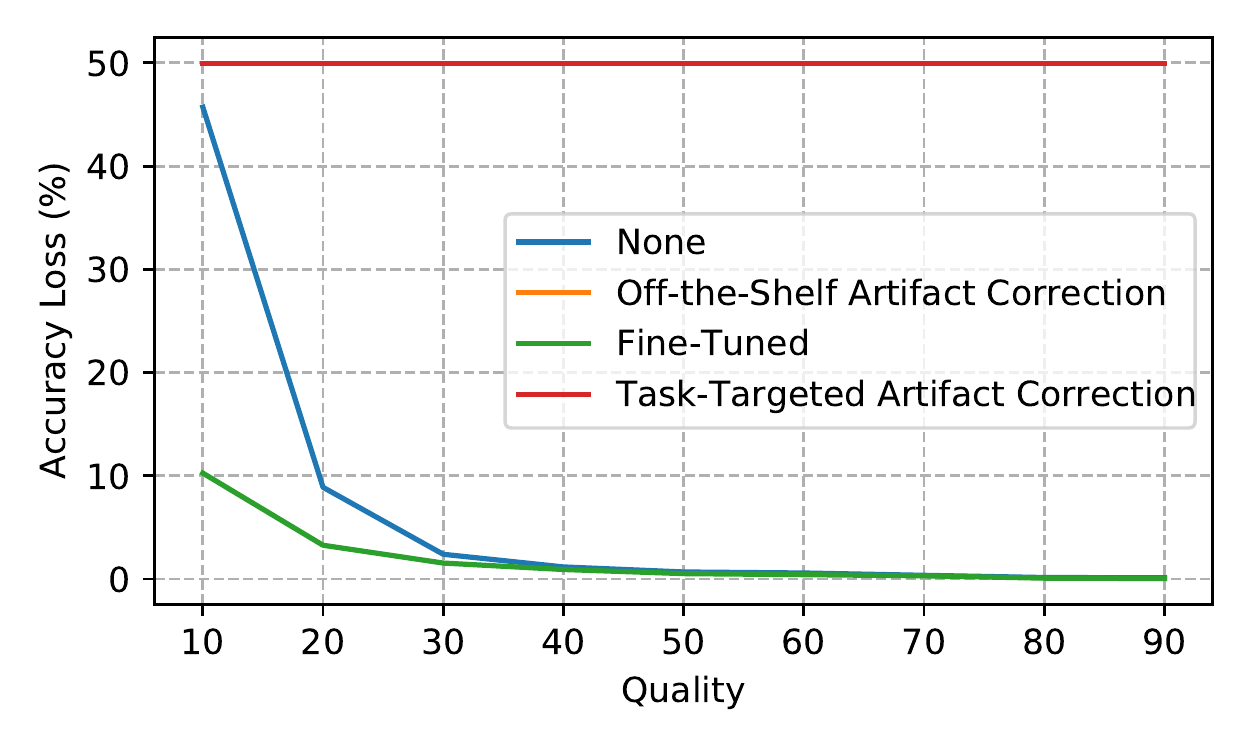}
        \vspace{-0.7cm}
        \caption{Wang \etal}
    \end{subfigure}
\end{figure}

\subsection{Tables of Results}

\resizebox{\textwidth}{!}{
    \begin{tabular}{llllrrrrrrrrr}
    \toprule
    Model                            & Metric                          & Reference              & Mitigation                        & Q=10  & Q=20  & Q=30  & Q=40  & Q=50  & Q=60  & Q=70  & Q=80  & Q=90  \\
    \midrule
    \multirow{4}{*}{EfficientNet B3} & \multirow{4}{*}{Top-1 Accuracy} & \multirow{4}{*}{83.98} & Supervised Fine-Tuning            & 79.78 & 81.84 & 82.47 & 82.68 & 82.78 & 82.75 & 82.83 & 82.85 & 82.83 \\
                                     &                                 &                        & None                              & 77.24 & 81.11 & 81.95 & 82.52 & 82.67 & 82.91 & 83.10 & 83.37 & 83.75 \\
                                     &                                 &                        & Off-the-Shelf Artifact Correction & 75.92 & 80.02 & 81.47 & 82.12 & 82.44 & 82.71 & 82.94 & 83.23 & 83.70 \\
                                     &                                 &                        & Task-Targeted Artifact Correction & 81.03 & 82.71 & 83.21 & 83.53 & 83.64 & 83.71 & 83.73 & 83.80 & 83.76 \\
    \midrule
    \multirow{4}{*}{InceptionV3}     & \multirow{4}{*}{Top-1 Accuracy} & \multirow{4}{*}{77.33} & Supervised Fine-Tuning            & 75.11 & 77.25 & 77.77 & 77.89 & 78.13 & 78.13 & 78.24 & 78.26 & 78.32 \\
                                     &                                 &                        & None                              & 69.38 & 74.15 & 75.44 & 75.98 & 76.38 & 76.69 & 76.95 & 77.14 & 77.30 \\
                                     &                                 &                        & Off-the-Shelf Artifact Correction & 71.21 & 75.04 & 76.09 & 76.42 & 76.68 & 76.79 & 76.97 & 77.06 & 77.13 \\
                                     &                                 &                        & Task-Targeted Artifact Correction & 73.65 & 75.89 & 76.53 & 76.82 & 76.93 & 76.99 & 77.09 & 77.15 & 77.10 \\
    \midrule
    \multirow{4}{*}{MobileNetV2}     & \multirow{4}{*}{Top-1 Accuracy} & \multirow{4}{*}{70.72} & Supervised Fine-Tuning            & 65.65 & 69.21 & 69.92 & 70.20 & 70.37 & 70.53 & 70.50 & 70.55 & 70.54 \\
                                     &                                 &                        & None                              & 57.23 & 65.55 & 67.87 & 68.95 & 69.47 & 69.98 & 70.24 & 70.60 & 70.86 \\
                                     &                                 &                        & Off-the-Shelf Artifact Correction & 57.33 & 65.25 & 67.76 & 68.93 & 69.60 & 70.07 & 70.40 & 70.71 & 70.58 \\
                                     &                                 &                        & Task-Targeted Artifact Correction & 64.64 & 68.63 & 69.71 & 70.18 & 70.32 & 70.44 & 70.50 & 70.52 & 70.34 \\
    \midrule
    \multirow{4}{*}{ResNet-101}      & \multirow{4}{*}{Top-1 Accuracy} & \multirow{4}{*}{76.91} & Supervised Fine-Tuning            & 74.63 & 76.50 & 77.07 & 77.20 & 77.27 & 77.29 & 77.43 & 77.44 & 77.53 \\
                                     &                                 &                        & None                              & 66.12 & 73.00 & 74.65 & 75.39 & 75.83 & 76.29 & 76.51 & 76.79 & 76.96 \\
                                     &                                 &                        & Off-the-Shelf Artifact Correction & 67.91 & 73.64 & 75.09 & 75.84 & 76.23 & 76.52 & 76.56 & 76.80 & 76.74 \\
                                     &                                 &                        & Task-Targeted Artifact Correction & 72.99 & 75.53 & 76.30 & 76.60 & 76.59 & 76.72 & 76.70 & 76.72 & 76.59 \\
    \midrule
    \multirow{4}{*}{ResNet-18}       & \multirow{4}{*}{Top-1 Accuracy} & \multirow{4}{*}{68.84} & Supervised Fine-Tuning            & 65.49 & 68.46 & 69.07 & 69.16 & 69.36 & 69.33 & 69.38 & 69.53 & 69.49 \\
                                     &                                 &                        & None                              & 57.62 & 65.26 & 67.07 & 67.68 & 68.08 & 68.30 & 68.61 & 68.84 & 68.92 \\
                                     &                                 &                        & Off-the-Shelf Artifact Correction & 61.19 & 66.39 & 67.87 & 68.39 & 68.61 & 68.77 & 68.97 & 68.99 & 68.90 \\
                                     &                                 &                        & Task-Targeted Artifact Correction & 63.83 & 67.06 & 68.04 & 68.24 & 68.35 & 68.48 & 68.52 & 68.60 & 68.50 \\
    \midrule
    \multirow{4}{*}{ResNet-50}       & \multirow{4}{*}{Top-1 Accuracy} & \multirow{4}{*}{75.31} & Supervised Fine-Tuning            & 73.18 & 75.46 & 76.02 & 76.24 & 76.36 & 76.42 & 76.52 & 76.52 & 76.55 \\
                                     &                                 &                        & None                              & 63.43 & 71.20 & 73.23 & 74.10 & 74.43 & 74.63 & 75.01 & 75.09 & 75.34 \\
                                     &                                 &                        & Off-the-Shelf Artifact Correction & 66.90 & 72.45 & 73.95 & 74.60 & 74.93 & 75.18 & 75.26 & 75.42 & 75.30 \\
                                     &                                 &                        & Task-Targeted Artifact Correction & 70.48 & 73.56 & 74.39 & 74.81 & 74.94 & 75.00 & 74.98 & 74.98 & 74.89 \\
    \midrule
    \multirow{4}{*}{ResNeXt-101}     & \multirow{4}{*}{Top-1 Accuracy} & \multirow{4}{*}{78.81} & Supervised Fine-Tuning            & 75.60 & 78.00 & 78.50 & 78.71 & 78.86 & 78.97 & 79.01 & 78.98 & 79.06 \\
                                     &                                 &                        & None                              & 68.83 & 74.84 & 76.39 & 77.05 & 77.60 & 78.00 & 78.16 & 78.56 & 78.75 \\
                                     &                                 &                        & Off-the-Shelf Artifact Correction & 71.19 & 75.88 & 77.14 & 77.80 & 78.15 & 78.30 & 78.57 & 78.66 & 78.61 \\
                                     &                                 &                        & Task-Targeted Artifact Correction & 74.73 & 77.33 & 78.08 & 78.29 & 78.55 & 78.62 & 78.68 & 78.73 & 78.68 \\
    \midrule
    \multirow{4}{*}{ResNeXt-50}      & \multirow{4}{*}{Top-1 Accuracy} & \multirow{4}{*}{76.99} & Supervised Fine-Tuning            & 74.21 & 76.23 & 76.79 & 77.01 & 77.08 & 77.18 & 77.16 & 77.30 & 77.17 \\
                                     &                                 &                        & None                              & 66.96 & 73.21 & 74.85 & 75.62 & 76.07 & 76.37 & 76.63 & 76.88 & 77.06 \\
                                     &                                 &                        & Off-the-Shelf Artifact Correction & 68.05 & 73.56 & 75.11 & 75.95 & 76.38 & 76.59 & 76.71 & 76.99 & 76.90 \\
                                     &                                 &                        & Task-Targeted Artifact Correction & 72.22 & 75.45 & 76.09 & 76.62 & 76.86 & 76.83 & 76.85 & 76.99 & 76.81 \\
    \midrule
    \multirow{4}{*}{VGG-19}          & \multirow{4}{*}{Top-1 Accuracy} & \multirow{4}{*}{73.44} & Supervised Fine-Tuning            & 69.50 & 72.66 & 73.29 & 73.74 & 73.83 & 73.85 & 73.95 & 74.14 & 74.11 \\
                                     &                                 &                        & None                              & 59.27 & 68.08 & 70.49 & 71.53 & 71.99 & 72.42 & 72.80 & 73.24 & 73.46 \\
                                     &                                 &                        & Off-the-Shelf Artifact Correction & 61.93 & 68.79 & 70.82 & 71.83 & 72.50 & 72.94 & 73.13 & 73.40 & 73.44 \\
                                     &                                 &                        & Task-Targeted Artifact Correction & 67.50 & 71.32 & 72.33 & 72.76 & 73.03 & 73.16 & 73.50 & 73.48 & 73.44 \\
    \bottomrule
\end{tabular}
}
\captionof{table}{Results for classification models.}

\resizebox{\textwidth}{!}{
    \begin{tabular}{llllrrrrrrrrr}
    \toprule
    Model                       & Metric               & Reference              & Mitigation                        & Q=10  & Q=20  & Q=30  & Q=40  & Q=50  & Q=60  & Q=70  & Q=80  & Q=90  \\
    \midrule
    \multirow{4}{*}{FasterRCNN} & \multirow{4}{*}{mAP} & \multirow{4}{*}{35.37} & Supervised Fine-Tuning            & 29.09 & 33.34 & 34.72 & 35.08 & 35.49 & 35.82 & 35.96 & 36.06 & 36.17 \\
                                &                      &                        & None                              & 20.35 & 30.03 & 32.59 & 33.43 & 34.04 & 34.31 & 34.73 & 34.93 & 35.25 \\
                                &                      &                        & Off-the-Shelf Artifact Correction & 28.45 & 31.86 & 33.10 & 33.85 & 34.05 & 34.47 & 34.70 & 34.77 & 34.71 \\
                                &                      &                        & Task-Targeted Artifact Correction & 31.43 & 33.85 & 34.29 & 34.81 & 34.81 & 34.97 & 35.01 & 34.88 & 34.81 \\
    \midrule
    \multirow{4}{*}{FastRCNN}   & \multirow{4}{*}{mAP} & \multirow{4}{*}{34.02} & Supervised Fine-Tuning            & 28.01 & 31.94 & 33.08 & 33.56 & 33.88 & 34.17 & 34.42 & 34.44 & 34.66 \\
                                &                      &                        & None                              & 19.99 & 29.04 & 31.22 & 32.19 & 32.65 & 33.00 & 33.34 & 33.40 & 33.80 \\
                                &                      &                        & Off-the-Shelf Artifact Correction & 27.62 & 30.91 & 32.04 & 32.56 & 32.78 & 33.18 & 33.28 & 33.48 & 33.44 \\
                                &                      &                        & Task-Targeted Artifact Correction & 30.11 & 32.31 & 33.07 & 33.31 & 33.39 & 33.53 & 33.69 & 33.68 & 33.59 \\
    \midrule
    \multirow{4}{*}{MaskRCNN}   & \multirow{4}{*}{mAP} & \multirow{4}{*}{32.84} & Supervised Fine-Tuning            & 26.32 & 30.48 & 31.79 & 32.21 & 32.55 & 32.83 & 33.11 & 33.20 & 33.32 \\
                                &                      &                        & None                              & 18.35 & 27.58 & 29.83 & 30.80 & 31.32 & 31.62 & 32.02 & 32.29 & 32.62 \\
                                &                      &                        & Off-the-Shelf Artifact Correction & 25.82 & 29.35 & 30.67 & 31.32 & 31.59 & 31.85 & 32.03 & 32.24 & 32.16 \\
                                &                      &                        & Task-Targeted Artifact Correction & 28.48 & 30.85 & 31.71 & 32.00 & 32.19 & 32.24 & 32.35 & 32.43 & 32.26 \\
    \midrule
    \multirow{4}{*}{RetinaNet}  & \multirow{4}{*}{mAP} & \multirow{4}{*}{33.57} & Supervised Fine-Tuning            & 27.64 & 31.97 & 33.03 & 33.50 & 33.80 & 34.12 & 34.30 & 34.33 & 34.40 \\
                                &                      &                        & None                              & 18.76 & 28.23 & 30.63 & 31.59 & 32.27 & 32.57 & 32.88 & 33.02 & 33.42 \\
                                &                      &                        & Off-the-Shelf Artifact Correction & 26.74 & 29.90 & 31.24 & 31.87 & 32.19 & 32.60 & 32.86 & 33.02 & 32.93 \\
                                &                      &                        & Task-Targeted Artifact Correction & 29.66 & 31.86 & 32.73 & 32.97 & 32.98 & 33.13 & 33.24 & 33.23 & 33.09 \\
    \bottomrule
\end{tabular}
}
\captionof{table}{Results for detection models.}

\resizebox{\textwidth}{!}{
    \begin{tabular}{llllrrrrrrrrr}
    \toprule
    Model                                            & Metric                & Reference              & Mitigation                        & Q=10  & Q=20  & Q=30  & Q=40  & Q=50  & Q=60  & Q=70  & Q=80  & Q=90  \\
    \midrule
    \multirow{4}{*}{HRNetV2 + C1}                    & \multirow{4}{*}{mIoU} & \multirow{4}{*}{40.59} & Supervised Fine-Tuning            & 34.76 & 37.35 & 38.74 & 38.78 & 39.27 & 39.75 & 39.98 & 39.86 & 39.96 \\
                                                     &                       &                        & None                              & 24.95 & 35.16 & 38.03 & 38.52 & 39.02 & 40.09 & 40.50 & 40.41 & 40.54 \\
                                                     &                       &                        & Off-the-Shelf Artifact Correction & 32.30 & 36.54 & 38.40 & 38.52 & 40.08 & 40.44 & 40.46 & 40.22 & 40.60 \\
                                                     &                       &                        & Task-Targeted Artifact Correction & 34.14 & 37.61 & 39.23 & 39.24 & 39.92 & 40.53 & 40.62 & 40.39 & 40.55 \\
    \midrule
    \multirow{4}{*}{MobileNetV2 (dilated) + C1 (ds)} & \multirow{4}{*}{mIoU} & \multirow{4}{*}{29.52} & Supervised Fine-Tuning            & 19.07 & 22.37 & 23.43 & 23.62 & 23.60 & 24.15 & 24.44 & 24.37 & 24.46 \\
                                                     &                       &                        & None                              & 13.92 & 24.03 & 27.13 & 27.75 & 27.73 & 28.86 & 29.37 & 29.35 & 29.43 \\
                                                     &                       &                        & Off-the-Shelf Artifact Correction & 21.17 & 25.27 & 27.31 & 27.16 & 29.14 & 29.32 & 29.26 & 29.06 & 29.54 \\
                                                     &                       &                        & Task-Targeted Artifact Correction & 24.74 & 27.37 & 28.44 & 28.33 & 29.19 & 29.56 & 29.54 & 29.38 & 29.52 \\
    \midrule
    \multirow{4}{*}{ResNet101 + UPerNet}             & \multirow{4}{*}{mIoU} & \multirow{4}{*}{41.08} & Supervised Fine-Tuning            & 35.32 & 37.41 & 38.27 & 38.28 & 38.55 & 38.59 & 38.72 & 38.58 & 38.70 \\
                                                     &                       &                        & None                              & 26.14 & 36.70 & 39.45 & 39.81 & 39.55 & 40.47 & 40.98 & 40.97 & 41.07 \\
                                                     &                       &                        & Off-the-Shelf Artifact Correction & 33.90 & 37.39 & 39.12 & 39.38 & 40.32 & 40.58 & 40.78 & 40.79 & 41.04 \\
                                                     &                       &                        & Task-Targeted Artifact Correction & 35.82 & 38.67 & 39.96 & 39.98 & 40.22 & 40.79 & 40.97 & 40.91 & 41.00 \\
    \midrule
    \multirow{4}{*}{ResNet101 (dilated) + PPM}       & \multirow{4}{*}{mIoU} & \multirow{4}{*}{40.26} & Supervised Fine-Tuning            & 31.86 & 35.45 & 36.73 & 36.94 & 36.91 & 37.33 & 37.67 & 37.55 & 37.65 \\
                                                     &                       &                        & None                              & 25.68 & 35.19 & 37.76 & 38.43 & 38.24 & 39.27 & 40.03 & 40.17 & 40.21 \\
                                                     &                       &                        & Off-the-Shelf Artifact Correction & 31.44 & 35.86 & 38.01 & 38.26 & 39.54 & 39.73 & 39.94 & 40.06 & 40.22 \\
                                                     &                       &                        & Task-Targeted Artifact Correction & 33.99 & 37.63 & 39.04 & 39.11 & 39.38 & 39.73 & 40.07 & 40.11 & 40.10 \\
    \midrule
    \multirow{4}{*}{ResNet18 (dilated) + PPM}        & \multirow{4}{*}{mIoU} & \multirow{4}{*}{36.65} & Supervised Fine-Tuning            & 29.84 & 32.33 & 33.08 & 33.01 & 33.38 & 33.61 & 33.50 & 33.29 & 33.33 \\
                                                     &                       &                        & None                              & 21.16 & 31.99 & 34.72 & 35.36 & 35.41 & 36.16 & 36.56 & 36.60 & 36.59 \\
                                                     &                       &                        & Off-the-Shelf Artifact Correction & 28.64 & 32.59 & 34.56 & 34.53 & 35.96 & 36.21 & 36.29 & 36.25 & 36.64 \\
                                                     &                       &                        & Task-Targeted Artifact Correction & 31.69 & 34.55 & 35.80 & 35.80 & 36.12 & 36.50 & 36.66 & 36.54 & 36.60 \\
    \midrule
    \multirow{4}{*}{ResNet50 + UPerNet}              & \multirow{4}{*}{mIoU} & \multirow{4}{*}{39.21} & Supervised Fine-Tuning            & 32.88 & 35.11 & 35.94 & 35.90 & 36.41 & 36.58 & 36.63 & 36.49 & 36.55 \\
                                                     &                       &                        & None                              & 24.29 & 34.78 & 37.34 & 37.71 & 37.70 & 38.57 & 39.12 & 39.13 & 39.16 \\
                                                     &                       &                        & Off-the-Shelf Artifact Correction & 31.83 & 35.52 & 37.20 & 37.26 & 38.44 & 38.67 & 38.87 & 38.86 & 39.12 \\
                                                     &                       &                        & Task-Targeted Artifact Correction & 34.36 & 36.94 & 38.17 & 38.07 & 38.55 & 38.93 & 39.14 & 39.06 & 39.09 \\
    \midrule
    \multirow{4}{*}{ResNet50 (dilated) + PPM}        & \multirow{4}{*}{mIoU} & \multirow{4}{*}{38.91} & Supervised Fine-Tuning            & 32.26 & 35.33 & 36.04 & 36.04 & 36.53 & 36.75 & 36.93 & 36.71 & 36.92 \\
                                                     &                       &                        & None                              & 23.05 & 33.95 & 36.66 & 37.07 & 37.40 & 38.58 & 38.93 & 38.70 & 38.86 \\
                                                     &                       &                        & Off-the-Shelf Artifact Correction & 28.36 & 32.69 & 35.24 & 35.31 & 37.74 & 38.04 & 38.18 & 38.13 & 38.73 \\
                                                     &                       &                        & Task-Targeted Artifact Correction & 31.92 & 35.43 & 37.04 & 36.92 & 38.05 & 38.69 & 38.79 & 38.52 & 38.74 \\
    \bottomrule
\end{tabular}
}
\captionof{table}{Results for segmentation models.}

\resizebox{\textwidth}{!}{
    \begin{tabular}{llllrrrrrrrrr}
    \toprule
    Model                        & Metric                          & Reference              & Mitigation                        & Q=10  & Q=20  & Q=30  & Q=40  & Q=50  & Q=60  & Q=70  & Q=80  & Q=90  \\
    \midrule
    \multirow{4}{*}{Chai et. al} & \multirow{4}{*}{Patch Accuracy} & \multirow{4}{*}{93.83} & Supervised Fine-Tuning            & 53.04 & 54.81 & 56.12 & 56.86 & 57.34 & 57.66 & 58.07 & 58.58 & 59.41 \\
                                 &                                 &                        & None                              & 49.87 & 49.46 & 49.48 & 49.86 & 50.33 & 50.74 & 51.12 & 51.83 & 53.26 \\
                                 &                                 &                        & Off-the-Shelf Artifact Correction & 50.56 & 50.20 & 50.34 & 50.63 & 51.14 & 51.63 & 52.26 & 53.22 & 54.12 \\
                                 &                                 &                        & Task-Targeted Artifact Correction & 51.99 & 52.46 & 52.70 & 52.98 & 53.21 & 53.33 & 53.48 & 53.58 & 53.28 \\
    \midrule
    \multirow{4}{*}{Wang et. al} & \multirow{4}{*}{Accuracy}       & \multirow{4}{*}{99.96} & Supervised Fine-Tuning            & 89.70 & 96.69 & 98.41 & 99.04 & 99.44 & 99.54 & 99.65 & 99.88 & 99.91 \\
                                 &                                 &                        & None                              & 54.25 & 91.08 & 97.56 & 98.79 & 99.28 & 99.38 & 99.60 & 99.81 & 99.84 \\
                                 &                                 &                        & Off-the-Shelf Artifact Correction & 50.00 & 50.00 & 50.00 & 50.00 & 50.00 & 50.00 & 50.00 & 50.00 & 50.00 \\
                                 &                                 &                        & Task-Targeted Artifact Correction & 50.00 & 50.00 & 50.00 & 50.00 & 50.00 & 50.00 & 50.00 & 50.00 & 50.00 \\
    \bottomrule
\end{tabular}
}
\captionof{table}{Results for forensics models.}

\subsection{Reference Results}

The following table gives the reference numbers of the pretrained weights as evaluated by our system on uncomressed images.

\begin{center}
    \begin{tabular}{lr}
    \toprule
    Model                      & Value                                                             \\
    \midrule
    \multicolumn{2}{@{}l}{ImageNet Classification, Metric: Top-1 Accuracy}                         \\
    \cmidrule[\cmidrulewidth]{1-2}
    ResNet 18                  & 68.84                                                             \\
    ResNet 50                  & 75.31                                                             \\
    ResNet 101                 & 76.91                                                             \\
    ResNeXt 50                 & 76.99                                                             \\
    ResNeXt 101                & 78.81                                                             \\
    VGG 19                     & 73.44                                                             \\
    MobileNetV2                & 70.72                                                             \\
    InceptionV3                & 77.33                                                             \\
    EfficientNet B3            & 83.98                                                             \\
    \midrule
    \multicolumn{2}{@{}l}{COCO Object Detection and Instance Segmentation, Metric: mAP}            \\
    \cmidrule[\cmidrulewidth]{1-2}
    FastRCNN                   & 34.02                                                             \\
    FasterRCNN                 & 35.38                                                             \\
    RetinaNet                  & 33.57                                                             \\
    MaskRCNN                   & 32.84                                                             \\
    \midrule
    \multicolumn{2}{@{}l}{ADE20k Semantic Segmentation, Metric: mIoU}                              \\
    \cmidrule[\cmidrulewidth]{1-2}
    HRNetV2 + C1               & 40.59                                                             \\
    MobileNetV2 (dilated) + C1 & 29.52                                                             \\
    ResNet 18 (dilated) + PPM  & 36.65                                                             \\
    ResNet 50 (dilated) + PPM  & 38.91                                                             \\
    ResNet 101                 & 41.08                                                             \\
    ResNet 101 (dilated) + PPM & 40.26                                                             \\
    \midrule
    \multicolumn{2}{@{}l}{Forensics (dataset varies), Metric: Accuracy (exact formulation varies)} \\
    \cmidrule[\cmidrulewidth]{1-2}
    Chai \etal                 & 93.84                                                             \\
    Wang \etal                 & 99.96                                                             \\
    \bottomrule
\end{tabular}
\end{center}

\clearpage

\section{Throughput}

\begin{figure}[t]
    \centering
    \includegraphics[width=0.47\textwidth]{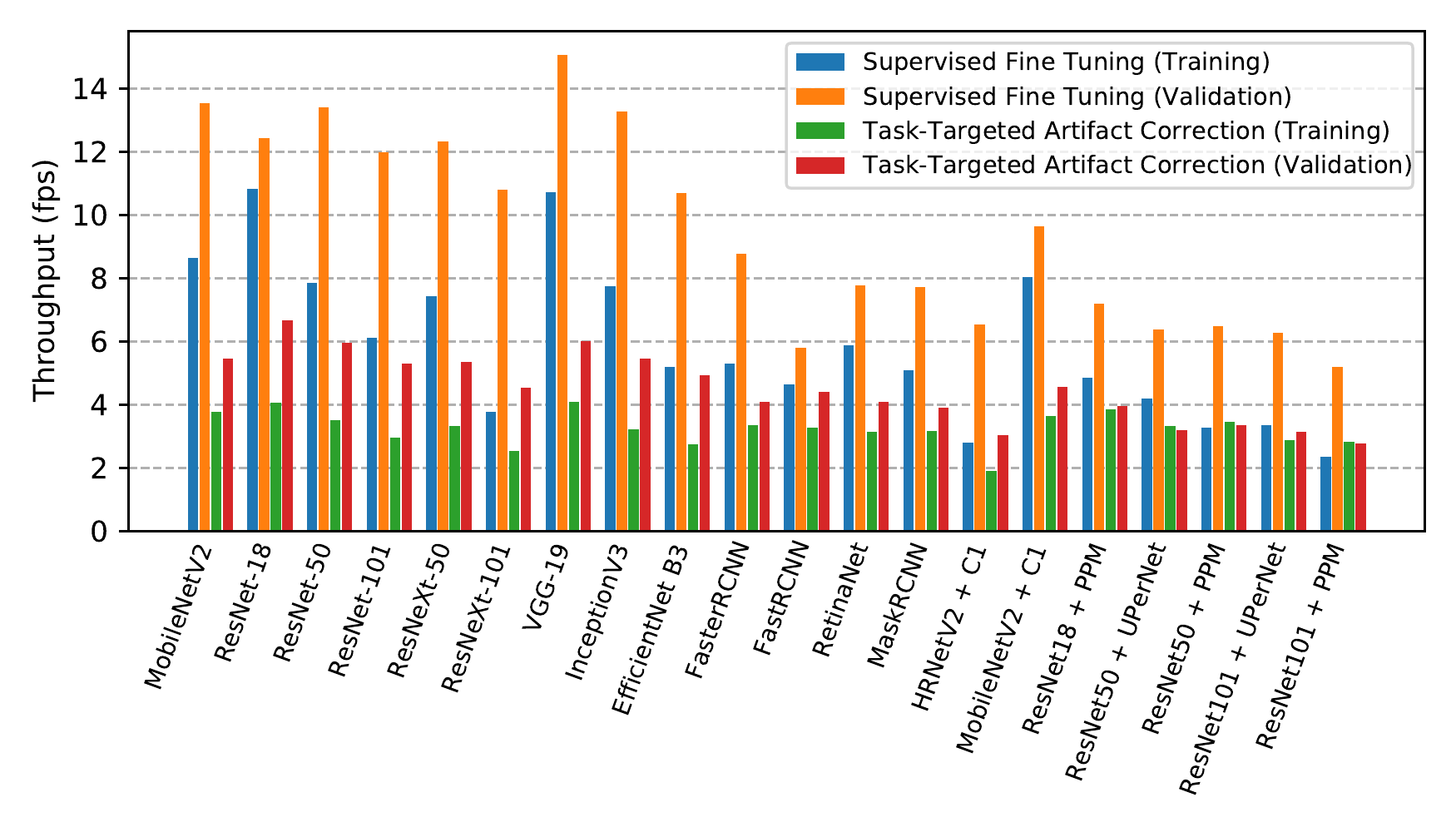}
    \vspace{-0.15in}
    \caption{Throughput results for all tested models.}
    \label{fig:throughput}
    \vspace{-0.003in}
\end{figure}

Although artifact correction is mentioned in prior works and presented here as a viable compression mitigation technique, we would be remiss if we did not note the slower throughput of these methods. In Figure \ref{fig:throughput} we show the training and inference throughput for batches of size 1 of both the artifact correction mitigation as well as the supervised fine tuning mitigation. These results are critical when considering which mitigation method is most viable for a particular application: although Task-Targeted Artifact Correction is more flexible, it comes with a cost in throughput.

\newpage
\section{Multihead Results}

Table \ref{tab:mh_and_transfer} shows the raw numbers for our transfer and multi-head experiments.

\begin{table}[t]
    \centering
    \footnotesize
    \renewcommand{\arraystretch}{1.2}
    \renewcommand{\tabcolsep}{2.4mm}
    \caption{Transfer and multihead results. Reference indicates the performance of the pretrained weights on uncompressed images. \textbf{Best} result in bold, \underline{second best} underlined.}
    \label{tab:mh_and_transfer}
    \vspace{-0.05in}
    \resizebox{0.47\textwidth}{!}{
        \setlength{\cmidrulewidth}{0.01em}
\begin{tabular}{@{}lrrrrr@{}}
    \toprule
    Mitigation                        & Q=10              & Q=20              & Q=30              & Q=40              & Q=50              \\
    \midrule
    \multicolumn{6}{@{}l}{\textbf{HRNetV2 + C1}, Reference: \textbf{40.59 mIoU} (Semantic Segmentation)}\\
    \cmidrule[\cmidrulewidth]{1-6}
    None                              & 24.95             & 35.16             & 38.03             & 38.52             & 39.02             \\
    Off-the-Shelf Artifact Correction & 32.30             & 36.54             & 38.40             & 38.52             & 40.08             \\
    Supervised Fine-Tuning            & \textbf{34.76}    & 37.35             & 38.74             & 38.78             & 39.27             \\
    Task-Targeted Artifact Correction & 34.14             & \underline{37.61} & \underline{39.23} & \underline{39.24} & \textbf{39.92}    \\
    MobileNetV2 Transfer              & 33.20             & 37.05             & 38.93             & 38.95             & 39.33             \\
    ResNet18 Transfer                 & 33.77             & 37.44             & 39.22             & 39.21             & 39.25             \\
    Multihead (Three Model)           & \underline{34.38} & \textbf{37.68}    & \textbf{39.39}    & \textbf{39.39}    & \underline{39.72} \\
    \midrule
    \multicolumn{6}{@{}l}{\textbf{Faster RCNN}, Reference: \textbf{35.37 mAP} (Object Detection)}\\
    \cmidrule[\cmidrulewidth]{1-6}
    None                              & 20.35             & 30.03             & 32.59             & 33.43             & 34.04             \\
    Off-the-Shelf Artifact Correction & 28.45             & 31.86             & 33.10             & 33.85             & 34.05             \\
    Supervised Fine-Tuning            & 29.09             & 33.34             & \textbf{34.72}    & \textbf{35.08}    & \textbf{35.49}    \\
    Task-Targeted Artifact Correction & \textbf{31.43}    & \textbf{33.85}    & \underline{34.29} & \underline{34.81} & \underline{34.81} \\
    MobileNetV2 Transfer              & 30.05             & 33.04             & 33.86             & 34.35             & 34.48             \\
    ResNet18 Transfer                 & 30.72             & 33.30             & 34.20             & 34.57             & 34.66             \\
    Multihead (Two Model)             & \underline{31.09} & 33.39             & 34.19             & 34.67             & 34.68             \\
    Multihead (Three Model)           & 30.96             & \underline{33.41} & \underline{34.29} & 34.68             & 34.70             \\
    \midrule
    \multicolumn{6}{@{}l}{\textbf{ResNet-101}, Reference: \textbf{76.91, Top-1 Accuracy} (Image Classification)}\\
    \cmidrule[\cmidrulewidth]{1-6}
    None                              & 66.12             & 73.00             & 74.65             & 75.39             & 75.83             \\
    Off-the-Shelf Artifact Correction & 67.91             & 73.64             & 75.09             & 75.84             & 76.23             \\
    Supervised Fine-Tuning            & \textbf{74.63}    & \textbf{76.50}    & \textbf{77.07}    & \textbf{77.20}    & \textbf{77.27}    \\
    Task-Targeted Artifact Correction & \underline{72.99} & \underline{75.53} & \underline{76.30} & \underline{76.60} & \underline{76.59} \\
    MobileNetV2 Transfer              & 72.18             & 75.35             & 76.15             & 76.49             & 76.58             \\
    ResNet18 Transfer                 & 71.80             & 75.05             & 76.00             & 76.40             & 76.49             \\
    \midrule
    \multicolumn{6}{@{}l}{\textbf{ResNet-50}, Reference: \textbf{75.31, Top-1 Accuracy} (Image Classification)}\\
    \cmidrule[\cmidrulewidth]{1-6}
    None                              & 63.43             & 71.20             & 73.23             & 74.10             & 74.43             \\
    Off-the-Shelf Artifact Correction & 66.90             & 72.45             & 73.95             & 74.60             & 74.93             \\
    Supervised Fine-Tuning            & \textbf{73.18}    & \textbf{75.46}    & \textbf{76.02}    & \textbf{76.24}    & \textbf{76.36}    \\
    Task-Targeted Artifact Correction & 70.48             & 73.56             & 74.39             & 74.81             & 74.94             \\
    Multihead (Two Model)             & \underline{71.66} & 74.14             & 74.90             & \underline{75.05} & 75.10             \\
    Multihead (Three Model)           & 71.49             & \underline{74.23} & \underline{74.96} & 75.05             & \underline{75.15} \\
    \bottomrule
\end{tabular}
    }
    \vspace{-0.05in}
\end{table}

\end{document}